\documentclass[final,3p,12pt]{elsarticle}
\journal{Preprint}
\let\today\relax
\makeatletter
\def\ps@pprintTitle{%
    \let\@oddhead\@empty
    \let\@evenhead\@empty
    \def\@oddfoot{\footnotesize\itshape
         {Preprint} \hfill\today}%
    \let\@evenfoot\@oddfoot
    }
\makeatother
\usepackage{amsmath}
\usepackage{amsfonts}
\usepackage{amssymb}
\usepackage{amsthm}
\usepackage{bm}
\usepackage{mathtools}
\usepackage{indentfirst}
\usepackage{booktabs}
\usepackage{float}
\usepackage[dvipsnames, table]{xcolor}
\usepackage{multirow}
\usepackage{subcaption}
\usepackage[hidelinks]{hyperref}
\usepackage{tikz}
\usepackage{array}
\usepackage[flushleft]{threeparttable}
\usepackage[export]{adjustbox}
\usepackage[column=0]{cellspace}
\tikzset{every picture/.style={line width=0.75pt}} 

\numberwithin{equation}{section}

\renewcommand{\appendix}{\par
  \setcounter{section}{0}
  \setcounter{subsection}{0}
  \gdef\thesection{\Alph{section}}
}

\usepackage{lineno}

\usepackage{ulem}

\begin{document}
\begin{frontmatter}
\title{Fusion-DeepONet: A Data-Efficient Neural Operator for Geometry-Dependent Hypersonic and Supersonic Flows}
\author[1]{Ahmad Peyvan}
\author[2]{Varun Kumar}
\author[1]{George Em Karniadakis}
\affiliation[1]{organization={Division of Applied Mathematics, 182 George Street, Brown University},
    city={Providence},state={RI}, postcode={02912},country={USA}
            }
\affiliation[2]{organization={School of Engineering, 184 Hope St, Brown University},
    city={Providence},state={RI}, postcode={02912},country={USA}
            }

\begin{abstract}
Shape optimization is essential in aerospace vehicle design, including reentry systems, missile configurations, and propulsion system components, as it directly influences aerodynamic efficiency, structural integrity, and overall mission success. Rapid and accurate prediction of external and internal flows accelerates design iterations and reduces development cost of the design concept across diverse operating conditions. To this end, we develop a new variant of Deep Operator Network (DeepONet), called Fusion-DeepONet as a fast surrogate model for geometry-dependent hypersonic and supersonic flow fields. We evaluated Fusion-DeepONet in learning two external hypersonic flows and a supersonic shape-dependent internal flow problem. First, we compare the performance of Fusion-DeepONet with state-of-the-art neural operators to learn inviscid hypersonic flow around semi-elliptic blunt bodies for two grid types: uniform Cartesian and irregular grids. Fusion-DeepONet provides comparable accuracy to parameter-conditioned U-Net on uniform grids while outperforming MeshGraphNet and Vanilla-DeepONet on irregular and arbitrary grids. Fusion-DeepONet requires significantly fewer trainable parameters than U-Net, MeshGraphNet, and FNO, making it computationally efficient. For the second hypersonic problem, we set up Fusion-DeepONet to map from geometry and free stream Mach number to the temperature field around a reentry capsule traveling at hypersonic speed. This fast surrogate is then improved to predict the spatial derivative of the temperature, resulting in an accurate prediction of heat flux at the surfaces of the capsule. To enhance the accuracy of spatial derivative prediction, we introduce a derivative-enhanced loss (DEL) term with the least computation overhead. For the third problem, we show that Fusion-DeepONet outperforms MeshGraphNet in learning geometry-dependent supersonic flow in a converging-diverging nozzle configuration attached to an isolator duct. For all the problems, we used high-fidelity simulations with a high-order entropy-stable DGSEM solver to generate training datasets with limited samples. We also analyze the basis functions of the Fusion-DeepONet model using a singular value decomposition approach. This analysis reveals that Fusion-DeepONet generalizes effectively to unseen geometries and grid points, demonstrating its robustness in scenarios with limited training data. The source code, along with the datasets, can be found at the following URL: \url{https://github.com/ahmadpeyvan/Fusion-DeepONet}.

\end{abstract}



\end{frontmatter}

{\renewcommand\arraystretch{0.75}
 
\section{Introduction}
High-speed flows involve complex interactions between shock waves, boundary layers, chemical reactions, and thermal effects, which vary significantly with the vehicle's geometry. These interactions require large-scale, high-fidelity, and computationally expensive simulations, particularly when exploring multiple geometric configurations. Given this, running full simulations for every potential design is computationally intractable. Multiple design parameters (e.g., shape, material properties, surface roughness) must be tested in hypersonic and supersonic vehicle design to find the optimal configuration. Geometry-dependent surrogate models can efficiently explore various geometries, helping engineers optimize designs without running costly CFD simulations. Moreover, real-time feedback is often necessary in scenarios such as launch vehicle guidance or rapid prototyping. Surrogate models can offer fast predictions during these operational stages, enabling decision-making on-the-fly based on geometric changes or modifications in flow conditions. Surrogate models based on multi-fidelity setup have traditionally been used for modeling flow around objects but have primarily focused on sub-sonic regimes \cite{multifid_airfoil_optim, multifid_design_optim, multifid_manifold, multifid_naca_parametric, multifid_reduced_basis, multifid_transonic_airfoil, multifid_unsteady_airfoil}. Recently, neural networks have emerged as an important class of surrogate models for inferring flow fields. With their ability to learn complex non-linear mapping between input and output functions, neural networks provide a promising landscape for surrogate model development to replace expensive CFD simulations. The research impetus on neural network-based surrogate models for aerodynamic flows has led to a surge in deep learning-based surrogates \cite{neural_aerodyn_Liao, neural_aerodyn_Renganathan, neural_design_Tao, neural_optim_Pin, neural_surrogate_Zhang, pinn_airfoil_Harada, neural_aerodyn_Bhatnagar}. However, these existing methods mainly focus on sub-sonic flows, likely due to the challenges of solving hypersonic problems that typically involve extreme variations in flow fields in space and time.

An important limitation of traditional neural network-based methods for scientific tasks is the lack of discretization invariance property. The discretization invariance is important since the solution space often requires interpolation over the entire domain rather than being limited to pre-defined grid points. As an alternative, operator networks offer the unique property of discretization invariance in temporal or spatial domains, allowing for better generalization capabilities over Vanilla neural networks. Some prominent operator-based frameworks include Deep Operator Network (DeepONet) \cite{deeponet}, Fourier Neural Operator (FNO) \cite{FNO}, Laplacian Neural Operator (LNO) \cite{LNO}, Wavelet Neural Operator (WNO) \cite{WNO}, to name a few. Operator-based methods have seen significant growth in their use as surrogate models for scientific tasks that require significant computational overhead with traditional numerical tools. Operator networks have also been explored for aerodynamic modeling, and some significant works include \cite{don_airfoil, Geom_FNO, MeshGraphNet, geometry_FNO, operator_aerodyn_icing}. While the application of these operator methods has been explored for aerodynamic problems such as flow around airfoils, their use has largely remained unexplored for predicting flow around hypersonic bodies, especially when the shape of the hypersonic body is changing. Hence, a research opportunity exists to determine the effectiveness of these methods for geometry-dependent hypersonic flow problems and further develop a general  operator-based framework to tackle such tasks.

Prediction of hypersonic and supersonic flows is challenging due to extremely high-gradient fields around the flying vehicle. For instance, the formation of the bow shock in front of blunt bodies creates a high-temperature area between the nose and the shock. High temperature reduces the density and creates near-vacuum conditions in some locations. Therefore, the data generation for hypersonic and supersonic flows is challenging; hence, only a few high-fidelity numerical solvers exist. Recently, Peyvan \emph{et al.} \cite{PEYVAN2023112310} developed a high-order entropy stable scheme that can robustly simulate supersonic to hypersonic flows and provide high-fidelity data. Due to the scarcity of high-speed flow data and the complications involved with such simulations, limited studies focus on developing surrogate models for such flows. For instance, Way \emph{et al.} \cite{way2024hypernetwork} created a surrogate model of hypersonic flows that maps from free stream flow conditions, including Mach number, angle of attack, and altitude, to the temperature and shear stresses on the surface of a hypersonic vehicle. However, their framework cannot handle a geometry-dependent flow field. In another case, Rataczak \emph{et al.} \cite{rataczak2024surrogate} created surrogate models for hypersonic aerodynamics using Gaussian process regression. They created three surrogate models. The first surrogate predicts the stagnation point heat flux and axial force coefficients. The second model predicts convective heat flux contours on the vehicle surface. Finally, the third surrogate is designed to predict the vehicle lift and drag coefficient by varying the angle of attack. Scherding \emph{et al.} \cite{scherding2025adaptive} employed reduced order nonlinear approximation with adaptive learning procedure to train a surrogate model for the thermodynamics in CFD solvers with application to unsteady hypersonic flows in chemical non-equilibrium. Schouler \emph{et al.} \cite{schouler2023machine} trained Krigging and fully connected neural networks to predict the pressure, heat flux stagnation coefficient, friction, and heat flux coefficient distributions in the rarefied portion of a reentry vehicle. The previous hypersonic and supersonic surrogate modeling lacks the consideration of the flying object geometry variation and learning of the flow field. 

In high-speed flow simulations, key quantities of interest typically include shear stress and heat flux acting on the surfaces of flying vehicles. Accurate prediction of these quantities necessitates precise computations of spatial gradients in the solution fields. Neural operators are conventionally trained by minimizing the mean squared error between predicted and ground truth solution fields. However, this approach lacks explicit consideration of derivative information in the target data. A few recent studies have addressed this gap. Cao et al. \cite{cao2025derivative} proposed a derivative-informed neural operator (DINO) framework, which trains surrogate models for costly parameter-to-observable mappings by simultaneously minimizing both output errors and Jacobian errors through a Sobolev loss. This method significantly reduces the required number of iterations during minimization of a cost function in an optimization loop. In the context of DeepONet, the Jacobian is calculated with respect to the network inputs. Similarly, Qiu et al. \cite{qiu2024derivative} extended the DINO concept by developing a derivative-enhanced Deep Operator Network, incorporating linear dimensionality reduction for high-dimensional parameter inputs and a derivative-based loss term. This enhancement improves both solution accuracy and the precision of solution-to-parameter derivatives, particularly in scenarios with limited training data. In our current study, we focus specifically on the derivative with respect to spatial coordinates. We identify the optimal form of derivative-informed loss to accurately predict quantities like heat flux and determine the derivative operator most compatible with high-fidelity CFD datasets.

Specifically, we develop a new Fusion-DeepONet framework and compare its performance with other neural operators to map geometry to the hypersonic flow field. We test Fusion-DeepONet to learn three geometry-dependent problems, including inviscid hypersonic flow over semi-ellipse bodies, viscous hypersonic flow around a reentry capsule, and internal supersonic compressible flow inside a converging-diverging nozzle. Temperature predictions are used to compute the heat flux exerted on the surfaces of reentry capsules. We summarize the main contributions of our work below:

\begin{itemize}
    \item We develop Fusion-DeepONet, a geometry-aware neural operator that accurately predicts hypersonic flow fields on uniform and irregular grids using scarce data. It achieves comparable accuracy to parameter-conditioned U-Net for uniform grid while outperforming MeshGraphNet and Vanilla-DeepONet for irregular grids. With the lowest number of trainable parameters among the models studied, Fusion-DeepONet ensures computational efficiency, scalability, and applicability to complex geometries.
    
    \item The singular value decomposition (SVD) of the trunk network reveals that Fusion-DeepONet learns both high- and low-frequency features of the hypersonic solution better than the vanilla model.

    \item We demonstrate that Fusion-DeepONet accurately infers internal and external viscous high-speed flows across geometric configurations.
    
    \item We introduce a derivative-enhanced loss function that enables Fusion-DeepONet to predict the temperature field around a reentry capsule with sufficient accuracy to yield accurate heat flux estimates on its surface.
\end{itemize}

The rest of the article is arranged as follows: We describe the governing equations solved for the data generation. Next, we explain the data generation procedure for training and testing the neural networks. The methodology section describes the structure of Fusion-DeepONet and the loss functions used for training. We then illustrate and interpret the results. Finally, we conclude the paper with a summary. The necessary information and data for reproducibility are included in the appendix section.

\section{Governing Equations}\label{sec:problems}
We are interested in learning geometry-dependent flow fields generated by solving the 2D compressible Navier-Stokes equations given by

\begin{equation}
\frac{\partial \mathbf{U}}{\partial t}+\frac{\partial \mathbf{F}^a}{\partial x} +\frac{\partial \mathbf{G}^a}{\partial y}  = \frac{1}{Re} \left[ \frac{\partial \mathbf{F}^v}{\partial x} +\frac{\partial \mathbf{G}^v}{\partial y}\right]
    \label{system_equation}
\end{equation}
where the vector of conserved variables $\mathbf{U}$ is defined as 
\begin{equation}
\mathbf{U}=\left(\rho, \rho u, \rho v, \rho E\right)^T,
    \label{Uconserv}
\end{equation}
and convective flux vectors $\mathbf{F}$ in x-direction and $\mathbf{G}$ in y-direction are:  

\begin{equation}
\mathbf{F}^a=\begin{pmatrix}
    \rho u \\\rho u^2+p\\ \rho uv\\u(\rho E+p)
\end{pmatrix},\quad \mathbf{G}^a=\begin{pmatrix}
    \rho v \\\rho vu\\ \rho v^2+p\\v(\rho E+p). 
\end{pmatrix}.
    \label{adevctive_flux}
\end{equation}
The viscous fluxes are expressed as
\begin{equation}
\mathbf{F}^v=\begin{pmatrix}
    0 \\\tau_{xx}\\ \tau_{xy}\\u\tau_{xx}+v\tau_{xy}-q_x
\end{pmatrix},\quad \mathbf{G}^v=\begin{pmatrix}
    0 \\\tau_{xy}\\ \tau_{yy}\\u\tau_{xy}+v\tau_{yy}-q_y. 
\end{pmatrix}.
    \label{viscous}
\end{equation}
In Eq.~\eqref{Uconserv}, $\rho$ denotes density. The terms $u$ and $v$ represent velocity components along x and y directions respectively. $\rho E$ indicates the total energy of the gas mixture, including internal and kinetic energies defined as
\begin{equation}
\rho E = \frac{p}{\gamma-1}+ \frac{1}{2}\rho(u^2+v^2),
    \label{total_energy}
\end{equation}
where $\gamma$ is the ratio of specific heat capacity for fluid medium with value $\gamma=1.4$ for air. In Eq.\eqref{total_energy}, $p$ denotes pressure. The Newtonian stress tensor using the Stokes's hypothesis is:  

\begin{align}
\tau_{xx} &= 2\mu\frac{\partial u}{\partial x}
  - \frac{2}{3}\,\mu\Bigl(\frac{\partial u}{\partial x} + \frac{\partial v}{\partial y}\Bigr),
\qquad
\tau_{yy} = 2\mu\frac{\partial v}{\partial y}
  - \frac{2}{3}\,\mu\Bigl(\frac{\partial u}{\partial x} + \frac{\partial v}{\partial y}\Bigr), 
\nonumber\\
\tau_{xy} &= \mu\Bigl(\frac{\partial u}{\partial y} + \frac{\partial v}{\partial x}\Bigr),
\label{eq:newtonian_components}
\end{align}
where $\mu$ is the constant non-dimensional dynamic viscosity. According to Fourier's law of heat conduction, the heat fluxes in Eq.\eqref{viscous} take the following form: 
\begin{equation}
q_x=-\kappa\frac{\partial T}{\partial x},\quad q_y=-\kappa\frac{\partial T}{\partial y}, 
    \label{eq:heat_conduct}
\end{equation}
where $\kappa$ is defined as $\kappa=\dfrac{\mu}{Pr}\dfrac{\gamma}{\gamma-1}$. $Pr$ denotes the Prandtl number and $Re$ refers to reference Reynolds number. In Eq~\eqref{eq:heat_conduct}, $T$ refers to the non-dimensional temperature. For non-dimensionalization, we have set the reference values of quantities as the free stream value. In detail, we set the reference velocity equal to the speed of sound computed using the free stream condition. The reference temperature is also computed using the ideal gas law with pressure and density set as free stream conditions and gas constant equal to $R=287.058$ JKg/K. The non-dimensional temperature follows the ideal gas law as 
\begin{equation}
p=\gamma \rho T.
    \label{eq:ideal_gas}
\end{equation}

\section{Data  Generation}
In this study, we intend to learn high-speed flows for three geometry-dependent problems, including inviscid hypersonic flow around semi-ellipses, viscous supersonic flow inside a converging-diverging nozzle attached to an isolator duct, and a reentry capsule traveling at hypersonic speed. All problems involve samples with varying geometries. We employ the Trixi.jl solver \cite{schlottkelakemper2020trixi,ranocha2022adaptive,schlottkelakemper2021purely} to solve Eq.~\eqref{system_equation}. In Trixi.jl, the entropy-stable Discontinuous Galerkin Spectral Element Method (ES-DGSEM) is used to solve the Euler and Navier-Stokes equations. For all three problems, we first select two or three geometric and flow parameters and then define an interval for the change. Then, we use random selection methods to create a list of cases for training and testing neural operators. In the following, we explain the specifics of the data generation procedure.

\subsection{Inviscid hypersonic flow around semi-ellipses}
We solve the hyperbolic part of Eq.~\eqref{system_equation} in a physical domain depicted in Fig.~\ref{fig:domain}. According to Fig.~\ref{fig:domain}, supersonic inflow is imposed on the blue boundaries while the extrapolation condition is imposed on the red lines. The geometry boundary, defined by the black curve, denotes the slip wall boundary condition. We use two parameters that define the semi-ellipse for generating 36 unique geometries: the major axes ($a\in[0.5, 3.0]$) and the minor axes ($b\in[0.5, 1.8]$) as shown in Fig.~\ref{fig:cases}. Latin hypercube sampling is used for defining these 36 samples from the parameter space, see appendix~\ref{append:datasets} for more details. Individual mesh files are generated using an automation script integrated with GMSH. Fig.~\ref{fig:mesh} (a)-(c) shows the quadrilateral mesh elements generated for three cases. The reference length size of elements is set at $0.25$ for all the points. According to Fig.~\ref{fig:mesh}, the initial and final meshes for the domain are entirely different due to the adaptive mesh refinement (AMR) strategy.

 The flow field is initialized using a free stream condition with Mach number set as $M=10$ and $\rho_0=1.4$, $u_0=-10$, $v_0=0$, and $p_0=1$. The specific heat ratio is set at $\gamma=1.4$. Shock stabilization is performed using the blending of low and high-order flux functions. The approximate Riemann solver of Lax and Friedrich is employed to patch the advective fluxes across elements. Polynomial order is set at $P=3$, translating into fourth-order spatial accuracy. The ES-DGSEM approach is constructed using the method of Hennemann \emph{et al.}\cite{hennemann2021provably}. The mesh refinement is accomplished using the Hennemann indicator\cite{hennemann2021provably} applied on $\rho\times p$ and three levels of refinement with a maximum threshold of 0.1. The AMR is activated every 5 time steps. A positivity preservation limiter is also used for density and pressure. For time stepping, the five-stage fourth-order Carpenter and Kennedy Runge-Kutta approach \cite{CK94a} is used. 

Our goal is to map the semi-ellipse's geometry into the solution of the Euler equations, see an example in Fig.~\ref{fig:field}. We divide the 36 samples into 28 training and 8 test sample sets, see appendix~\ref{append:datasets}. Splitting is done using a uniform random distribution sampling. Figure~\ref{fig:cases} depicts the geometries of the semi-ellipse employed for training and test datasets. The solution of the Euler system is computed on an unstructured grid, as shown in Fig.~\ref{fig:mesh}. Also, the number of grid points varies between samples. We use irregular and regular grid spacing to study the effect of grid point spacing on solution mapping. In the following section, we describe the pre-processing steps applied to the raw data obtained by the Trixi.jl solver to prepare the dataset for operator learning.

\begin{figure}[!t]
\begin{center}
\includegraphics[width=0.5\textwidth]{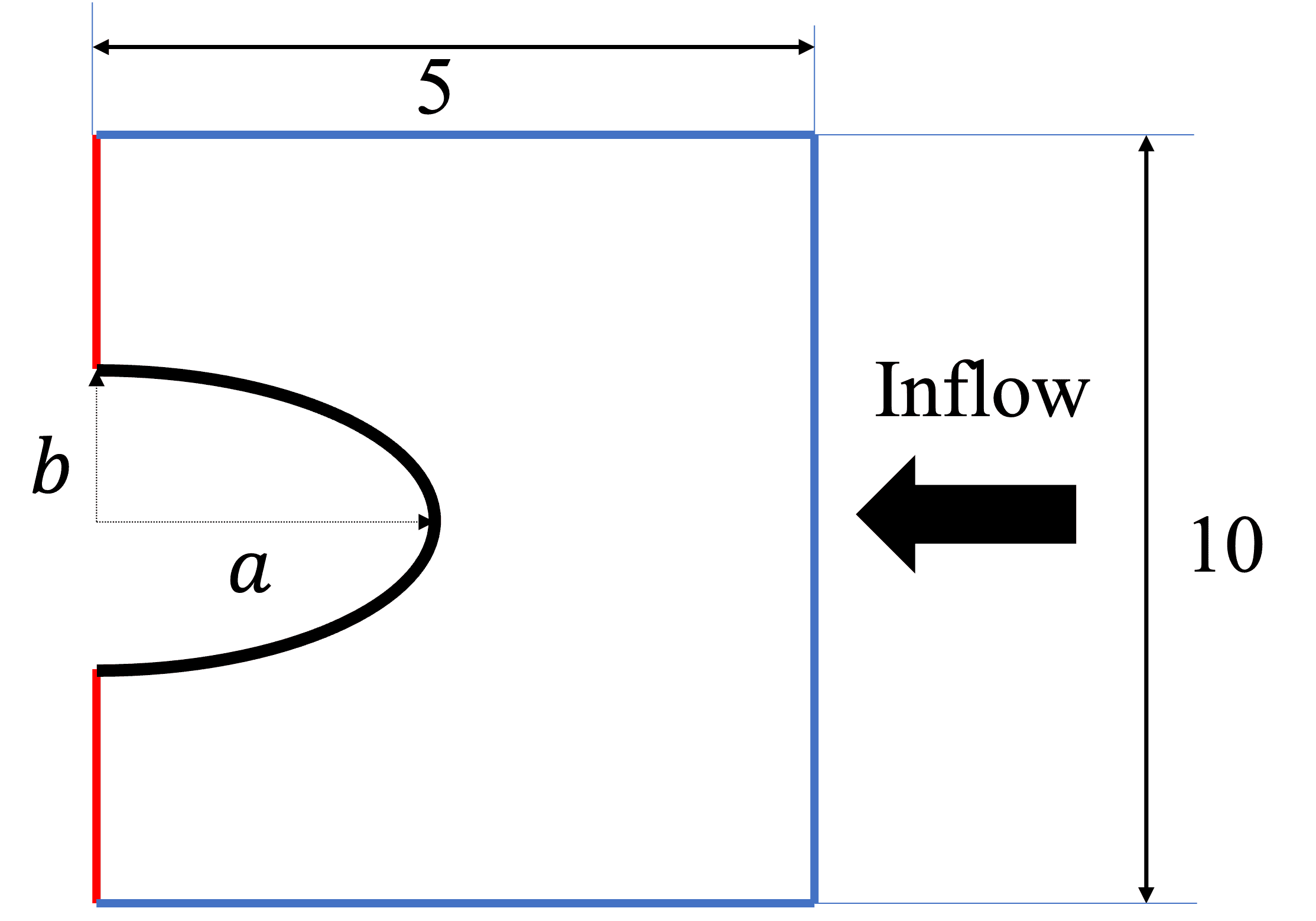}
\caption{Schematic of the  domain used for solving compressible Euler equations. The blue lines are defined as supersonic inflow boundary condition. The red lines are defined as extrapolation boundary. The black line is a semi-ellipse where the wall-slip boundary conditions are defined.}
\label{fig:domain}
\end{center}
\captionsetup{justification=centering}
\end{figure}

\begin{figure}[!t]
\begin{center}
\includegraphics[width=0.95\textwidth]{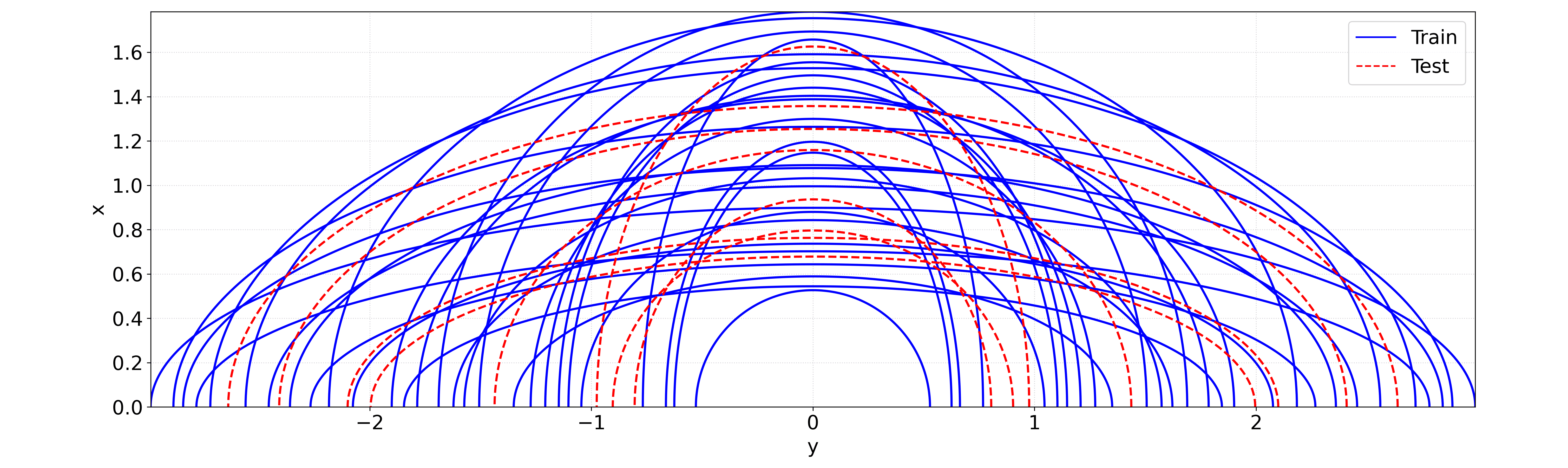}
\caption{Geometry of all the semi-ellipses used for training and testing. We have used 28 for training and 8 for testing. The blue lines refer to training samples and the dashed red lines refer to testing samples.}
\label{fig:cases}
\end{center}
\captionsetup{justification=centering}
\end{figure}

\subsubsection{Uniform Cartesian grid}
In this study, we evaluate the parameter-conditioned U-Net and FNO neural operators to learn the mapping between geometry parameters and flow fields. Therefore, we used linear interpolation to map the solution from an irregular grid into a structured Cartesian grid that covers the entire domain shown in Fig.~\ref{fig:domain}, including the area inside the semi-ellipse. In the y and x directions, we employed 256 discretization points and generated a total of $256^2$ uniformly spaced grid points. We generate a mask matrix for each case with zero or one entry as outlined in \cite{synergistic_deeponet}. Zero value is assigned to coordinate points inside the semi-ellipse, and one is assigned to points outside the semi-ellipse and inside the rectangular domain. Finally, all the training and testing dataset samples are projected onto the same Cartesian grid.

\subsubsection{Unstructured irregular Grid}
We utilize neural operators like DeepONet and MeshGraphNet, which can directly work with the irregular grids generated by the Trixi.jl solver. These grids adapt dynamically to the solution of the Euler equation, resulting in varying grid point coordinates and number of points for each case. For the DeepONet network, maintaining a consistent input shape for the trunk network is necessary when using a batch size larger than one. To address this, we standardize the number of grid points to match the maximum number of points in the training dataset. If a sample has fewer points, we pad it by repeating its last coordinate to ensure uniformly sized array for operator network training and evaluation.

\begin{figure}[t!]
  \begin{center}
    \begin{tabular}{ccc}

\includegraphics[width=0.28\textwidth,height=0.5\textwidth]{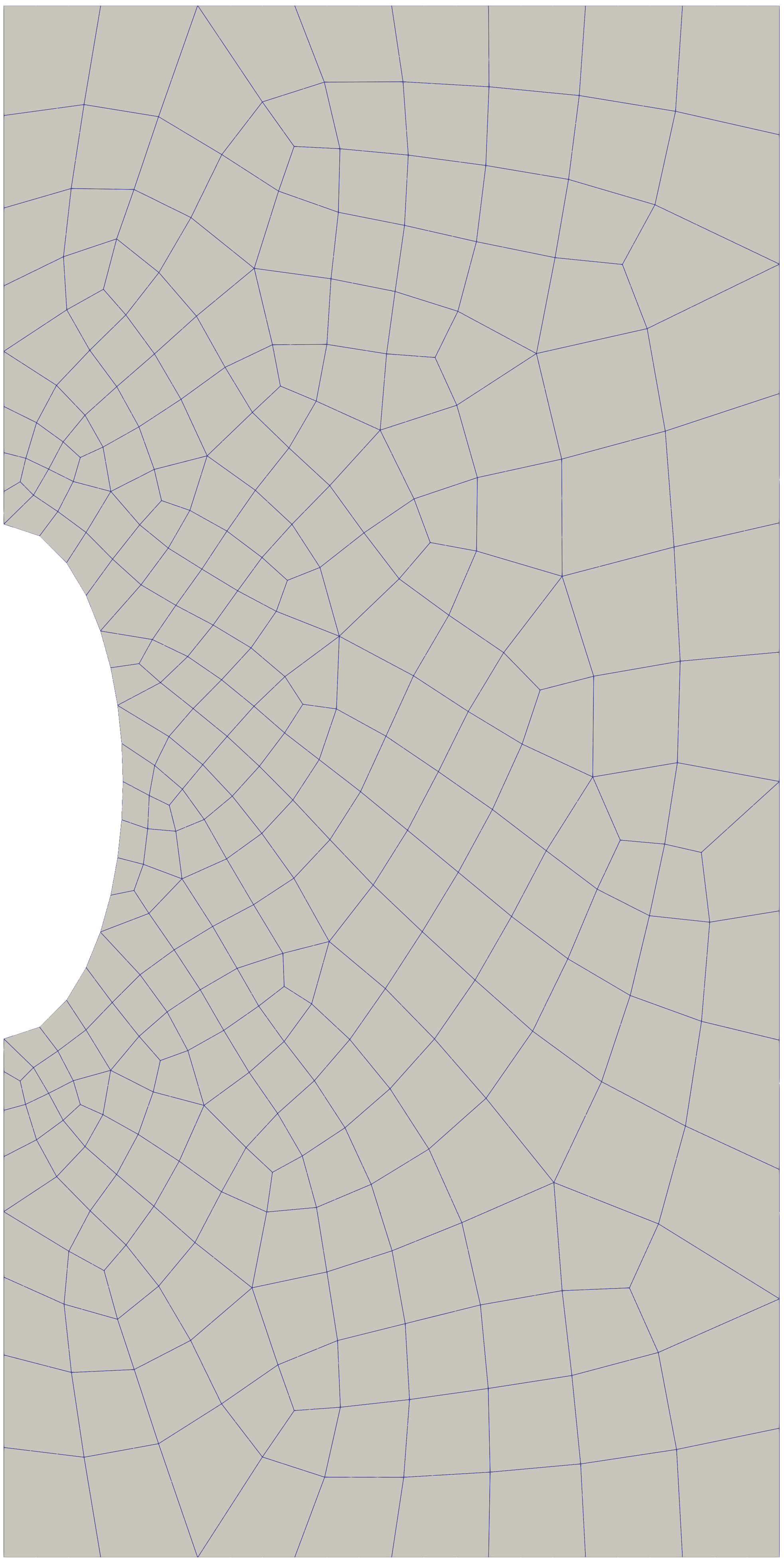}
 &
\includegraphics[width=0.28\textwidth,height=0.5\textwidth]{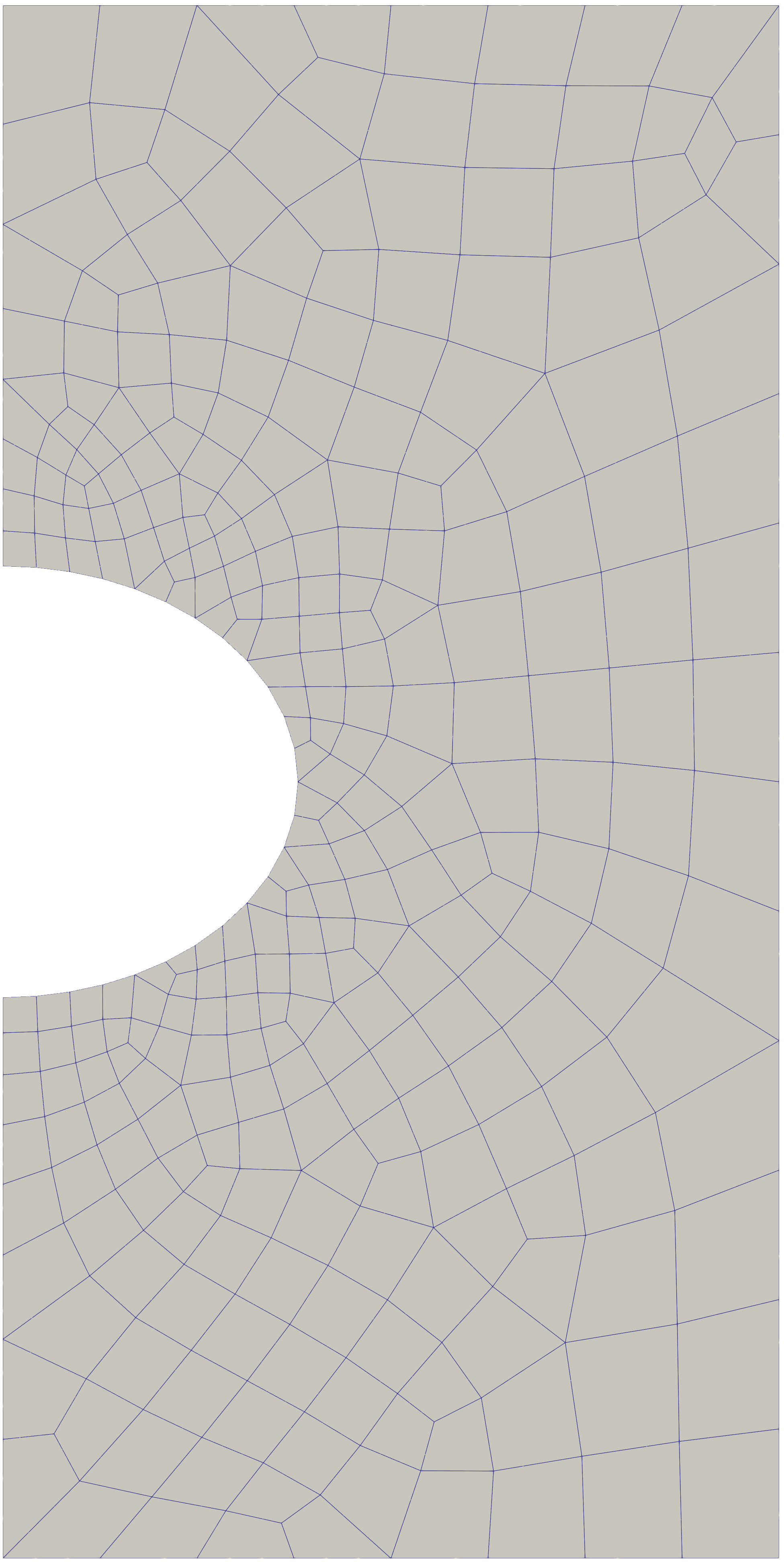}

 &
\includegraphics[width=0.28\textwidth,height=0.5\textwidth]{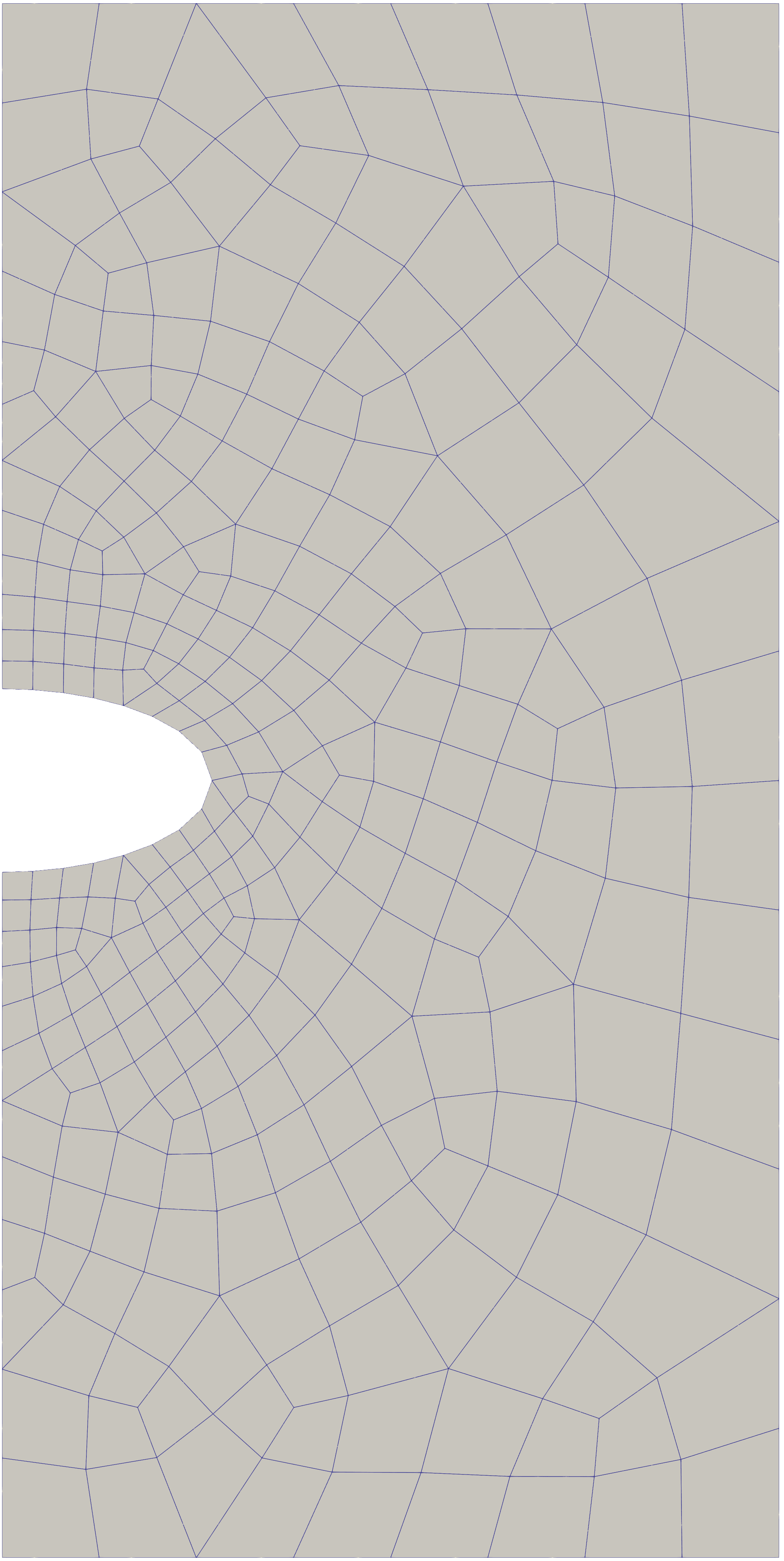}
  \\
  (a) Mesh 1 at $t=0$& (b) Mesh 3 at $t=0$ & (b) Mesh 5 at $t=0$
\\
\includegraphics[width=0.28\textwidth,height=0.5\textwidth]{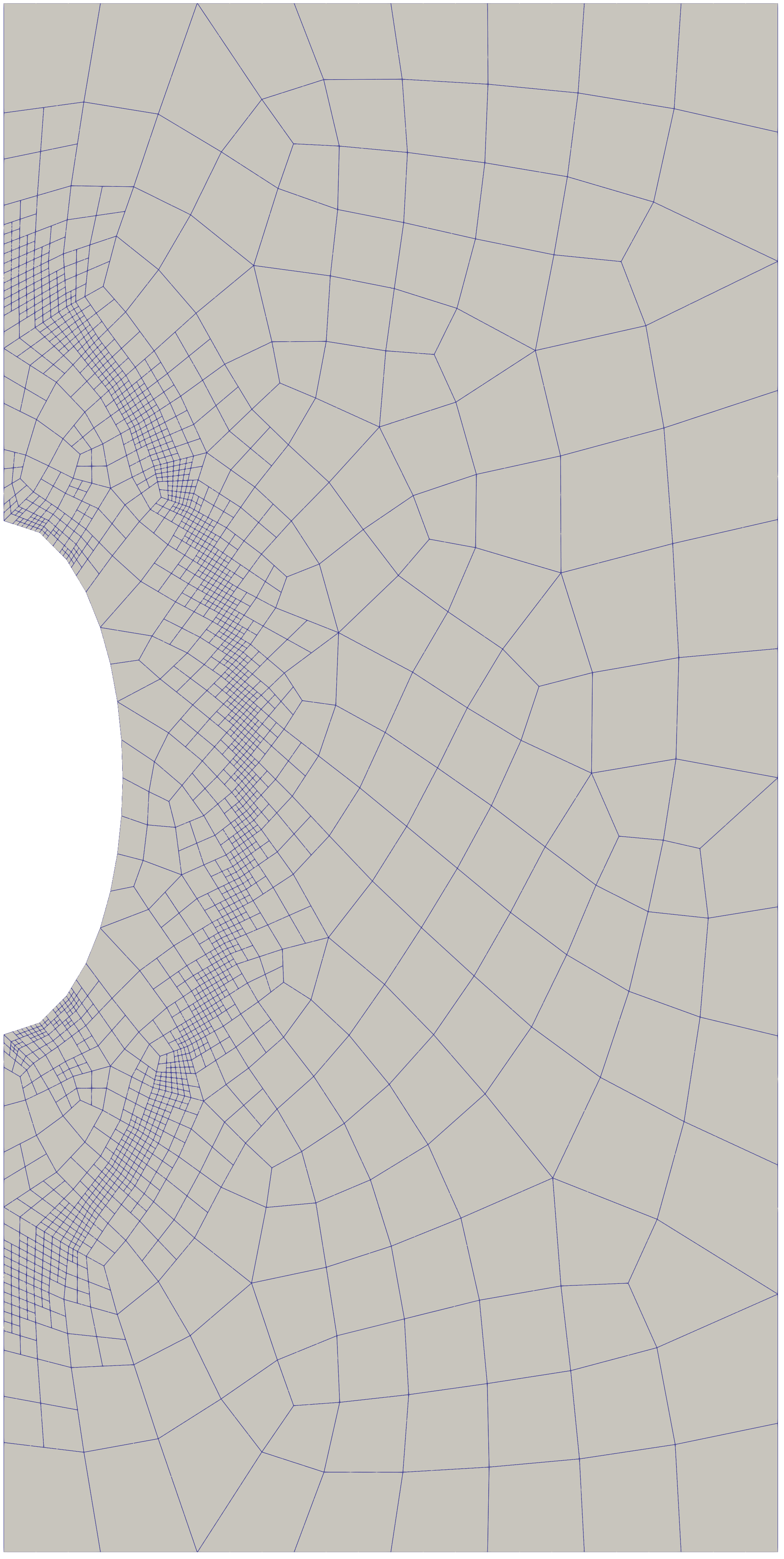}
 &
\includegraphics[width=0.28\textwidth,height=0.5\textwidth]{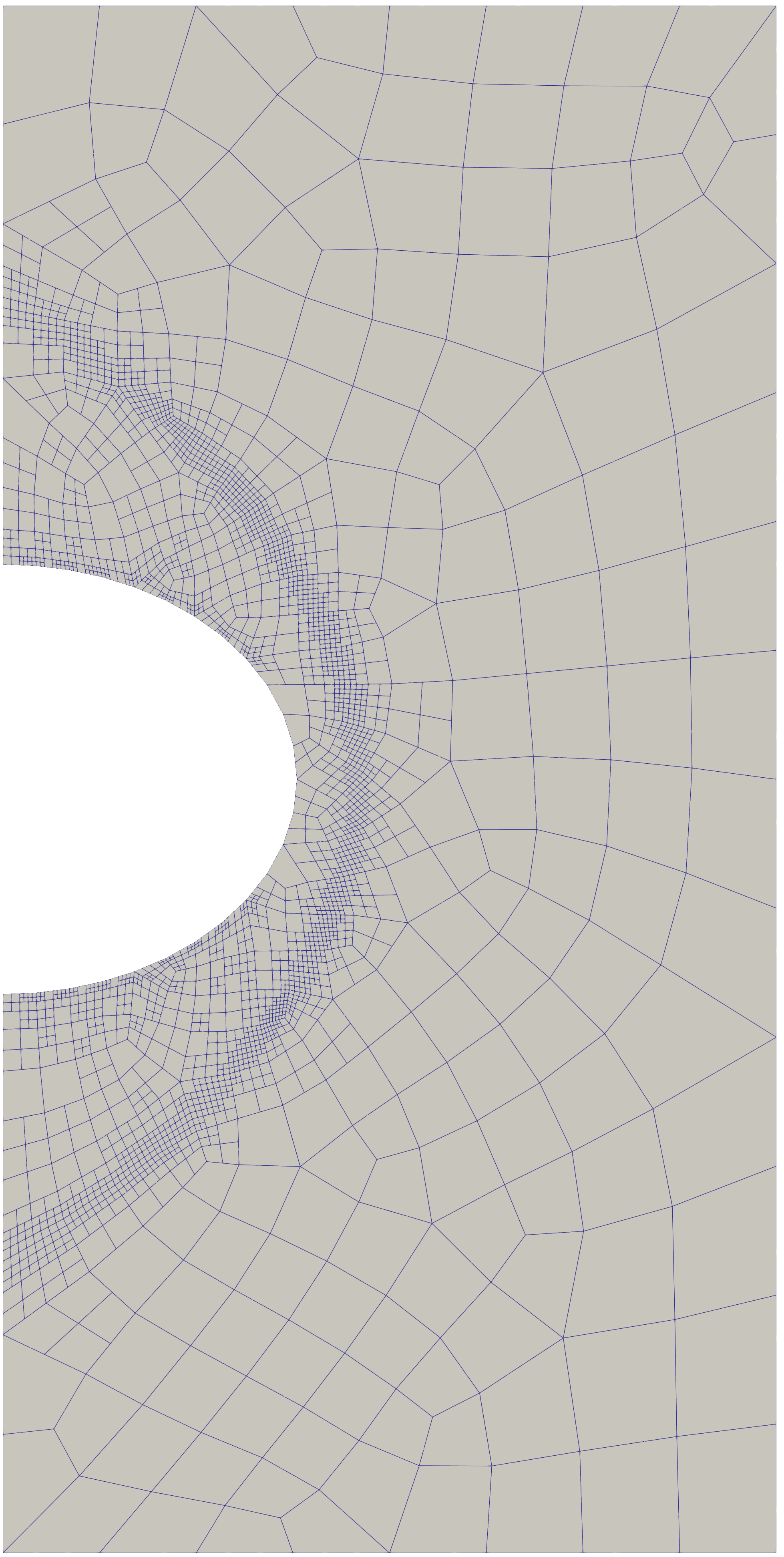}

 &
\includegraphics[width=0.28\textwidth,height=0.5\textwidth]{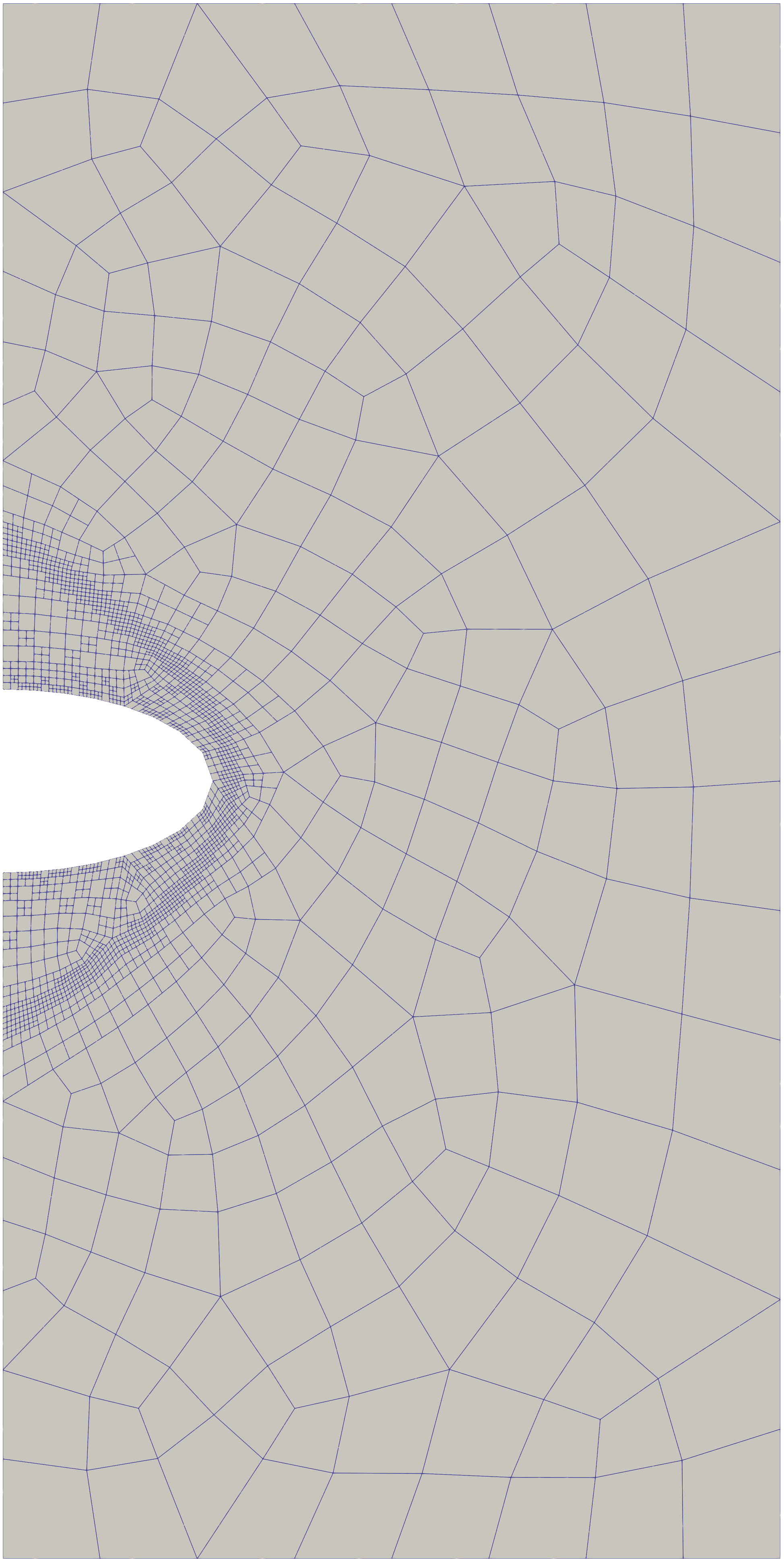}
  \\
  (d) Mesh 1 at $t=1$& (e) Mesh 3 at $t=1$ & (f) Mesh 5 at $t=1$
\end{tabular} 
\caption{Initial mesh configuration and adaptive mesh at final time. (a) Mesh case 1 at $t=0$, (b) Mesh case 3 at $t=0$, (c) Mesh case 5 at $t=0$, (d) Mesh case 1 at $t=1$, (e) Mesh case 3 at $t=1$, (f) Mesh case 5 at $t=1$.}
    \label{fig:mesh}
  \end{center}
\end{figure}

\begin{figure}[t!]
  \begin{center}
    \begin{tabular}{cccc}

\includegraphics[width=0.24\textwidth,height=0.42\textwidth]{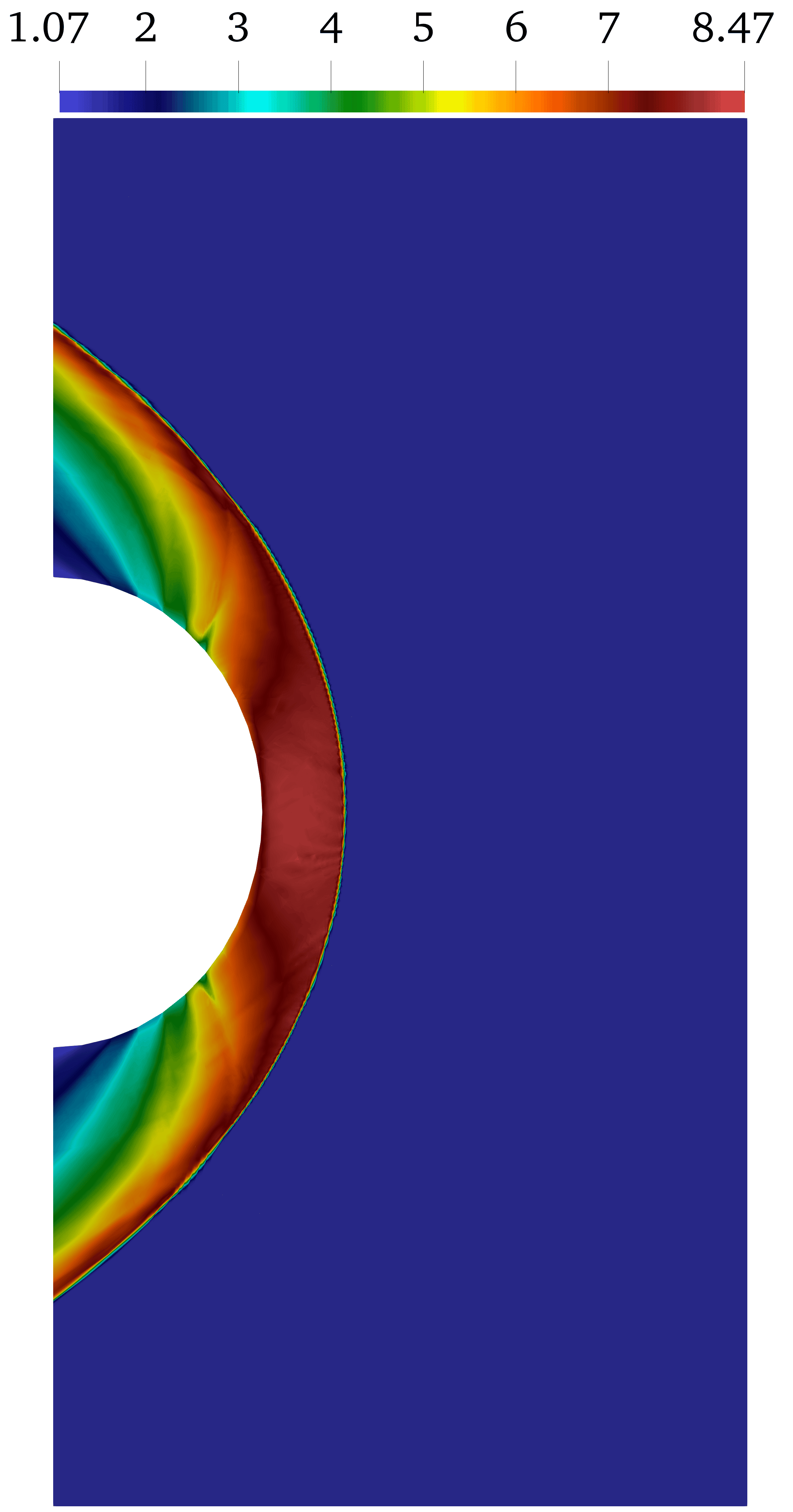}
 &
\includegraphics[width=0.24\textwidth,height=0.42\textwidth]{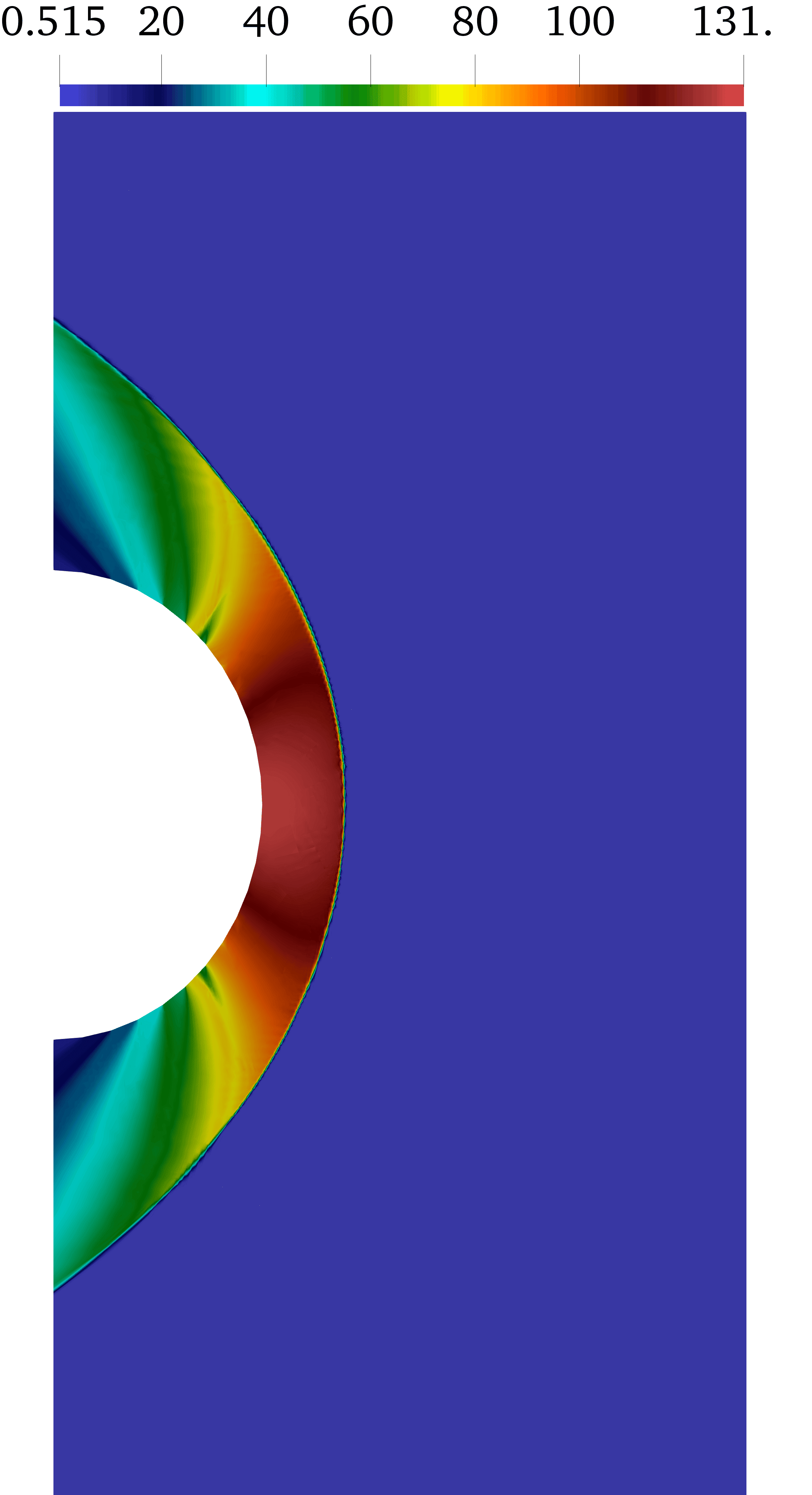}&\includegraphics[width=0.24\textwidth,height=0.42\textwidth]{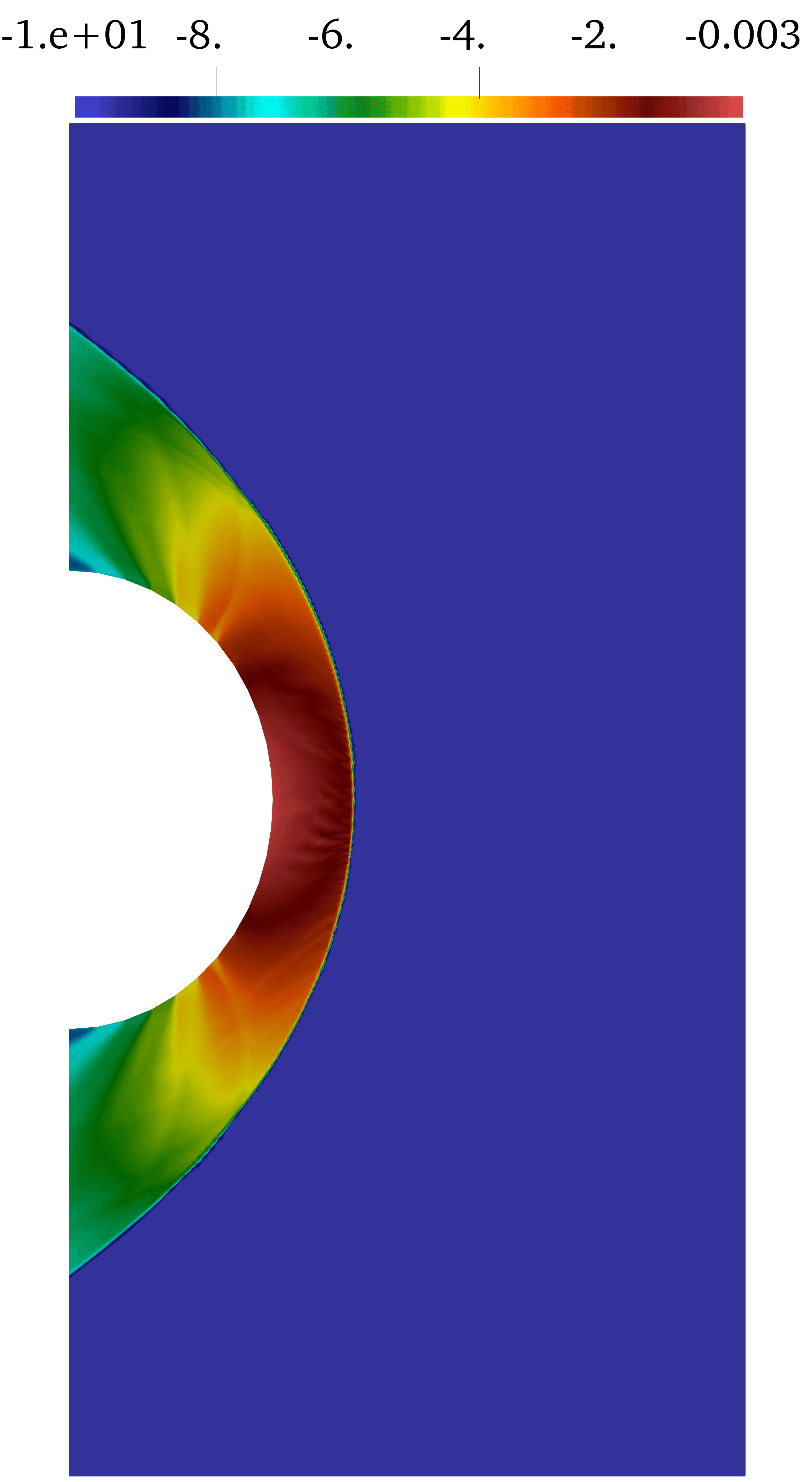}
 &
\includegraphics[width=0.24\textwidth,height=0.42\textwidth]{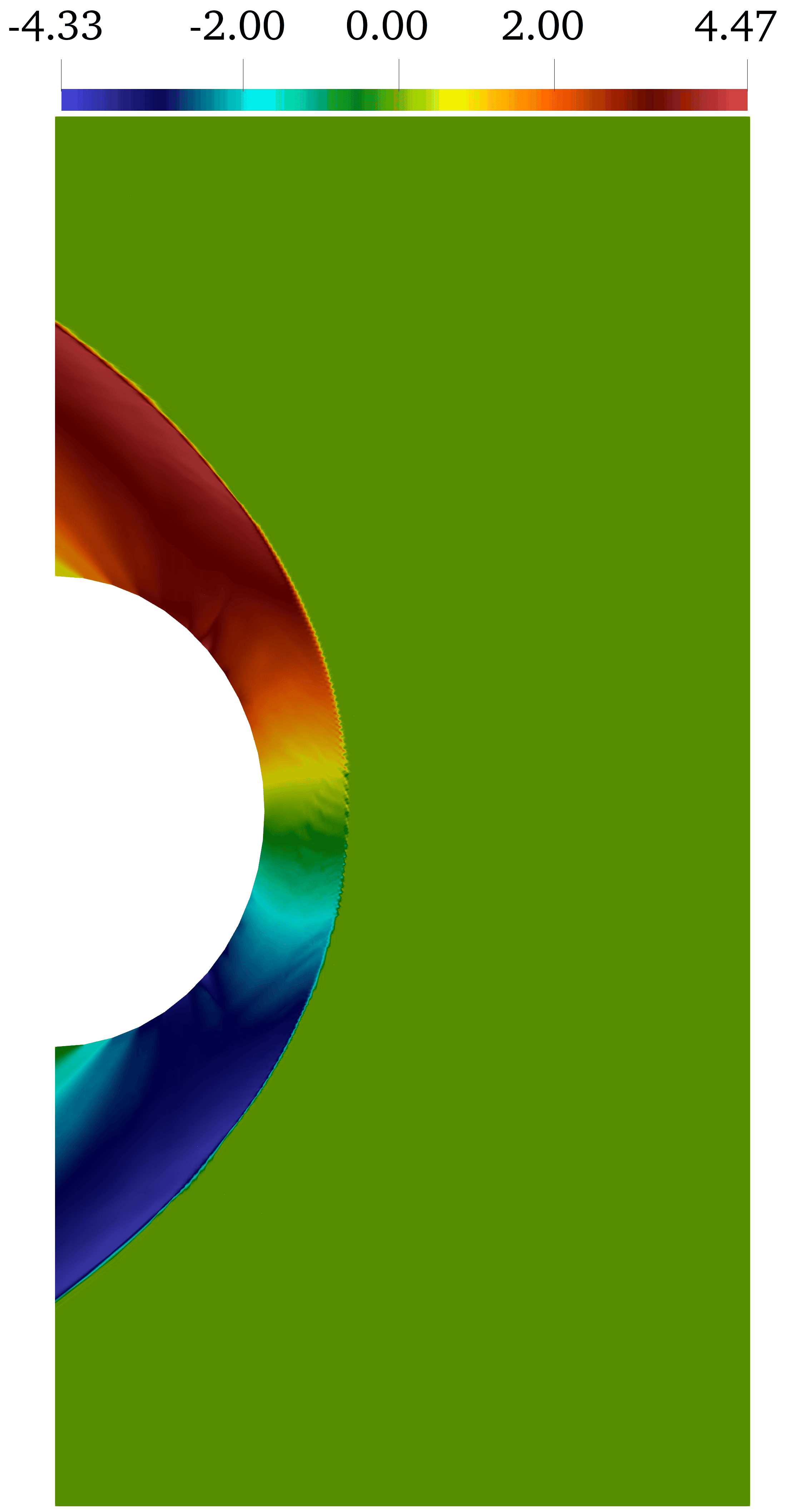}
  \\
  (a) Density& (b) Pressure  &(c) x-Velocity& (d) y-Velocity  
  
\end{tabular} 
\caption{Solution fields for case 1 at $t=1$. (a) Density contours, (b) Pressure contours, (c) x-Velocity, (d) y-Velocity.}
    \label{fig:field}
  \end{center}
\end{figure}

\subsection{Hypersonic viscous flow around Reentry capsule}
We design a challenging problem for which geometric and flow parameters are considered simultaneously. A $\beta$-apex blunt cone geometry\cite{sahoo2007experiments} is parameterized to generate a flow field surrogate. The schematic for the physical domain is shown in Fig.~\ref{fig:domain_capsule}(a). A supersonic inflow boundary condition is imposed at the blue boundaries, and an extrapolation boundary condition is imposed on the red boundary. The surface of the capsule is selected as an isothermal wall boundary condition with $T_w=25$ $^oC$. According to a magnified view of the capsule in Fig.~\ref{fig:domain_capsule}(a), the radius ($R$) of the blunt section and the $\alpha$ angle are designated as geometric parameters that vary such that $\alpha \in [40^o, 60^o]$ and $R \in [13.06, 20]$. We then added Mach number of the free stream as an extra input parameter defined as  $M\in [5, 8]$. The free stream conditions are set as $p_\infty=850$ Pa, $\rho_\infty=0.0141$ $\textrm{kg}/\textrm{m}^3$, $Re_\infty=4\times 10^4$ and $Pr=0.72$. The reference length is set as $L_f=1$ m. 

We employ the Latin hypercube sampling \cite{mckay2000comparison} method to randomly draw samples inside the three-dimensional parametric space constructed by $(R,\alpha, M)$. We selected 60 samples with various geometries shown in Fig.~\ref{fig:caspule} and free-stream Mach numbers. For each of the 60 samples, we automatically generated the geometry of the physical domain using GMSH and generated a structured mesh with only quadrilateral elements. We discretize the physical domain using 8,501 elements in total, see Fig.~\ref{fig:domain_capsule}(b). We then employ the Trixi.jl solver to solve Eq.~\ref{system_equation} for twelve flow-through times to allow the flow to reach steady-state conditions. Each flow-through time is defined as the non-dimensional time that the free-stream flow starts from the inflow boundary and exits from the outflow boundary, and can be computed as $14H/M$. At the end of the simulations, we select the last snapshot of the time-dependent solution as the steady-state flow field that the machine learning model will learn. After completing the simulations, we create a geometry and Mach-dependent flow field dataset. Since Trixi.jl employs the DGSEM scheme, it uses nodal solution points on the sides of quadrilateral elements. Therefore, duplicate solution points exist on the elements' internal edges. We eliminate all duplicate points before training the machine learning model. From 60 samples, we randomly select 48 samples for training and 12 for testing datasets, see appendix~\ref{append:datasets} for dataset details.


\begin{figure}[t!]
  \begin{center}
    \begin{tabular}{cc}

\includegraphics[width=0.5\textwidth,height=0.45\textwidth]{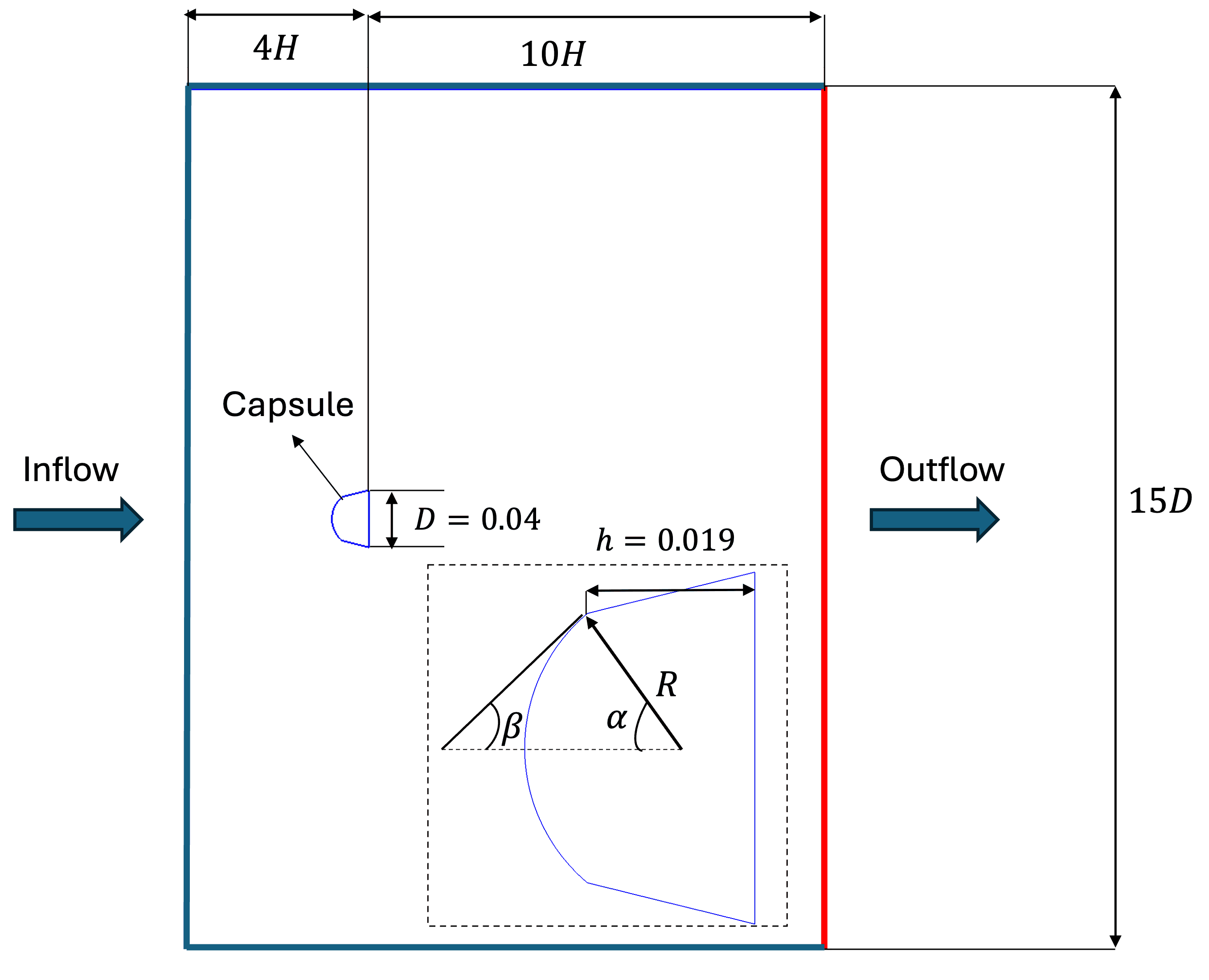}
 &
\includegraphics[width=0.35\textwidth,height=0.45\textwidth]{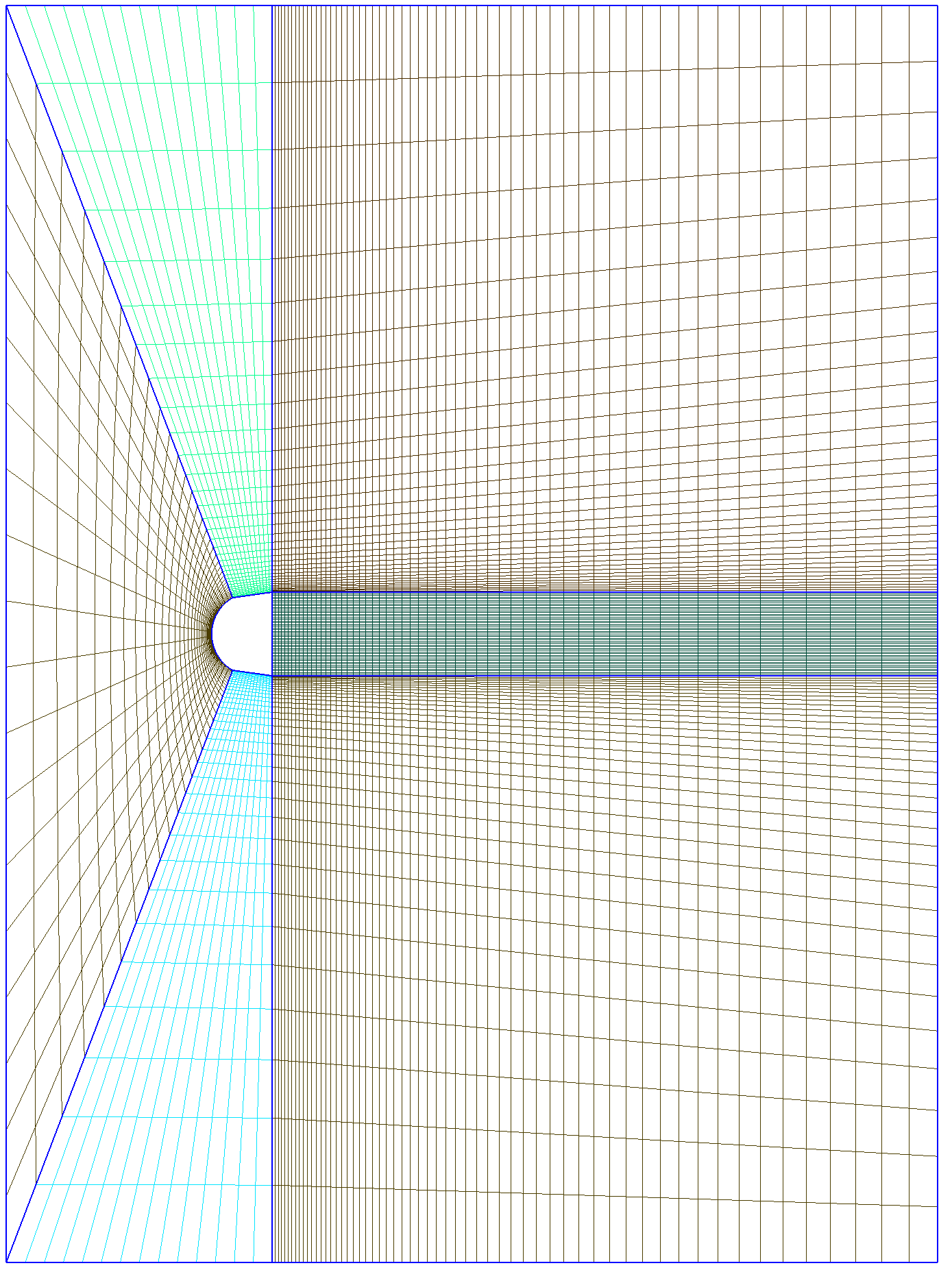}
  \\
  (a) Physical domain schematic& (b) Sample grid  
\end{tabular} 
\caption{(a)Schematic of the domain used for solving compressible Navier-Stokes equations. A capsule geometry is place inside a rectangle. Supersonic inflow boundary conditions are imposed at the blue boundaries. Supersonic outflow is also imposed at the red boundary. Isothermal wall boundary condition is imposed on the capsule surfaces. A zoomed-in view of the capsule is shown inside the dashed outline box. The geometry of the capsule is parameterized using $\alpha \in [40, 60]$ and $R\in [13.06, 20]$ and the flow condition is parametrized using Mach number as $M\in[5,8]$. All the dimensions are in meters and $H=0.03177$.(b) Domain discretization using quadrilateral elements}
    \label{fig:domain_capsule}
  \end{center}
\end{figure}

\begin{figure}[!t]
\begin{center}
\includegraphics[width=0.6\textwidth]{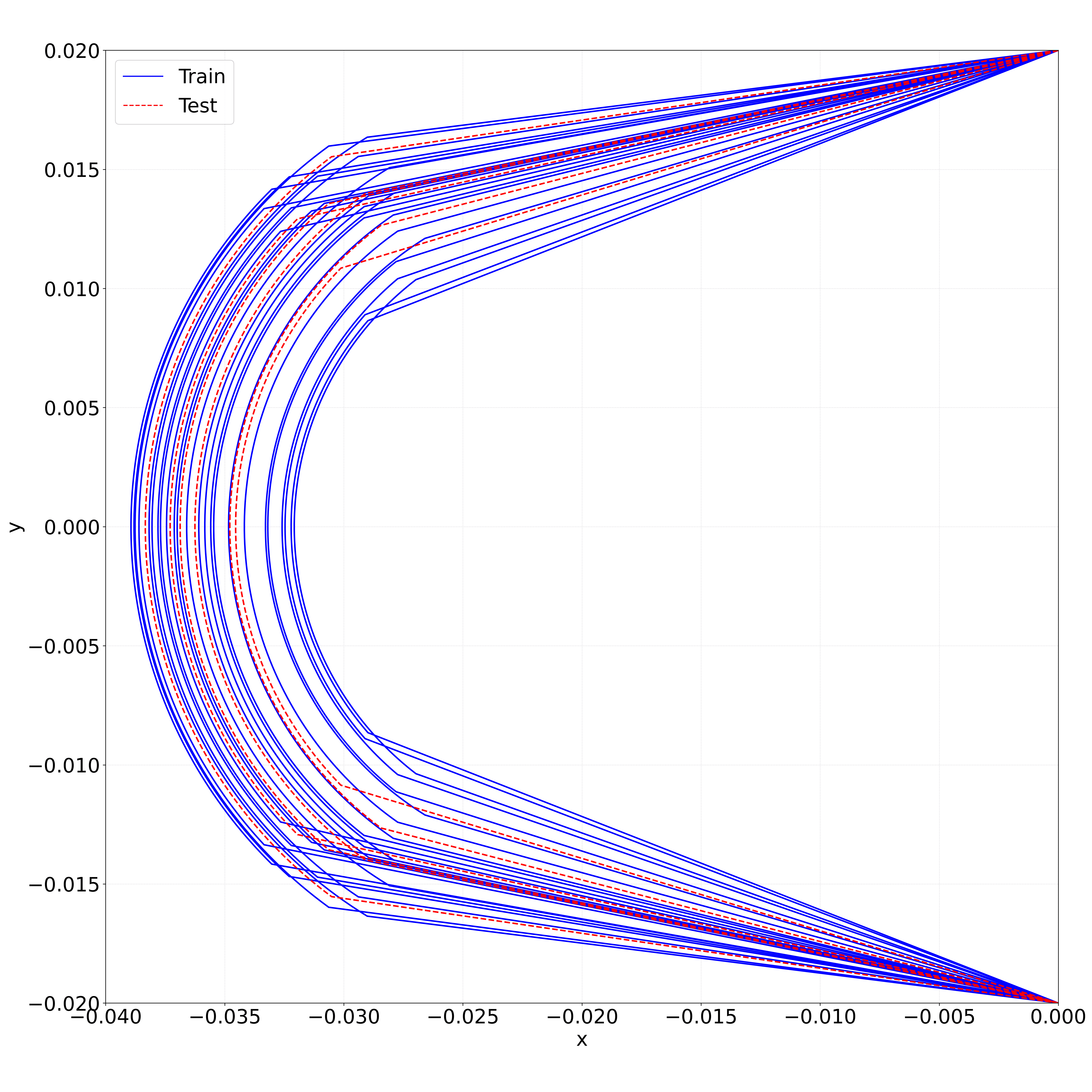}
\caption{Geometry of all capsules used for operator network learning. We use 48 samples for training and 12 for testing, see appendix~\ref{append:datasets}. The blue lines refer to training samples and the dashed red lines refer to testing samples.}
\label{fig:caspule}
\end{center}
\captionsetup{justification=centering}
\end{figure}

\subsection{Viscous flow inside convergent-divergent nozzle}
The last example includes learning the viscous flow field for an internal flow problem. The physical domain consists of a converging-diverging nozzle attached to a duct as shown in Fig.~\ref{fig:domain_cd}. The compressible flow enters the nozzle at Mach 0.6 and a Reynolds number of 1000. Although the inflow Mach number is set at 0.6, the flow transforms to a supersonic regime in the diverging part of the nozzle. Inflow conditions are defined as $p_\infty=89214.769$ Pa and $\rho_\infty=1.074\textrm{kg}/\textrm{m}^3$ and gas constant $R=287.0$ JKg/K. The $Pr=0.72$ is selected, and the isothermal wall temperature is set at $T_w=2T_\infty$. At the inflow boundary, we impose subsonic inflow boundary condition \cite{jacobs2003numerical}, and at the outflow boundary, the subsonic outflow \cite{jacobs2003numerical}. The physical domain is discretized using quadrilateral elements with a polynomial order of 3 for the DGSEM solver. We initialize the entire domain using the inflow conditions and simulate for 20 flow-through times to reach a steady state solution.

Our objective is to generate a surrogate model that takes three geometric parameters and predicts flow field variables such as temperature, density, pressure, and velocity. The geometric parameters consist of inlet height $2 h_i$, outlet height $2 h_o$, and the x-coordinate of the location of the throat area of the converging-diverging nozzle ($10x_t$). The three parameters vary such that $h_i \in [3, 4]$, $h_o \in [1.5, 2.5]$, and $x_t \in [0.45, 0.65]$. By changing the geometric parameters, the overall shape of the nozzle is constructed using a fifth-order polynomial. The top wall of the nozzle is constructed by a fifth-order polynomial constrained by the following geometric conditions: 
\begin{equation}
\begin{split}
y = a_5x^5+a_4x^4+a_3x^3+a_2x^2+a_1x+a_0, \\
y(x=0)=\frac{h_i}{2},\quad y(x=L_N)=\frac{h_o}{2},\quad y(x=x_t)=1\\
\frac{dy}{dx}\Big|_{x=0}=0,\quad\frac{dy}{dx}\Big|_{x=L_N}=0,\quad \frac{dy}{dx}\Big|_{x=x_t}=0.
\end{split}
\end{equation}
The bottom wall is the mirror image of the top wall. We draw 60 samples using Latin hypercube sampling from the three-dimensional parametric space and generate the geometry for all the cases using GMSH. We then discretize the domain using quadrilateral elements that are clustered close to the isothermal walls using 1824 elements. From the 60 generated cases, we randomly select $70\%$ (i.e., 42) trajectories for training and 18 for testing datasets, see Fig.~\ref{fig:cd_cases} and appendix~\ref{append:datasets} for dataset details. Next, we compute the steady-state solution using Trixi.jl and extract the solution field including $p$, $\rho$, $u$, $v$, and $T$ from the output files. We remove all the duplicate points in the domain since the spectral element method uses nodes on the common edges of the computational elements. Ultimately, we accumulate the solution, coordinates, and geometric parameters into three separate tensors to feed into the training loop.


\begin{figure}[t!]
  \begin{center}
    \begin{tabular}{c}

\includegraphics[width=0.8\textwidth,height=0.4\textwidth]{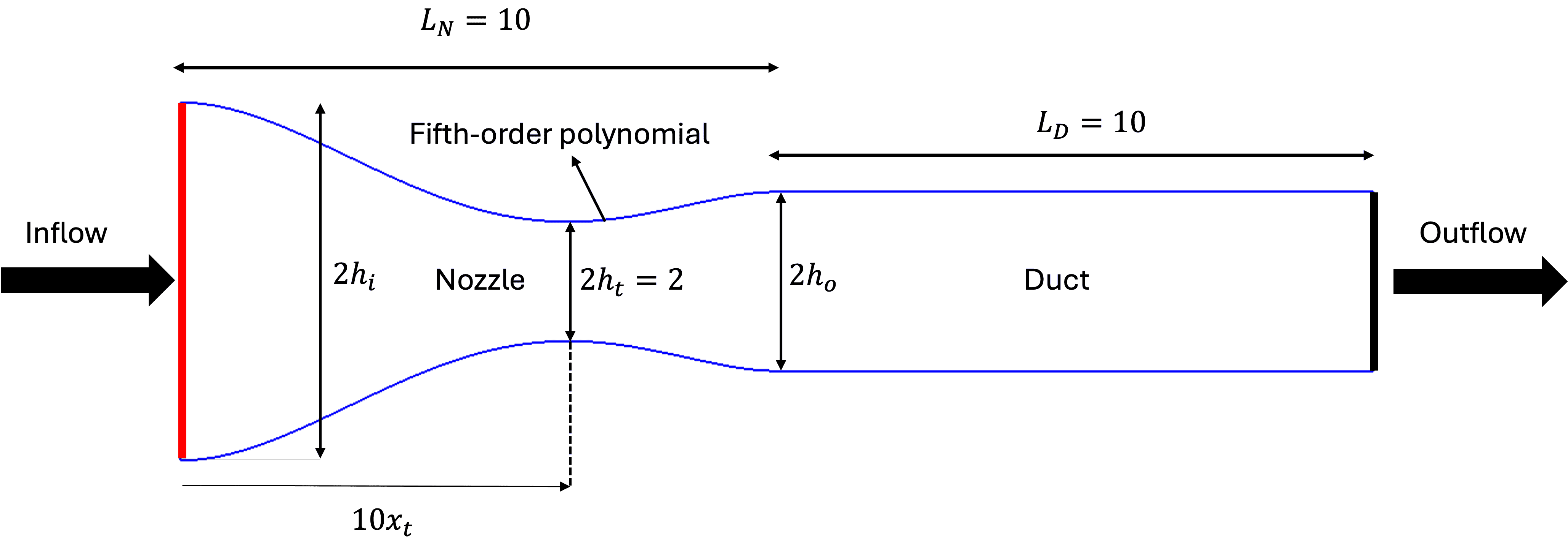}
  \\
  (a) Physical domain schematic
  \\
\includegraphics[width=0.7\textwidth,height=0.3\textwidth]{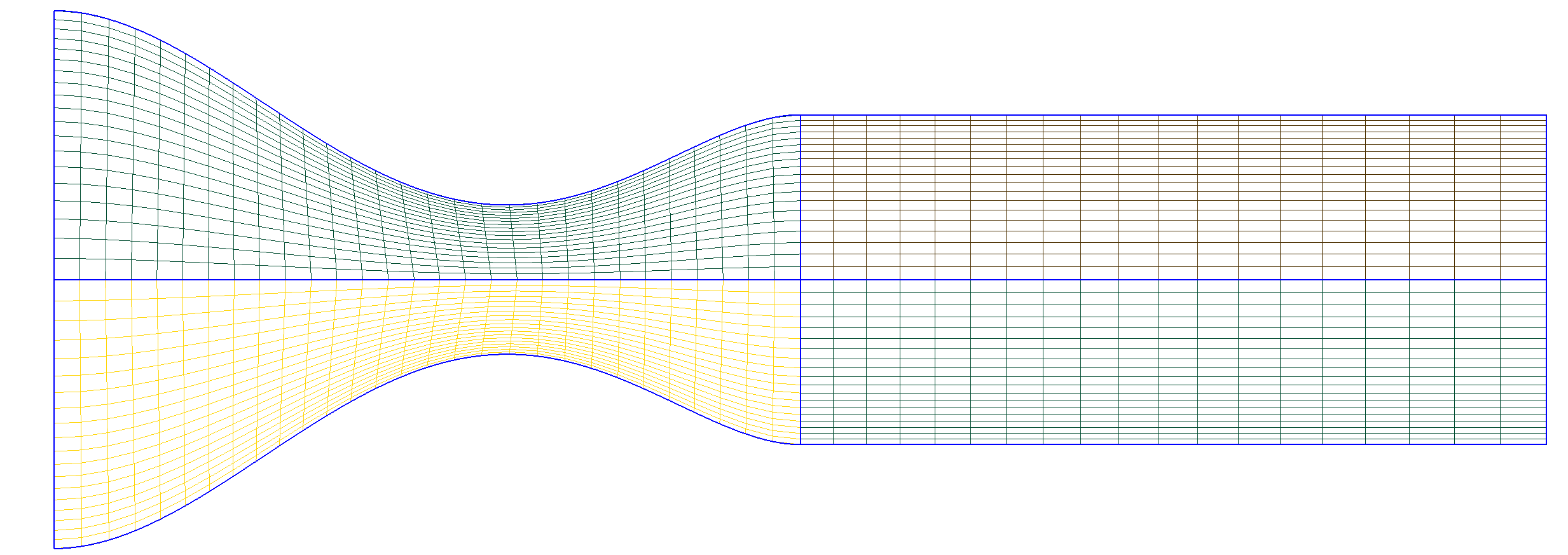}
\\ (b) Sample grid  
\end{tabular} 
\caption{(a) Schematic of the domain used for solving compressible Navier-Stokes equations. The domain consists of a 
converging-diverging nozzle and a duct. The length of the nozzle is 10 and the length of the duct is 10. The throat diameter is set at 2. The inflow Mach number is 0.6 and the Reynolds number is 1000 based on a reference length scale of one. The wall boundary condition is isothermal and $T_w=2T_\infty$. The walls of the nozzle are represented using a fifth-order polynomial by imposing sixth geometric constraints. For this problem, we use nozzle inlet diameter $h_i$, nozzle outlet diameter $h_o$ and the location of the throat $x_t$ as the geometric parameters. (b) A sample discretization of the physical domain.}
    \label{fig:domain_cd}
  \end{center}
\end{figure}

\begin{figure}[!t]
\begin{center}
\includegraphics[width=0.95\textwidth]{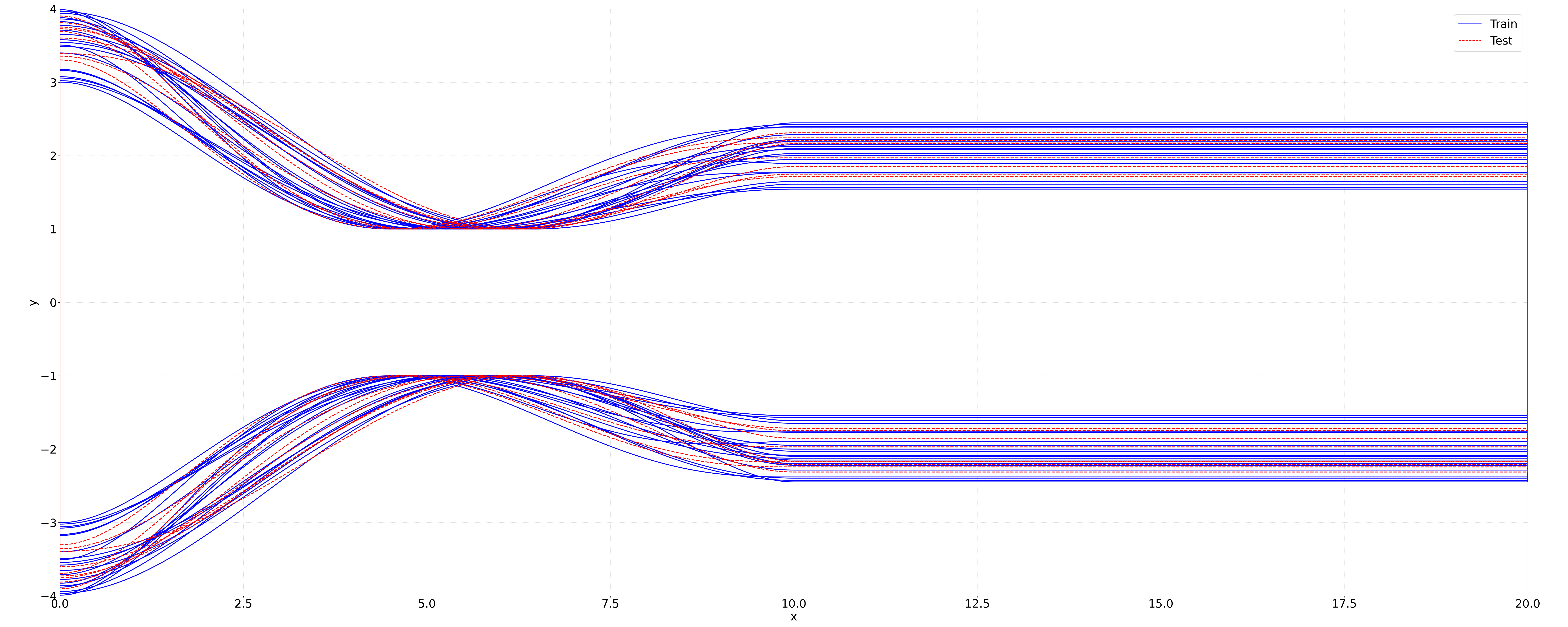}
\caption{Geometry of all the converging diverging nozzles used for operator network learning. We use 42 samples for training and 18 for testing, see appendix~\ref{append:datasets} for dataset details. The blue lines refer to training samples and the dashed red lines refer to testing samples.}
\label{fig:cd_cases}
\end{center}
\captionsetup{justification=centering}
\end{figure}
\section{Methodology}
Here, we describe the architecture of Fusion-DeepONet model and how to improve its ability to predict not only the solution field but also the derivative of the solution field, specifically close to the boundary of the physical domain. Our quantities of interest in hypersonic flows are usually shear stress, pressure, and heat flux acting on the surfaces of the flying object. By predicting the quantities of interest on the solid surfaces, we can optimize the geometry to reduce drag force, increase lift force, and minimize heat flux going into the body. Therefore, we aim to develop a method to accurately predict the flow field and its derivatives on the physical boundary of the varying physical domain.  

\subsection{Fusion-DeepONet}

We present Fusion-DeepONet, a multi-scale conditioned neural field. Conditional neural fields were first conceptualized by Wang \emph{et al.} \cite{wang2024cvit}, who introduced the idea of global and local conditioning in neural fields. DeepONet, in this context, can be characterized as a globally conditioned neural field. In DeepONet, the trunk network serves as the neural field, with its final linear layer being conditioned by the output of the branch network. Conditioning the neural field at its final layer for all coordinate points defines global conditioning. Additionally, Seidman \emph{et al.} \cite{seidman2022nomad} proposed the NoMaD architecture, which conditions the neural field locally (for each coordinate point) by concatenating the input function encoding with the input encoding of the neural field. In Fusion-DeepONet, we introduce a new type of conditioning that conditions each trunk network hidden layer using the latent encoding of branch hidden layers while keeping the global conditioning of the trunk network output. In the following, we explain the architecture of the Fusion-DeepONet.

Fusion-DeepONet, like the Vanilla-DeepONet, is built from two sub-networks: a branch network that processes the input function and a trunk network that processes the coordinates. We begin by defining the architectures of each sub-network. For $L\in\mathbb{N}$ and the vector $\vec n^b = (n_0^b, n_1^b, \dots, n_L^b*n_v) \in \mathbb{N}^{L+1}$ with $n_v$ denoting number of output variables and $n_l^b=n$ for $l=1,\cdots,L-1$ denoting the width of layer $l^\textrm{th}$, an $L-1$-layer feed-forward branch network is given recursively by
\begin{equation}
\begin{aligned}
\mathbf Z^0 &= \mathbf x^b \\
\mathbf Z^l &= \phi\bigl(\mathbf W^{b,l} \mathbf Z^{l-1} + \mathbf b^{b,l}, \,\mathbf a^{b,l}\bigr),
\quad l =1,\dots,L-1,
\end{aligned}
\label{eq:branch_architecture}
\end{equation}
where $\phi(\cdot,\mathbf a^{b,l})$ is an adaptive activation with learnable parameters $\mathbf a^{b,l} \in \mathbb{R}^{N_R}$ with $N_R$ denoting number of Rowdy activation function learnable coefficients. In Eq.~\eqref{eq:branch_architecture}, $\mathbf x^b \in \mathbb{R}^{N\times N_p}$ denotes the vector of branch inputs, i.e., geometric or flow parameters, with $N$ training samples and $N_p=n_0$ number of input parameters. The weight matrix is defined as $\mathbf W^{b,l}\in \mathbb R^{n_{l-1}^b\times n_l^b}$ for $l=1,\cdots, L-1$. The bias matrix is also defined as $\mathbf b^{b,l} \in \mathbb R^{1\times n_l^b}$. The matrix $\mathbf Z^l \in \mathbb R^{N \times n_l^b}$ denotes the output of branch network layers for $l=1, \cdots, L$. Similarly, for $\vec n^t = (n_0^t, n_1^t, \dots, n_L^t) \in \mathbb{N}^{L+1}$ with $n_l^t=n$ for $l=1,\cdots,L-1$, the trunk network is constructed as 
\begin{equation}
\begin{aligned}
\mathbf Y^0 &= \mathbf x^t,\\
\mathbf Y^l &= \phi\bigl(\mathbf W^{t,l} \mathbf Y^{l-1} + \mathbf b^{t,l}, \,\mathbf a^{t,l}\bigr),
\quad l =1,\dots,L-1,
\end{aligned}
\label{eq:trunk_architecture}
\end{equation}
where $\phi(\cdot,\mathbf a^{t,l})$ is an adaptive activation with learnable parameters $\mathbf a^{t,l} \in \mathbb{R}^{N_R}$ with $N_R$ denoting number of Rowdy activation function learnable coefficients. In Eq.~\eqref{eq:trunk_architecture}, $\mathbf x^t \in \mathbb{R}^{N\times N_{pts}\times N_c}$ denotes the vector of trunk inputs, i.e., coordinate points for training samples, with $N$ training samples, $N_{pts}$ number of solution points for each sample, and $N_{c}=n_0^t$ number of coordinate directions. The weight matrix is defined as $\mathbf W^{t,l}\in \mathbb R^{n_{l-1}^t\times n_l^t}$ for $l=1,\cdots, L-1$. The bias matrix is also defined as $\mathbf b^{t,l} \in \mathbb R^{1\times n_l^t}$. The tensor $\mathbf Y^l \in \mathbb R^{N\times N_{pts} \times n_l^t}$ denotes the output of trunk network's layers for $l=1, \cdots, L-1$.

Next, we introduce connection coefficients that condition the trunk net layers.  Define
\begin{equation}
\begin{aligned}
\mathbf S^0 &= \mathbf Z^1,\\
\mathbf S^m &= \mathbf Z^{m+1} + \mathbf S^{m-1}, 
\quad m = 1,\dots,L-2,
\end{aligned}
\label{eq:skip_connections}
\end{equation}
where $\mathbf S^m \in \mathbb R^{N \times n_{m+1}^b}$. We want to condition the output of trunk layers $\mathbf Y^{l} \in \mathbb R^{N\times N_{pts}\times n_l^t}$ using $\mathbf S^m \in \mathbb R^{N \times n_{m+1}^b}$; however, the tensor dimensions does not match at the first glance and we previously assumed that number of neurons follows $n_{m+1}^b=n_l^t=n$. To be able to perform elment-wise multiplication, we add a single axis to $\mathbf S^m \in \mathbb R^{N \times n}$ and construct $\tilde{\mathbf S}^m \in \mathbb{R}^{N\times 1 \times n}$ to account for the extra axis of trunk output that accounts for coordinate points. Then conditioning proceeds by element-wise multiplication,
\begin{equation}
\widehat{\mathbf Y}^l \;=\; \mathbf Y^l \;\odot\; \tilde{\mathbf S}^{\,l-1}.
\quad l = 1,\dots,L-2.
\label{eq:conditioning}
\end{equation} 
This element-wise multiplication conditions the output of the trunk layer over the first axis that accounts for a number of samples ($N$) and the last axis that accounts for the number of neurons using unique coefficients while using common coefficients for all the coordinate points. In Eq.~\eqref{eq:conditioning}, $\widehat{\mathbf Y}^l \in \mathbb R^{N\times N_{pts} \times n_l^t}$ denotes the conditioned trunk layer output. 
Finally, following the DeepONet convention, the last layers are linear:
\begin{equation}
\begin{aligned}
\mathbf Z^L &= \mathbf W^{b,L} \mathbf Z^{L-1} + \mathbf b^{b,L},\\
\widehat{\mathbf Y}^L &= \mathbf W^{t,L} \widehat{\mathbf Y}^{L-1} + \mathbf b^{t,L}.
\end{aligned}
\label{eq:branchtrunk_output}
\end{equation}
The Fusion-DeepONet output is then the dot product of these two final vectors:
\[
\mathcal{G}(\mathbf x^b,\mathbf x^t) \;=\; \mathbf Z^L \cdot \widehat{\mathbf Y}^L.
\]

\begin{figure}[!t]
\begin{center}
\includegraphics[width=0.95\textwidth]{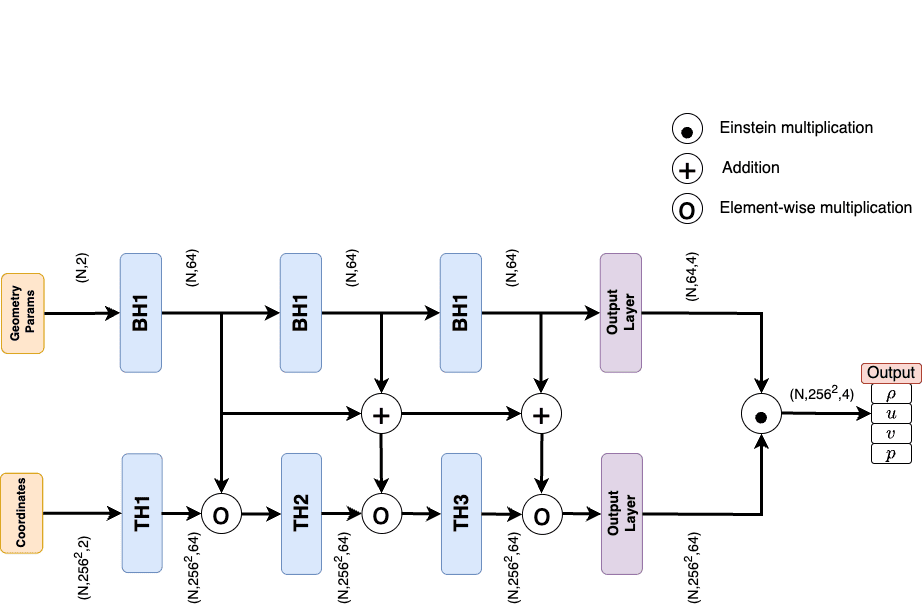}
\caption{Fusion-DeepONet schematic. All the tensor dimensions are shown. $\textrm{BH}I$ refers to the branch $I^\textrm{th}$ hidden layer and $\textrm{TH}I$ denotes trunk net $I^\textrm{th}$ hidden layer. $256^2$ denotes the number of grid points used in the uniform grid. For the irregular grid, the number of points is the maximum point count in the training dataset. $N$ denotes number of training samples.}
\label{fig:Deeponet_fusion}
\end{center}
\captionsetup{justification=centering}
\end{figure}

For Fusion-DeepONet, every hidden layer output of the neural field is conditioned using the branch hidden layer output while retaining the global conditioning of the trunk network. In Fusion-DeepONet, as shown in Fig.~\ref{fig:Deeponet_fusion}, the output of the first hidden layer of the branch is used to condition the output of the first, second, and third hidden layers of the trunk network. As mentioned in the study by Peyvan \emph{et al.} \cite{riemannonet}, the first layers of the fully connected neural nets learn the low-mode features of the target data. Therefore, the low-mode features encoded by the branch layer modulate the low-to-highest-mode functions of the basis functions of the target solution; hence, the conditioning of the neural field occurs on multiple scales. Cross-scale conditioning of the neural field significantly improves the accuracy of the geometry-dependent discontinuous flow field. We later show that the fusion of branch layers into a trunk significantly reduces the over-fitting due to the limited training data.

According to Fig.~\ref{fig:Deeponet_fusion}, we assume that the number of hidden layers in the branch and trunk networks are the same and that the width of the hidden layers of the trunk and branch networks are the same, except for the last linear layer of the branch network. The width of the last linear layer of the branch network is selected to account for multiple target fields such as density, x-velocity, y-velocity, pressure, or temperature. The Fusion-DeepONet consists of extra connections and pathways among the neurons of the neural operators. Therefore, we expect that the forward and backward passes of the Fusion-DeepONet will require extra computational cost compared to the Vanilla-DeepONet for the same number of trainable parameters. We emphasize that the modulation of the hidden layers of the trunk network comes from the branch layers upstream of the current hidden layer of the trunk.

\subsection{Derivative enhanced loss function}
\label{method_der}
To accurately predict the shear stress and heat flux on the solid surfaces in the physical domain, we train the neural operator in a way that can simultaneously learn the solution field, such as temperature, and also the spatial gradients such as $\partial T/\partial x$ and $\partial T/\partial y$. Training the neural operator to learn, for instance, temperature, does not guarantee accurate prediction of the temperature derivative with respect to spatial coordinates. Therefore, we modified the loss function so the model would be aware of the spatial derivative during the training updates. Qiu \emph{et al.}\cite{qiu2024derivative} introduced a derivative-enhanced DeepONet (DE-DeepONet), which integrates derivative information (G$\hat{\textrm{a}}$teaux derivatives) into the MSE loss function to improve the accuracy of predicting the directional derivative of the neural operators with respect to the branch input parameters. In this study, we wish to improve the prediction accuracy of the machine learning model of the derivative with respect to the spatial coordinates. Therefore, we introduce a derivative-enhanced loss (DEL) function that informs the neural operator about the derivative of the solution field. 

Let $\hat{\mathcal{G}}(x,y,m;\theta)$ denotes the output of the neural operator with $\theta$ trainable parameters of the network and $m$ the input geometric parameter; we minimize the following loss function

\begin{equation}
L(\theta)=\|\mathcal{G}(m)-\hat{\mathcal{G}}(m;\theta)\|^2_{L^2(\Omega)}+\lambda_1 \|d\mathcal{G}(m)-\hat{d}\hat{\mathcal{G}}(m;\theta)\|^2_{L^2(\Omega)},
    \label{eq:loss}
\end{equation}
where the $L^2$ norm of the function $f$ is defined as $\|f\|^2_{L^2(\Omega)}=\int_{\Omega}\left|f(\mathbf x)\right|^2d\mathbf x$. In Eq.~\eqref{eq:loss}, $d$ and $\hat{d}$ denote derivative operators that act on the ground truth and the neural operator prediction, respectively. In this study, we investigate the best approach to generate a derivative operator to use in the derivative loss to enhance the accuracy of Fusion-DeepONet in predicting the derivation of the flow field. We evaluate several approaches that best represent the derivative operation with the least computational costs. 

First, we use the same operator for $d$ and $\hat{d}$, and represent the derivative using the following formulation. Let $(x_i,y_i)$ and $(x_{i+1},y_{i+1})$ be two consecutive coordinate points in the tensor of coordinates, and $\phi_i$ and $\phi_{i+1}$ be the values of the predicted solution on these two points. We construct a derivative operator such that
\begin{equation}
d=\hat{d} = \frac{\phi_i-\phi_{i+1}}{\sqrt{(x_i-x_{i+1})^2+(y_i-y_{i+1})^2}}.
    \label{eq:derive_discrete}
\end{equation}
Therefore, we apply this operator to both the predicted value by the neural operator and the ground truth. This derivative operator is straightforward; however, it provides useful information to force the neural operator to learn derivative information. This operator is not computationally expensive and can be vectorized. It is clear that this operator does not represent the discrete form of $\nabla \phi$; however, it provides necessary information to improve the neural operator's predictability for the field's derivatives.

Second,  since we train neural operators using discrete points from a point cloud structure, we define a local least‐squares approximation of the gradient at each point. Then, we project that gradient onto a given direction v to compute the directional derivative. We calculate the directional spatial derivative of the output using the finite differencing method. For each point $i$, we first find $k$ closest points. At each point $i$, we approximate the true scalar field $u(x,y)$ by a local linear model 
\begin{equation}
u(\mathbf{x}_i+\Delta \mathbf{x})\approx u(\mathbf{x}_i)+\nabla u(\mathbf{x}_i).\Delta \mathbf{x}.
    \label{eq:approx_der}
\end{equation}
Now, we estimate $\nabla u(\mathbf{x}_i)$ from the $k$ neighbors around $\mathbf{x}_i$. Let $\mathbf{A}$ be a $k\times2$ matrix for each central point $i$. The rows of matrix $\mathbf{A}_i$ are the displacement vectors $\Delta\mathbf{x}_{i\rightarrow i_j}=\mathbf{x}_{i_j}-\mathbf{x}_i$ and $\mathbf{b}_i$ be a length-$k$ vector of the corresponding changes in the scalar field $\Delta u_{i\rightarrow i_j}=u(\mathbf{x}_{i_j})-u(\mathbf{x}_{i})$. We solve a least squares problem 
\begin{equation}
\min_\mathbf{g}\|\mathbf{A}_i \mathbf{g}-\mathbf{b}_i\|_2^2
    \label{eq:least}
\end{equation}
for the gradient vector $\mathbf{g}$. The matrix form for Eq.~\eqref{eq:least} can be expressed as 
\begin{equation}
\left(\mathbf{A}_i^\top \mathbf{A}_i\right)\mathbf{g}=\mathbf{A}_i^\top \mathbf{b}_i,
    \label{eq:norm_eq}
\end{equation}
thus 

\begin{equation}
\nabla u(\mathbf{x}_i)\approx \mathbf{g}_i = \left(\mathbf{A}_i^\top \mathbf{A}_i\right)^{-1}\mathbf{A}_i^\top\mathbf{b}_i.
    \label{eq:grad}
\end{equation}
Finally, the directional derivatives can be computed by 
\begin{equation}
D_v u(\mathbf{x}_i)=\nabla u(\mathbf{x}_i).v.
    \label{eq:deriv}
\end{equation}
In summary, this is a scattered‐data finite‐difference scheme: we fit a local linear model around each point using its nearest neighbors, then read off the directional derivative from that model. To reduce the computational cost, we precompute the nearest points' index and the ground truth's directional gradient. For this approach, we apply this least squares derivative operator to discretely compute the directional derivative of the neural operator prediction and ground truth. In the results section, we investigate the accuracy and computational cost of the two approaches to enhance the derivative predictability of the neural operator.

\section{Results}

This section presents the results of three geometry-dependent surrogate modeling of high-speed flows using Fusion-DeepONet. We start with two hypersonic problems: inviscid hypersonic flow around semi-elliptic blunt geometries and hypersonic viscous flow over a reentry capsule. We then finish the section by learning an internal supersonic flow in a converging-diverging nozzle connected to an isolator duct. The flow field of all three problems involves strong discontinuities in the solution, where we have shock waves, which are challenging for operator networks to learn. Additionally, using scarce training data in training severely impacts the generalization error of such networks. Further, by evaluating operator networks on unstructured grids commonly employed by numerical solvers for high-speed flow problems, we disrupt the bijectivity in input-output mapping, which is required for practical neural network training. According to Figs.~\ref{fig:mesh}, \ref{fig:caspule}, and \ref{fig:cd_cases}, the geometry of the physical domain in all three problems changes significantly from case to case, and along with it, the grid points change. Learning complex input-output mapping is essential for real-world geometry optimization problems, such as designing the complex 3D geometry of aerial vehicles traveling at hypersonic speeds.

\subsection{Inviscid hypersonic flow around semi-ellipses}
For this problem, we randomly selected 28 cases for training and 8 cases for testing randomly. We used the same training and testing datasets for all neural operators. We train six different neural operators, including Vanilla-DeepONet, POD-DeepONet, and Fusion-DeepONet with Rowdy adaptive activation functions, parameter-conditioned U-Net, FNO, and MeshGraphNet with individual architecture setups discussed in Appendix~\ref {append:neural_ops}. We used the mean squared error of the solution field as the loss function for training all operator models.

We have three versions of the vanilla, POD-DeepONet, and Fusion-DeepONets that are learning the hypersonic field over uniform and irregular grids. The architecture details of all variants of DeepONet are shown in Table~\ref {tab:dno_details}. For Fusion and Vanilla-DeepONet, we use three hidden layers with 64 neurons in the branch network and three with 64 neurons in the trunk network. The adaptive Rowdy activation function \cite{riemannonet} is employed for both trunk and branch networks. The base function for the rowdy activation function is $\tanh$. Adam optimizer and a variable learning rate are used to train the DeepONet framework for 50,000 epochs. We vary the learning rate using the exponential decay function with the step size, decay step, and decay rate set as $0.001$, $2000$, and $0.91$ respectively. For the POD-DeepONet, we should note that there is no trainable trunk network; we only use the basis functions computed by the singular value decomposition (SVD) of the 28 training samples. For the last layer of the branch network, we select the number of neurons to be $64\times 4$, where $4$ corresponds to the four variables predicted by the neural operator. 

For the uniform grid, we input $256^2$ coordinates with shape $(28,65536,2)$ into the trunk net and a matrix that contains the values of $a$ and $b$, which are the geometric parameters with the shape of $(28,2)$ into the branch network. In the case of a uniform Cartesian grid, we multiply the mask matrix into the difference of predictions and the ground truth to zero out the values inside the semi-ellipse and diminish their effect on the update of the neural network parameters. The Vanilla and Fusion-DeepONets then predict all four variables, including density, x-direction velocity, y-direction velocity, and pressure, at the same time using a single trunk network. However, for the POD-DeepONet, we employ four trunk networks, assigning each to the density prediction, x-direction velocity, y-direction velocity, and pressure. For the irregular grid, the input of the trunk net is a tensor with the shape of $(28,npts,2)$ where the $npts$ refers to the maximum number of grid points across the training samples to learn the flow over an irregular grid. The prediction output will then take a shape as $(28,npts,4)$, where the last axis refers to $\rho$, $u$, $v$, and $p$, respectively. The total number of parameters used for Vanilla and Fusion-DeepONet is $37,854$, including the weights and biases, and the learnable parameters of the Rowdy activation function. 

\begin{table}[!t]
\caption{Architecture details for all variants of DeepONets}
\label{tab:dno_details}
\begin{center}
\begin{tabular}{@{}lccc@{}}
\toprule
\multicolumn{1}{c}{\textbf{Variants}} & \textbf{Trunk layers} & \textbf{Branch Layers} & \textbf{Activation} \\ \midrule
Vanilla & {[}2, 64, 64, 64, 64{]} & {[}2, 64, 64, 64, 64*4{]} & Rowdy \\
POD & {[}28*4{]} & {[}2, 64, 64, 64, 28*4{]} & Rowdy \\
Fusion & {[}2, 64, 64, 64, 64{]} &  {[}2, 64, 64, 64, 64*4{]} & Rowdy \\ \bottomrule
\end{tabular}%
\end{center}
\end{table}

\begin{table}[!t]
\caption{Relative $L^2$ norm of the primitive variables errors for different frameworks showing the accuracy achieved. GS refers to Grid Spacing, \#P refers to number of trainable parameters.}
\label{tab:result_error}
\resizebox{\columnwidth}{!}{%
\begin{tabular}{@{}lclcccccc@{}}
\toprule
\textbf{Framework} & \textbf{GS} & \textbf{\#P} & $\%\textrm{L}^2 (\rho)$&$\%\textrm{L}^2 (u)$&$\%\textrm{L}^2 (v)$&$\%\textrm{L}^2 (p)$&$\%\textrm{L}^2$ & t(sec)\\ \midrule
Fusion-DeepONet & Uniform   & 37,854    & $\mathbf{3.41}$  & $\mathbf{0.68}$   & 5.91   & 4.80   & $\mathbf{3.70}$   &1471.05 \\
Vanilla-DeepONet & Uniform   & 37,854 & 33.2  & 8.43   & 52.4   & 51.57  & 36.41  & 78.6 \\
POD-DeepONet & Uniform   & 15,807 & 25.29  & 5.46   & 39.21   & 38.76  & 27.18  & 177.4 \\
U-Net           & Uniform   &  7,056,116      & 3.68    & 2.39      & 9.36   & 2.74   & 4.54   & 3164.9\\
FNO             & Uniform     & 1,71,556,804     & 10.4      & 8.41      & 1.97   & 12.08   & 8.21   & 1650.3 \\
Fusion-DeepONet & Irregular   & 37,854 & $\mathbf{6.18}$   & $\mathbf{2.79}$   & $\mathbf{9.29}$   & $\mathbf{7.29}$   & $\mathbf{6.39}$   & 536.2\\
Vanilla-DeepONet & Irregular   & 37,854 & 39.89   & 16.77   & 52.66   & 47.48   & 39.20   & 369.78\\
MeshGraphNet    & Irregular     & 277,061      &  47.33     &  30.22     & 58.81   & 62.40   & 49.46   & 1178.3\\
\bottomrule
\end{tabular}%
}
\end{table}

Table~\ref {tab:result_error} shows the prediction error for the testing dataset and the computational time of training. All the training is performed using a single NVIDIA 3090 GeForce RTX GPU. The hypersonic flow is predicted on the uniform and irregular grid. Therefore, we compare various neural operators trained using both types of grids. Focusing on the uniform grid, Fusion-DeepONet shows the best accuracy in predicting all four variables compared to Vanilla-DeepONet, parameter-conditioned U-Net, and FNO neural operators. Fusion-DeepONets and U-Nets provide comparable results, while FNO falls short in matching the prediction accuracy compared to these two frameworks on a uniform grid. The Vanilla-DeepONet performs the worst. In terms of computational cost, Vanilla-DeepONet has the lowest computational cost compared to other frameworks. The Fusion-DeepONet employs the least number of parameters compared to U-Net and FNO while providing comparable accuracy to U-Net. The computational time for training Fusion-DeepONet over 50,000 epochs is higher compared to Vanilla-DeepONet. This is due to two main reasons. First, the forward and backward passes in Fusion-DeepONet involve addition and multiplication operations defined in Eqs.~\eqref{eq:skip_connections} and \eqref{eq:conditioning}. Second, the input to the trunk network in Fusion-DeepONet has a shape of $(28, 65536, 2)$, whereas in Vanilla-DeepONet it is $(65536, 2)$. In Vanilla-DeepONet, the trunk network is not conditioned on the output of the branch network, allowing the use of the same set of spatial coordinates across all 28 training samples. As a result, Fusion-DeepONet incurs a higher computational cost. However, in scenarios where each sample has a unique coordinate grid (e.g., irregular grids), Fusion-DeepONet offers higher accuracy at the cost of about a 30\% increase in computational overhead compared to Vanilla-DeepONet. In conclusion, Fusion-DeepONet predicts accurately and employs minimum trainable parameters compared to U-Net and FNO.

Learning flow fields on irregular grids presents unique challenges, as only a limited number of operator networks can handle unstructured grid coordinates without requiring additional pre-processing. Among these, Vanilla-DeepONet and the more recent MeshGraphNet are capable of operating on arbitrary mesh grids for output functions. Fusion-DeepONet, being a variant of Vanilla-DeepONet, retains its core capabilities and is therefore included in this comparison. For a fair evaluation, we use the same number of parameters for both the vanilla and Fusion-DeepONet models on irregular grids as in their uniform grid counterparts. As shown in Table~\ref{tab:result_error}, Fusion-DeepONet is the only model that achieves accurate predictions of output flow fields with limited training data. It consistently outperforms both MeshGraphNet and Vanilla-DeepONet across all predicted variables while also requiring lower computational cost for training. Additionally, we assess Fusion-DeepONet’s performance on a high-pressure ratio Sod shock tube problem, as introduced in \cite{riemannonet}. The results are detailed in Appendix~\ref{sod}.

\begin{figure}[!t]
\begin{center}
\includegraphics[width=\textwidth]{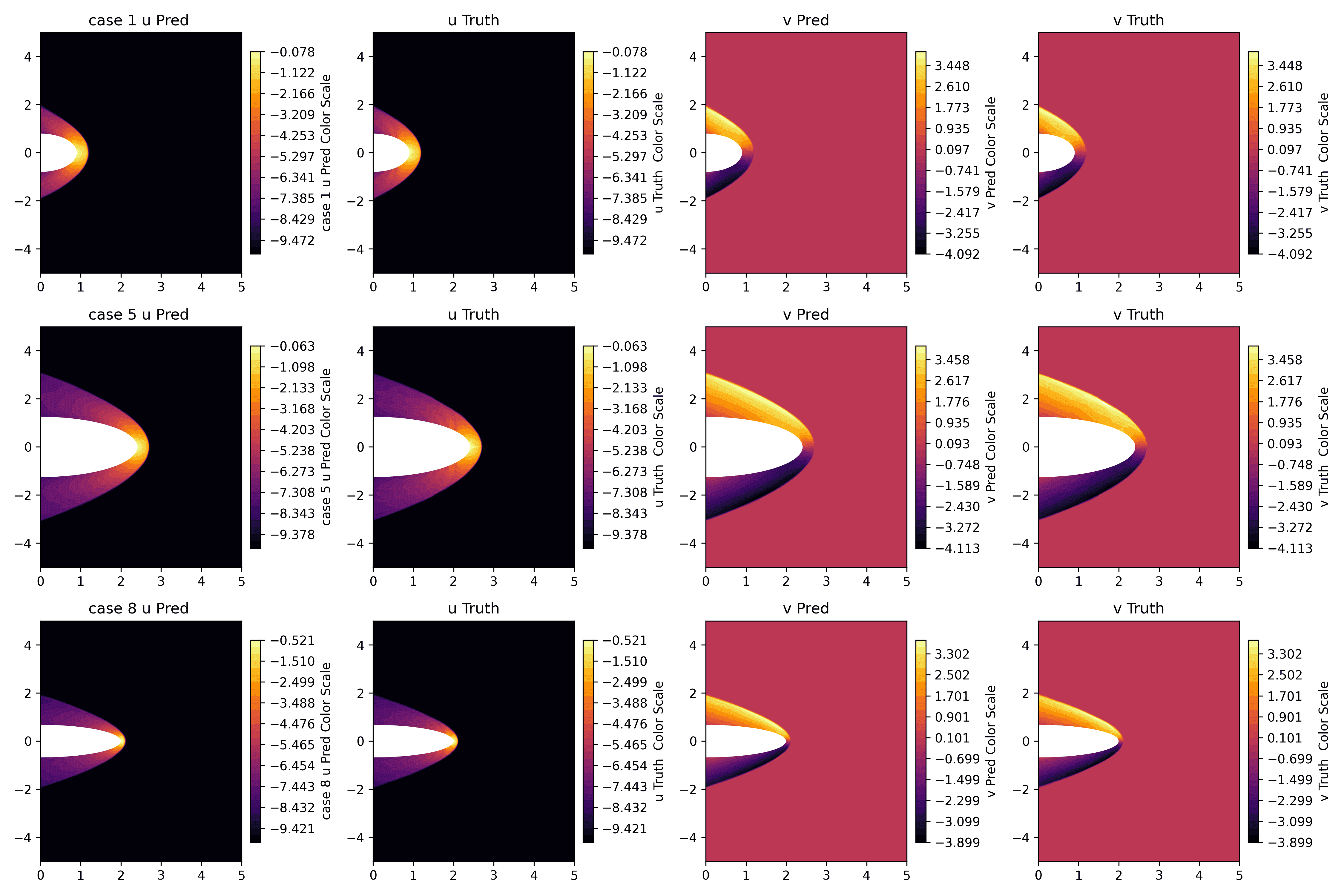}
\caption{Prediction of Fusion-DeepONet for three unseen samples. Each row depicts $u$ and $v$ prediction and ground truth for an unseen sample on uniform grid. the first two columns show the predicted $u$ and the ground truth. The last two columns depict the predicted $v$ and ground truth. Row 1: test case 1, Row 2: test case 5, Row 3: test case 8. See appendix~\ref{append:datasets} for the values of input parameters for cases 1,5 and 8.}
\label{fig:var_samples_un}
\end{center}
\captionsetup{justification=centering}
\end{figure}

\subsubsection{Training on uniform grid}

In this section, we dive deeper into the predicted results using a uniform grid and try to explain the reason behind the exceptional performance of the Fusion-DeepONet framework. According to Fig.~\ref{fig:var_samples_un}, the Fusion-DeepONet framework can accurately predict the hypersonic flow field for a significant geometrical variation between test cases. Figure~\ref{fig:Neural_comp} compares prediction accuracy among Fusion-DeepONet, U-Net, and FNO for an unseen sample by the neural operators. Qualitatively, all the operators can predict the general shape of the solution. However, Fusion-DeepONet better predicts the bow shock edges surrounding the semi-ellipse, as shown by the point-wise error plot in the third column. The FNO prediction of the bow shock edge shows wiggles, and its point-wise error is the highest. The comparison of other variables, such as $u$, $v$, and $p$, shows a similar conclusion as the density plots. Results from FNO show higher error values when compared to Fusion-DeepONet and U-Net for a regular grid setup for all output variables except for the $v$ velocity component, where the error is the lowest. This may be due to using the same modes used to filter all four output variables during training simultaneously. The design parameters for the geometry were not used as inputs to the FNO model, which instead relies on the grayscale mask to map the shape of the geometry. Including the geometric parameters along with the node coordinates may improve FNO predictions; however, this would increase the memory footprint of the input data since each design parameter would have to be replicated across all nodes. Figure~\ref{fig:loss_ug} depicts the MSE loss function value versus the Fusion and Vanilla-DeepONets training epoch. Fusion-DeepONet helps reduce over-fitting with the help of fusion connections and the conditioning of the trunk network. Therefore, the generalization error in Fusion-DeepONet is less than that of Vanilla-DeepONet.

\begin{figure}[!h]
\begin{center}
\includegraphics[width=\textwidth]{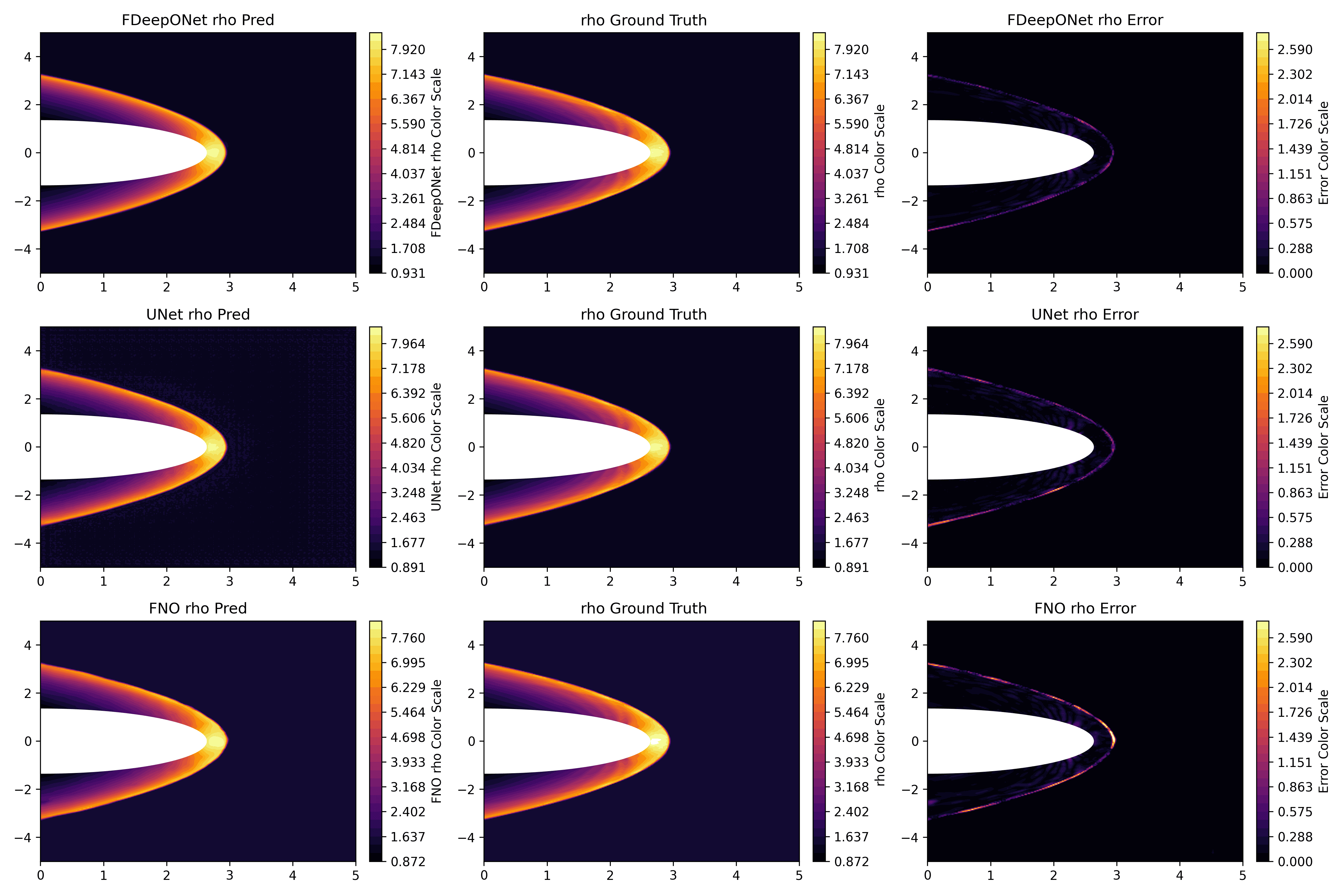}
\caption{Comparison of density prediction using Fusion-DeepONet, U-Net, and FNO neural operators for the unseen case 2 on uniform Cartesian grid. Each row corresponds to a specific neural operator. First row: Fusion-DeepONet, Second row: U-Net, and Third row: FNO. First column shows prediction by the neural operator and second column shows the ground truth with a similar color range as the prediction case. Third column shows absolute value of point-wise error. The color bar for the point-wise error is the same for various neural operators for comparison. See appendix~\ref{append:datasets} for value of input parameters of test case 2.}
\label{fig:Neural_comp}
\end{center}
\captionsetup{justification=centering}
\end{figure}

\begin{figure}
  \begin{center}
    \begin{tabular}{cc}
\includegraphics[width=0.4\textwidth]{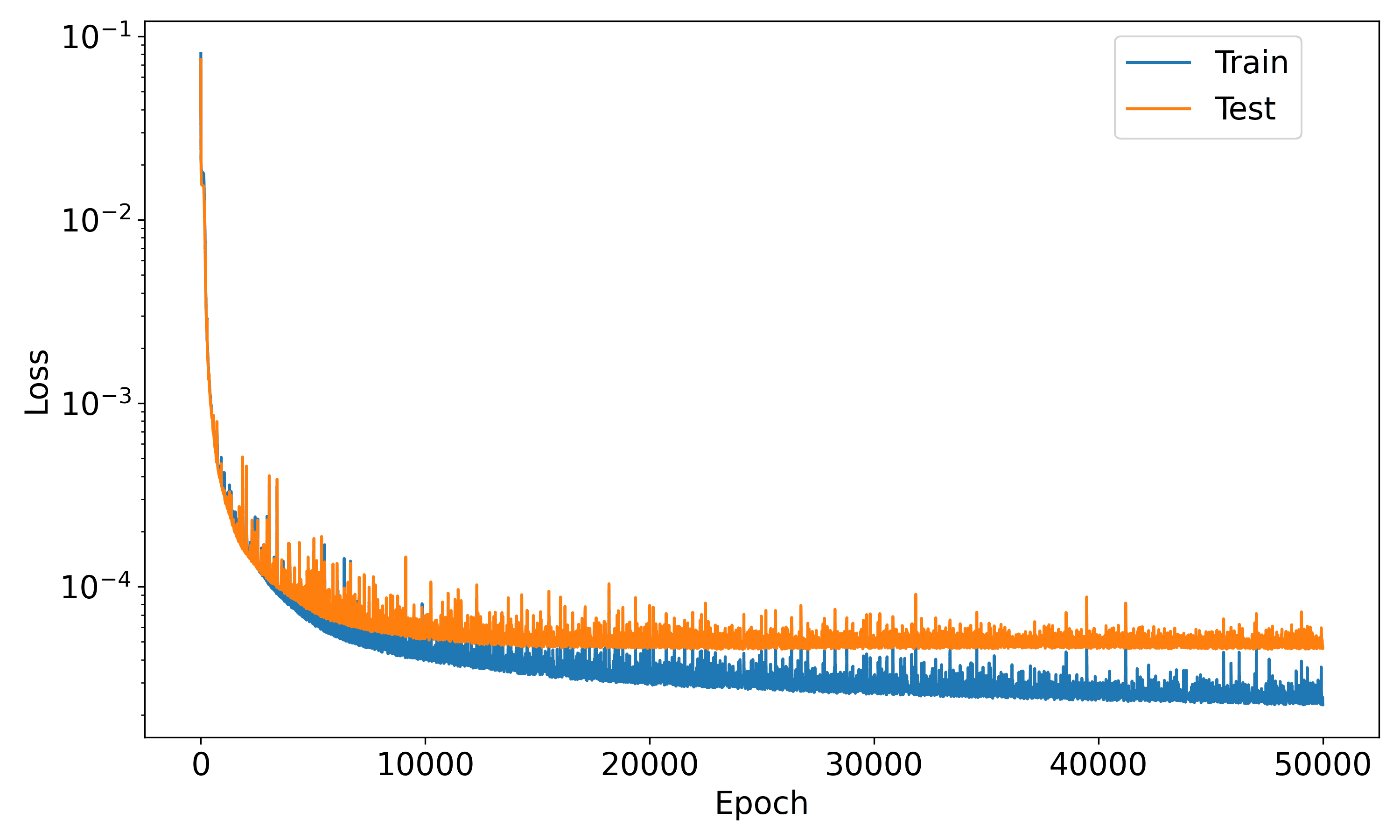}
 &
\includegraphics[width=0.4\textwidth]{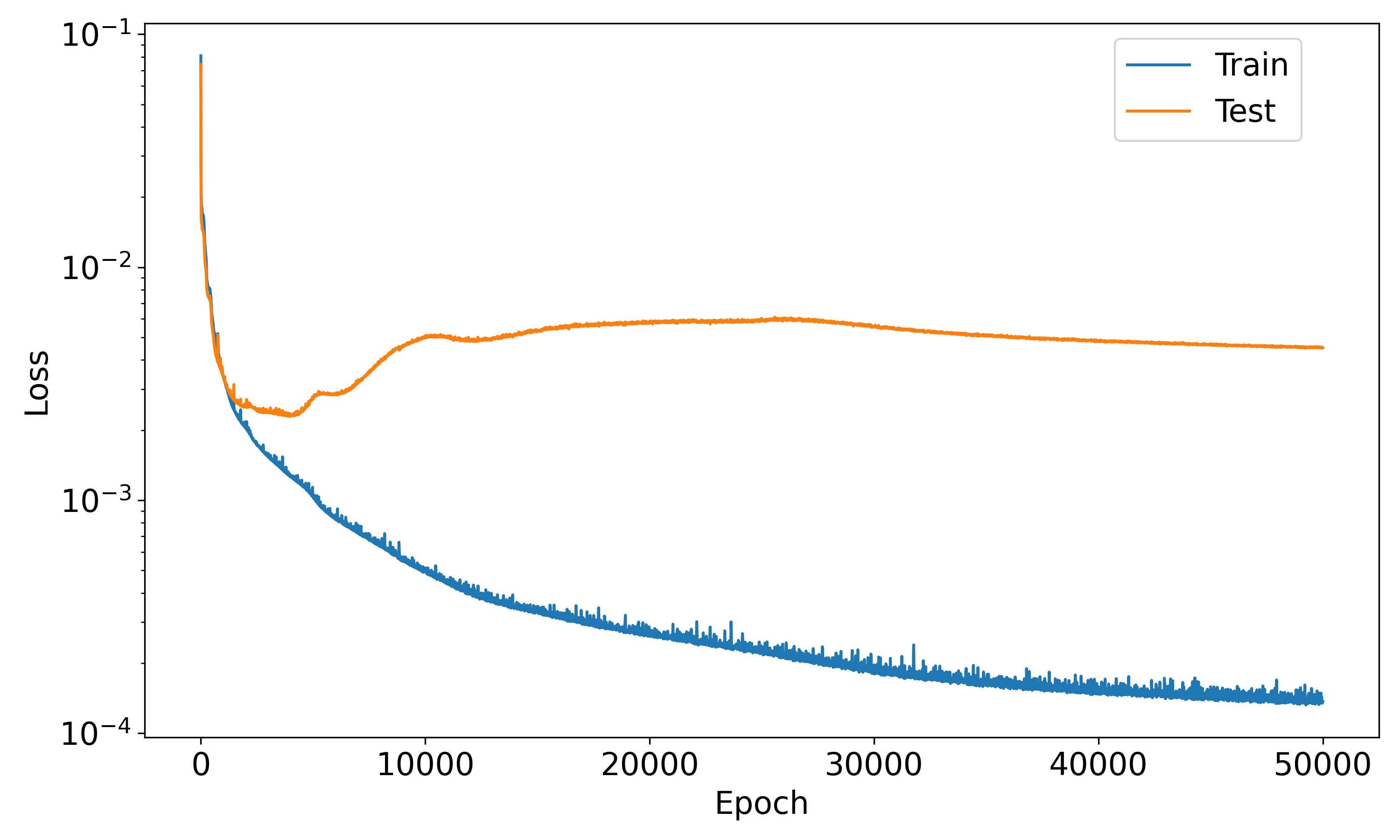}
\\
(a) Loss for Fusion-DeepONet& (b) Loss for Vanilla-DeepONet  
\end{tabular} 
\caption{Loss versus epochs for Fusion and Vanilla-DeepONets uniform grid.}
    \label{fig:loss_ug}
  \end{center}
\end{figure}

\subsubsection{Training on irregular grid}
Prediction of a hypersonic field over an irregular grid poses a significant challenge to neural operators. In the literature, few neural operators can handle irregular mesh and sample-dependent grid points directly. This study uses the Vanilla-DeepONet and MeshGraphNet frameworks to evaluate Fusion-DeepONet performance on irregular grids. We modified the structure of the DeepONet framework into a unique conditioned neural field that can build geometry-informed basis functions. The geometry-informed basis functions encode the grid points and geometry parameters and adapt to the hypersonic flow field simultaneously. MeshGraphNet utilizes message-passing to update node and edge feature embeddings using the initial nodal coordinates and Euclidean distance between the nodes as the node and edge features of the graph network. By utilizing the node coordinates with node connectivity across the whole domain, MeshGraphNet enables the creation of better feature embeddings that can act as basis functions for the nodal predictions. By predicting fields ($\rho, u, v, p$) defined as the output node features trained in a supervised manner, MeshGraphNet maps the structural features of the geometry to the resulting flow fields, thereby eliminating the need for using geometric parameters explicitly.

Figure~\ref{fig:Neural_comp_un} visually compares the predicted solution by Fusion-DeepONet, Vanilla-DeepONet, and MeshGraphNet operators. The Fusion-DeepONet operator can accurately predict the $u$ field without spurious oscillations or excessive diffusion of sharp interfaces in the solution. MeshGraphNet performs the worst for the irregular grid problem and cannot predict the sharp edge of the bow shock or the solution profile between the semi-ellipse and the bow shock. The Fusion-DeepONet excels in learning non-linear geometry-dependent solutions on an irregular grid. Comparing the Vanilla-DeepONet versus the Fusion-DeepONet, we can deduce that the fusion of branch layers into the trunk network improves learning generalization over arbitrary grids. For completeness, we depict the effectiveness of Fusion-DeepONet in learning geometry-dependent hypersonic field for three unseen samples in Fig.~\ref{fig:var_samples_us}. Figure~\ref{fig:loss_ig} demonstrates the MSE loss function value for Fusion-DeepONet and Vanilla-DeepONet trained on cases with irregular unstructured grids. The generalization error for Vanilla-DeepONet is much larger than that of Fusion-DeepONet.

\begin{figure}[!t]
\begin{center}
\includegraphics[width=\textwidth]{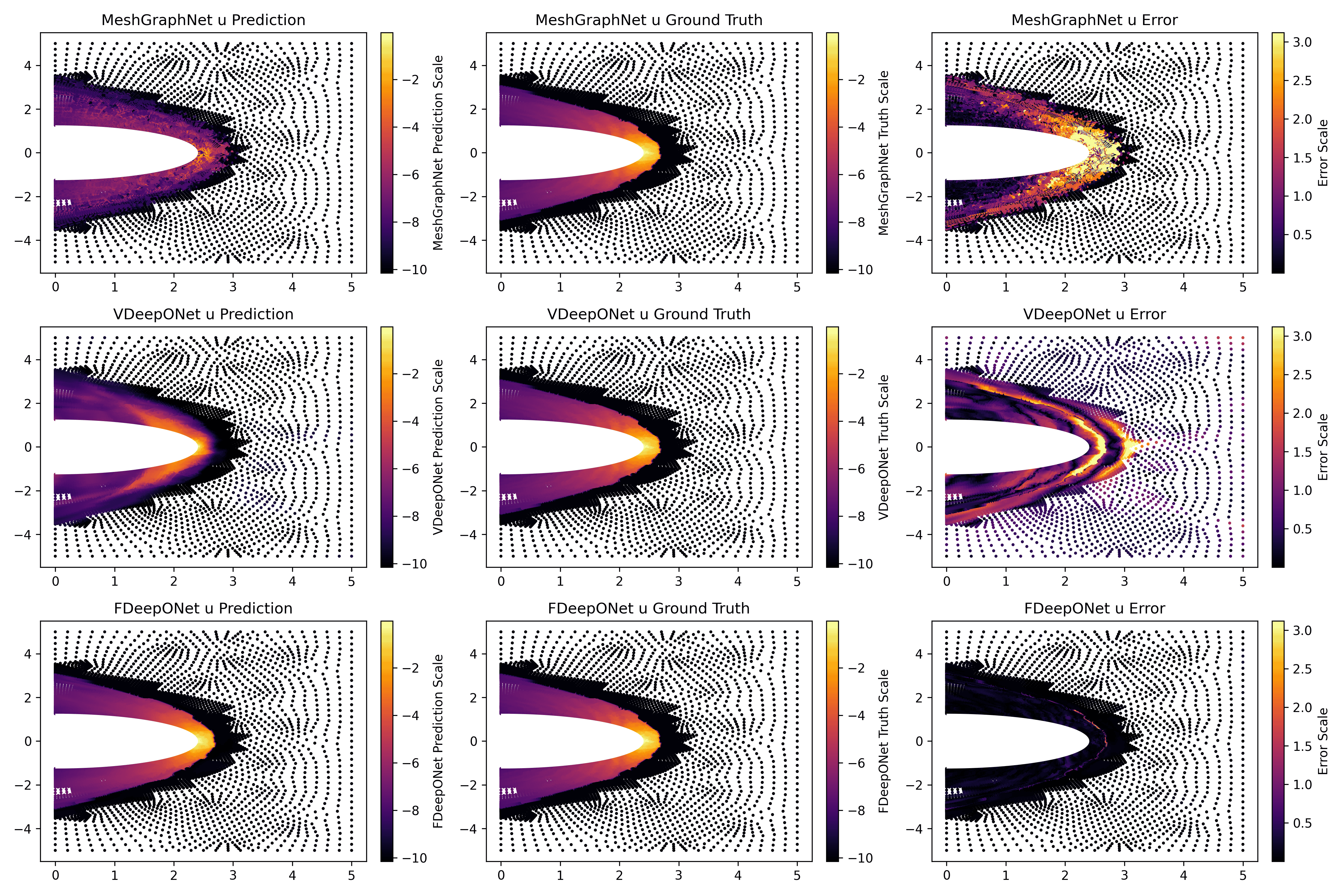}
\caption{Comparison of x-velocity prediction using MeshGraphNet, Vanilla-DeepONet, and Fusion-DeepONet neural operators for the unseen case 5 on an irregular unstructured grid. Each row corresponds to a specific neural operator. First row: MeshGraphNet, Second row: Vanilla-DeepONet, and Third row: Fusion-DeepONet. First column shows prediction by the neural operator and second column shows the ground truth. Third column shows absolute value of point-wise error. The color bar for the point-wise error is the same for various neural operators for comparison. See appendix~\ref{append:datasets} for value of input parameters of test case 5.}
\label{fig:Neural_comp_un}
\end{center}
\captionsetup{justification=centering}
\end{figure}

\begin{figure}[!t]
\begin{center}
\includegraphics[width=\textwidth]{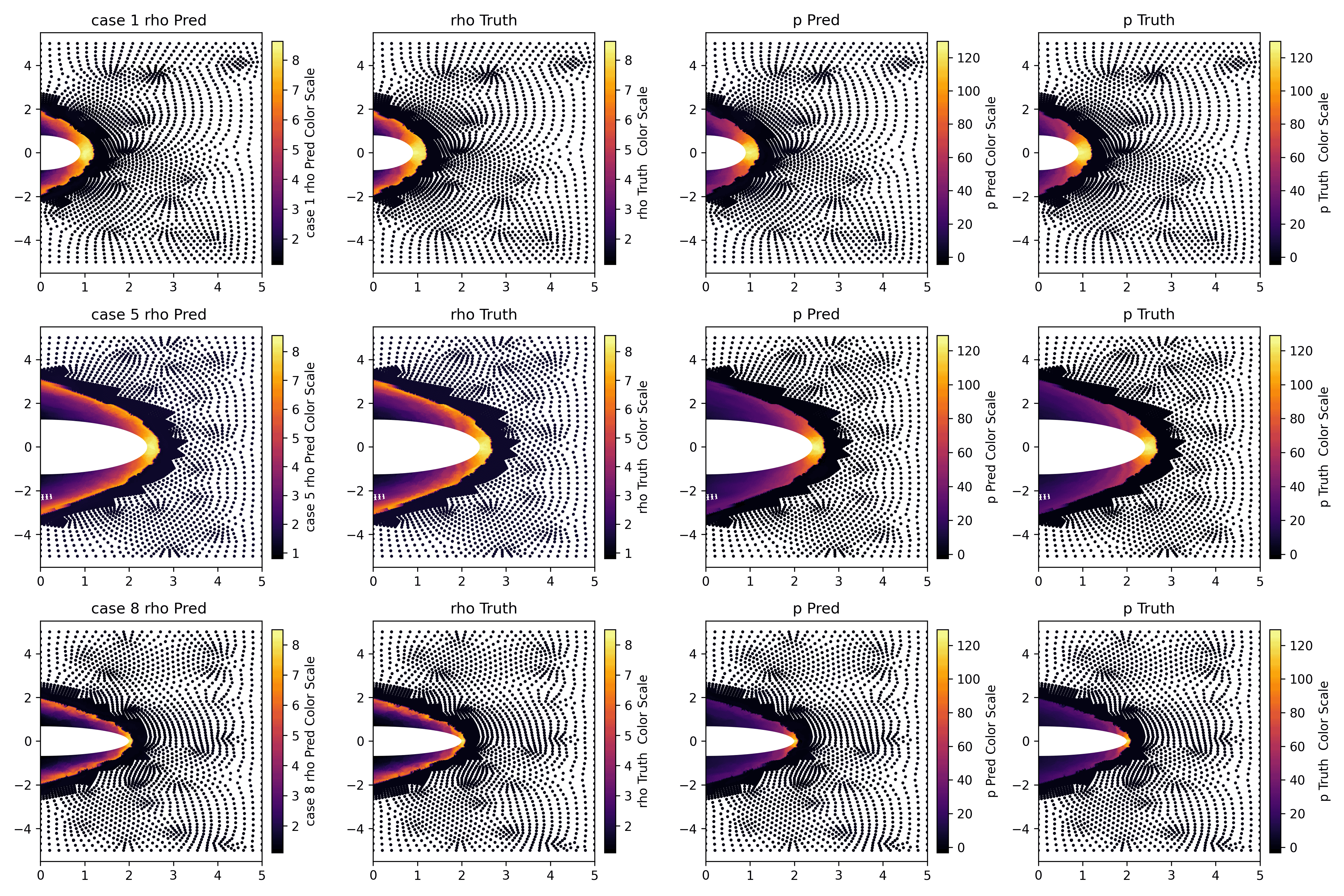}
\caption{Prediction of Fusion-DeepONet for three unseen samples on irregular unstructured grids. Each row depicts $\rho$ and $p$ prediction and ground truth for an unseen sample. the first two columns show the predicted $\rho$ and the ground truth. The last two columns depict the predicted $p$ and ground truth. Row 1: case 1, Row 2: case 5, Row 3: case 8. See appendix~\ref{append:datasets} for the values of input parameters for test cases 1,5 and 8}
\label{fig:var_samples_us}
\end{center}
\captionsetup{justification=centering}
\end{figure}

\begin{figure}[!t]
  \begin{center}
    \begin{tabular}{cc}
\includegraphics[width=0.4\textwidth]{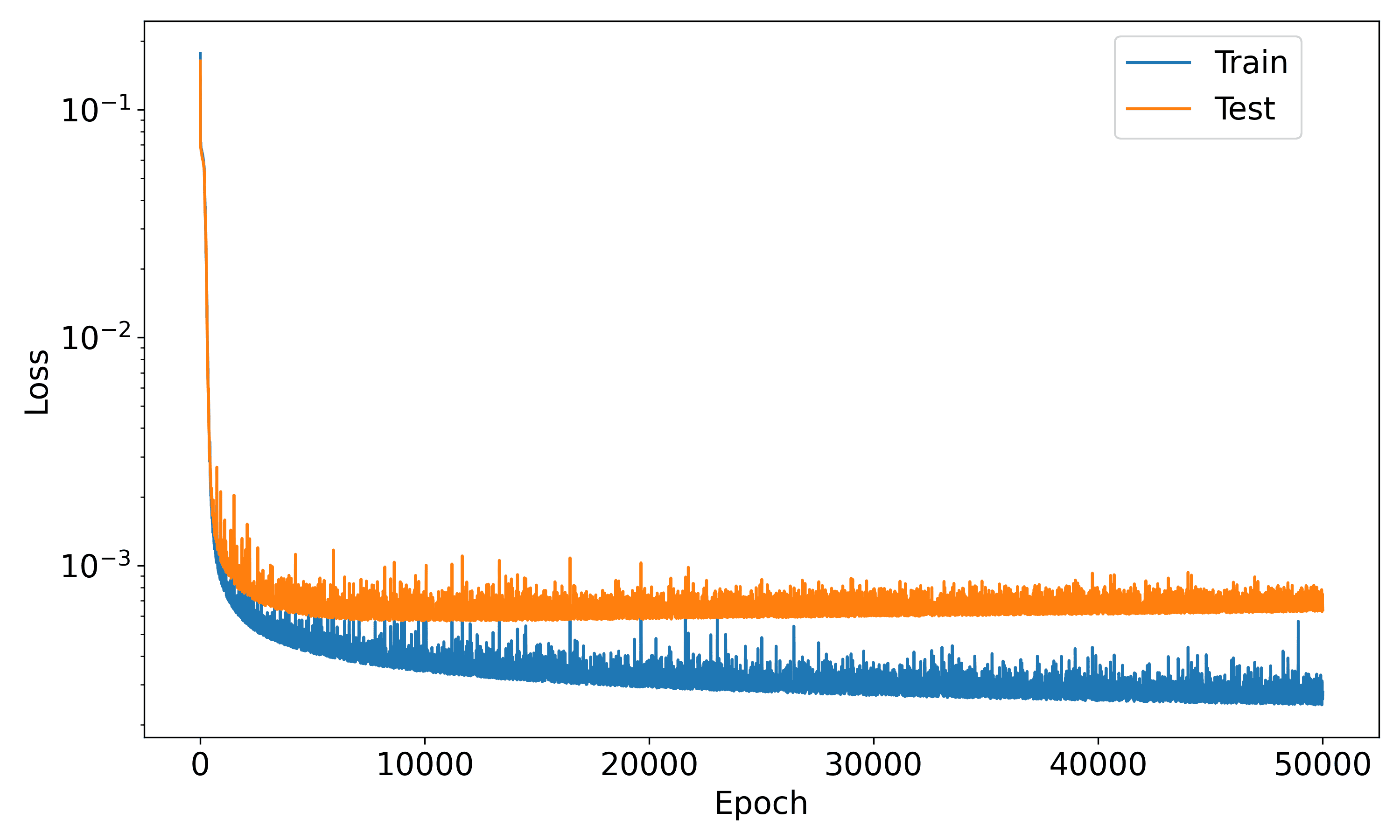}
 &
\includegraphics[width=0.4\textwidth]{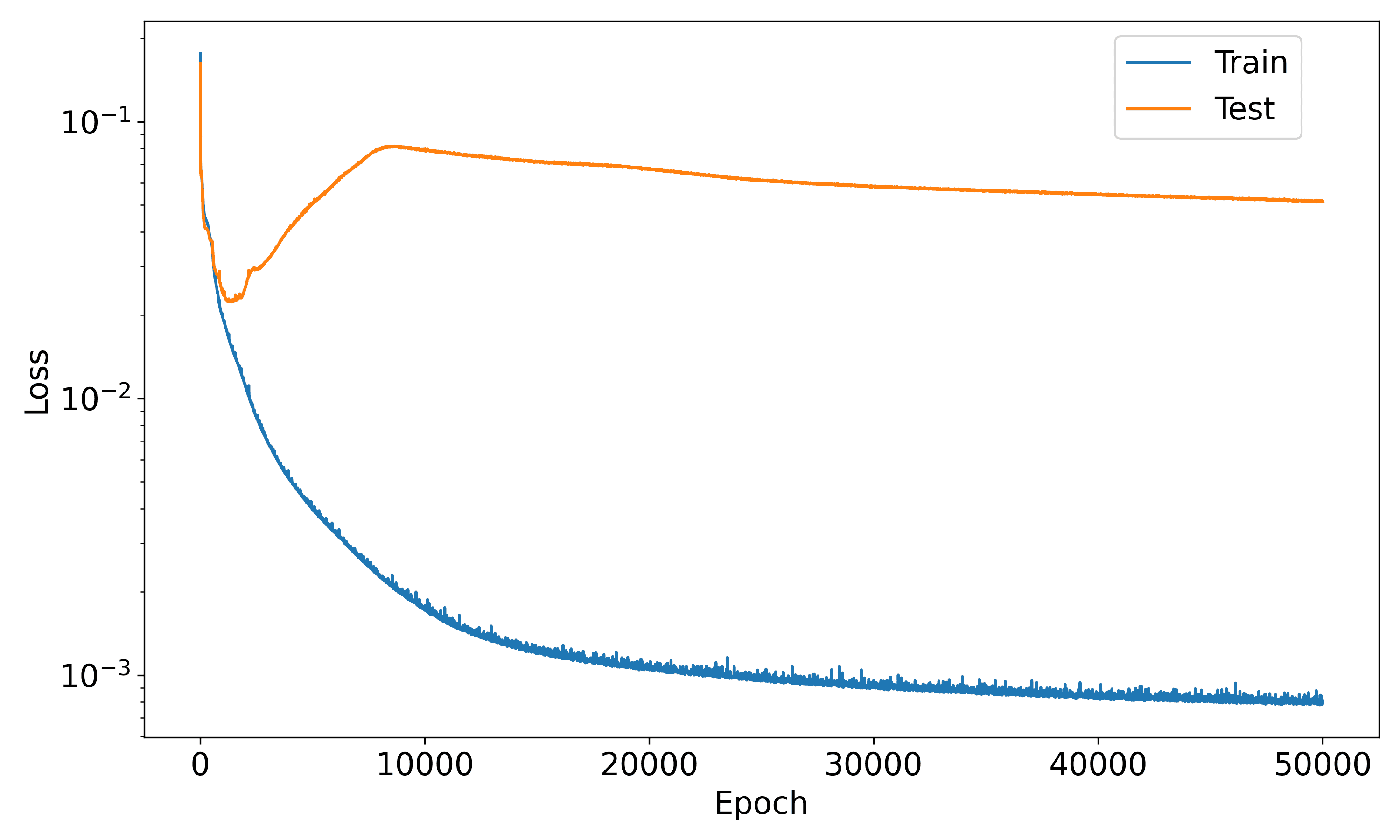}
\\
(a) Loss for Fusion-DeepONet& (b) Loss for Vanilla-DeepONet  
\end{tabular} 
\caption{Loss versus epochs for Fusion and Vanilla-DeepONets on irregular grid.}
    \label{fig:loss_ig}
  \end{center}
\end{figure}

\subsubsection{Interpretation}
This section analyzes the reasons for Fusion-DeepONet's good performance for geometry-dependent problems. Within the framework of the DeepONet, the trunk network encodes the coordinate points into a set of basis functions that are conditioned using the encoded output of the branch hidden layers. We borrow the analysis method of Peyvan \emph{et al.} \cite{riemannonet} to decompose the output of each hidden layer of the trunk network into interpretable entities. We performed this analysis on both uniform and irregular grids. Using the same set of grid points for all the training samples differentiates entirely from using individual discretization for each sample. We first analyze the trunk network of the Fusion-DeepONet and Vanilla-DeepONet in learning the uniform grid problem.  

\begin{figure}[t!]
  \begin{center}
    \begin{tabular}{cc}

\includegraphics[width=0.48\textwidth]{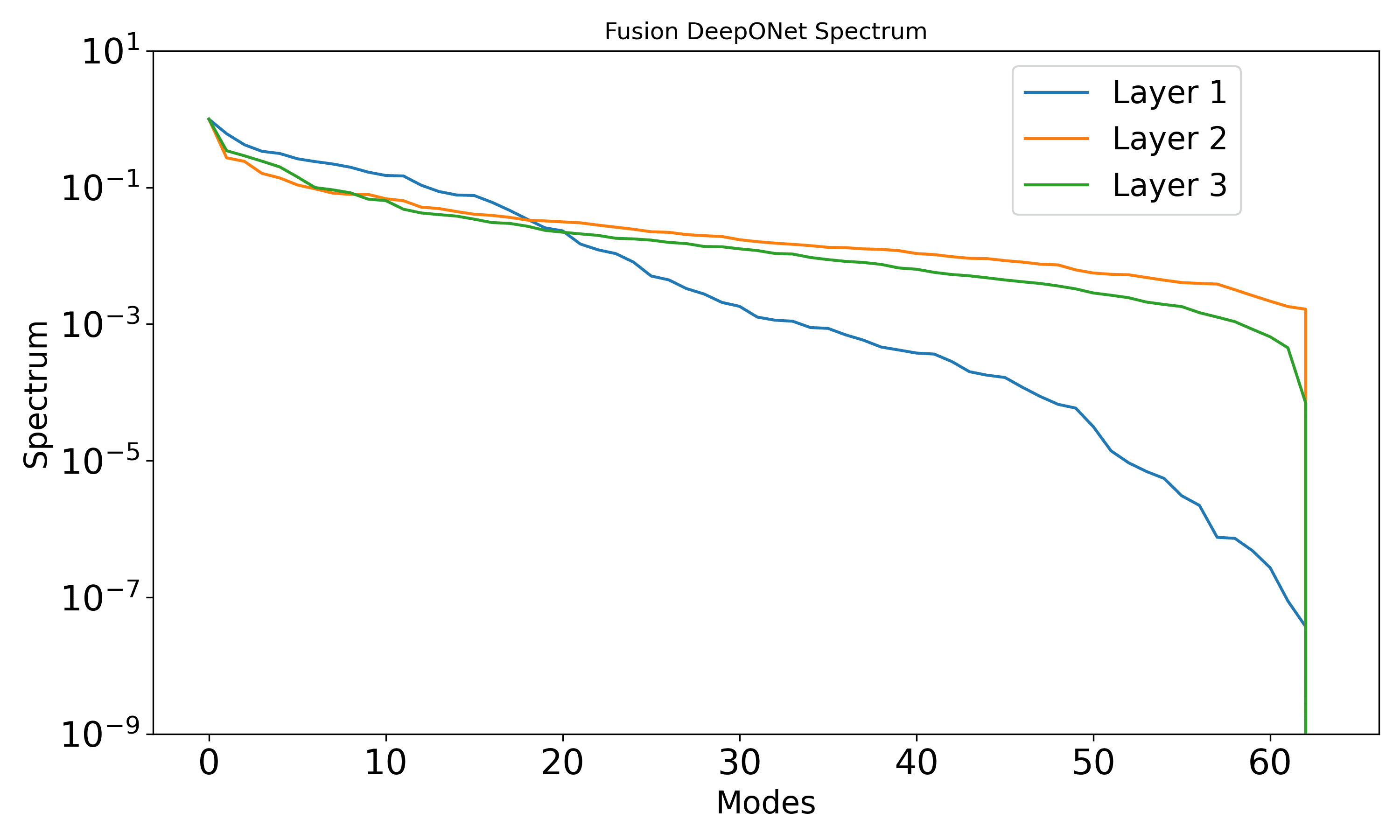}
 &
\includegraphics[width=0.48\textwidth]{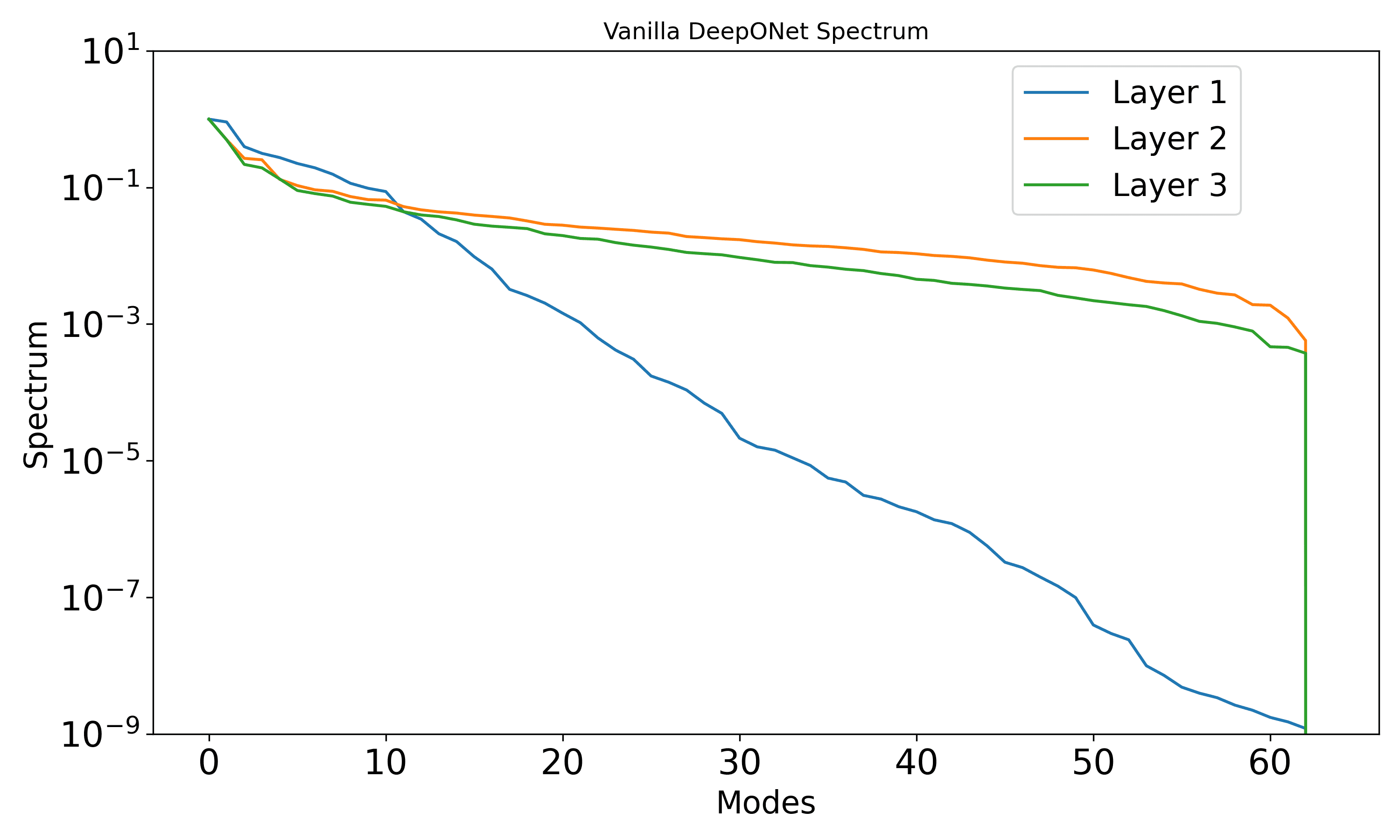}
  \\
  (a) Energy Spectrum Fusion-DeepONet& (b) Energy Spectrum Vanilla-DeepONet   
\end{tabular} 
\caption{Energy spectrum for modes of trunk network hidden layers outputs trained with uniform grid training dataset for Vanilla and Fusion-DeepONet.}
    \label{fig:spectrum_uni}
  \end{center}
\end{figure}

\paragraph{Uniform grid:}
We apply SVD on all the hidden layers of the trunk net to compute 64 eigenvalues corresponding to 64 eigenvectors. Therefore, we evaluate the energy spectrum for each of the 64 modes corresponding to the number of neurons of each hidden layer. Figure~\ref{fig:spectrum_uni}(a) and (b) compare the energy spectrum of the output of hidden layers in the trunk network of Fusion and Vanilla-DeepONets. According to the energy spectra of the first hidden layer output, we observe that the Fusion-DeepONet extracts more information both in the low and highest modes. As a result, the first layer collects high-frequency information due to the connection of the first layer of the branch to the last layer of the trunk network. Conditioning the output of the trunk network layers brings extra information from the geometric parameter encoding into the trunk network to modulate the data-driven basis functions. Comparing the $50^\textrm{th}$ mode of the first layer Fusion and Vanilla-DeepONet shows Fusion-DeepONet's ability to learn high-frequency features. In contrast, the Vanilla-DeepONet can only extract noisy data as shown in Fig.~\ref{fig:basis_function_compare} (b).

\begin{figure}[t!]
  \begin{center}
    \begin{tabular}{cc}

\includegraphics[width=0.48\textwidth]{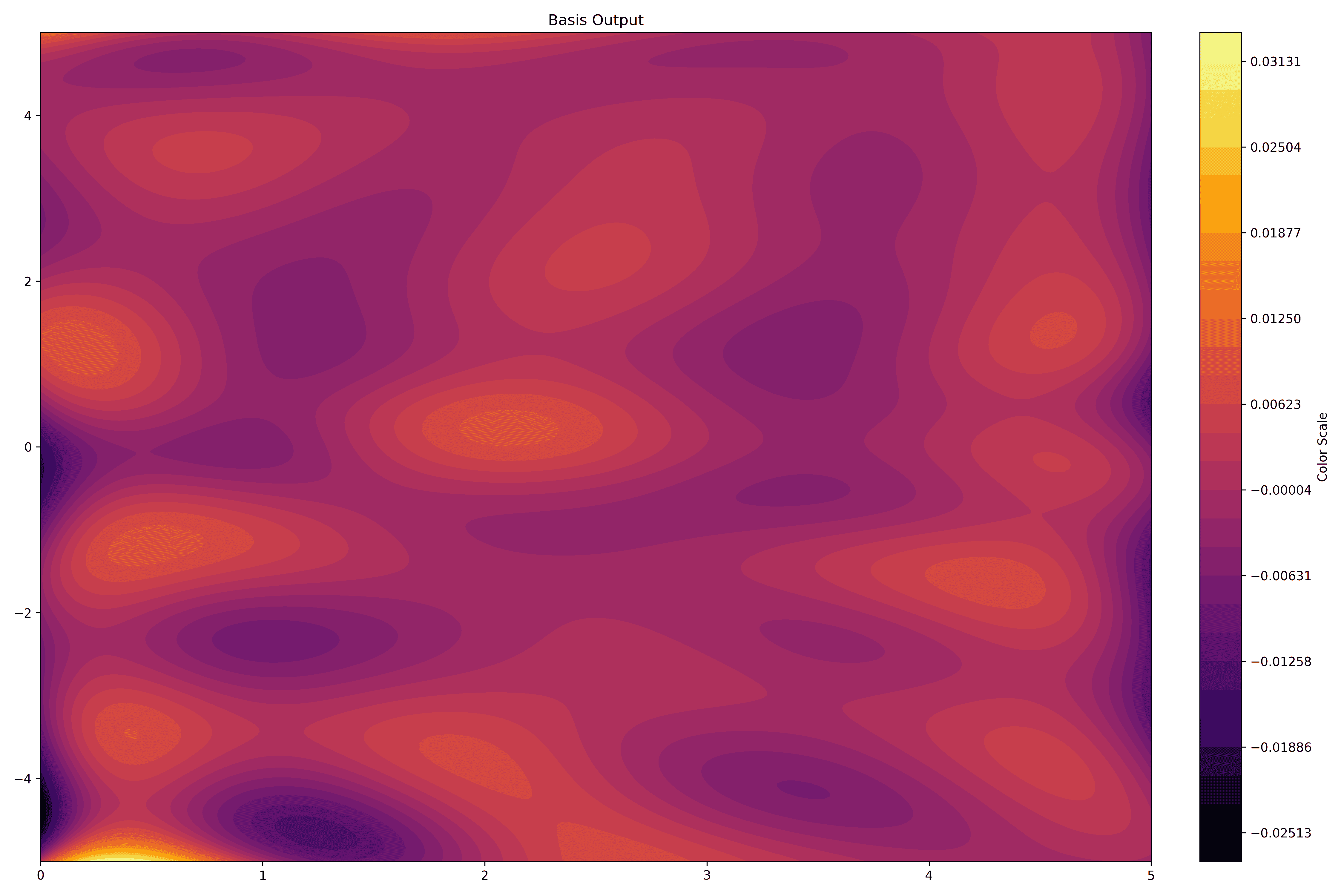}
 &
\includegraphics[width=0.48\textwidth]{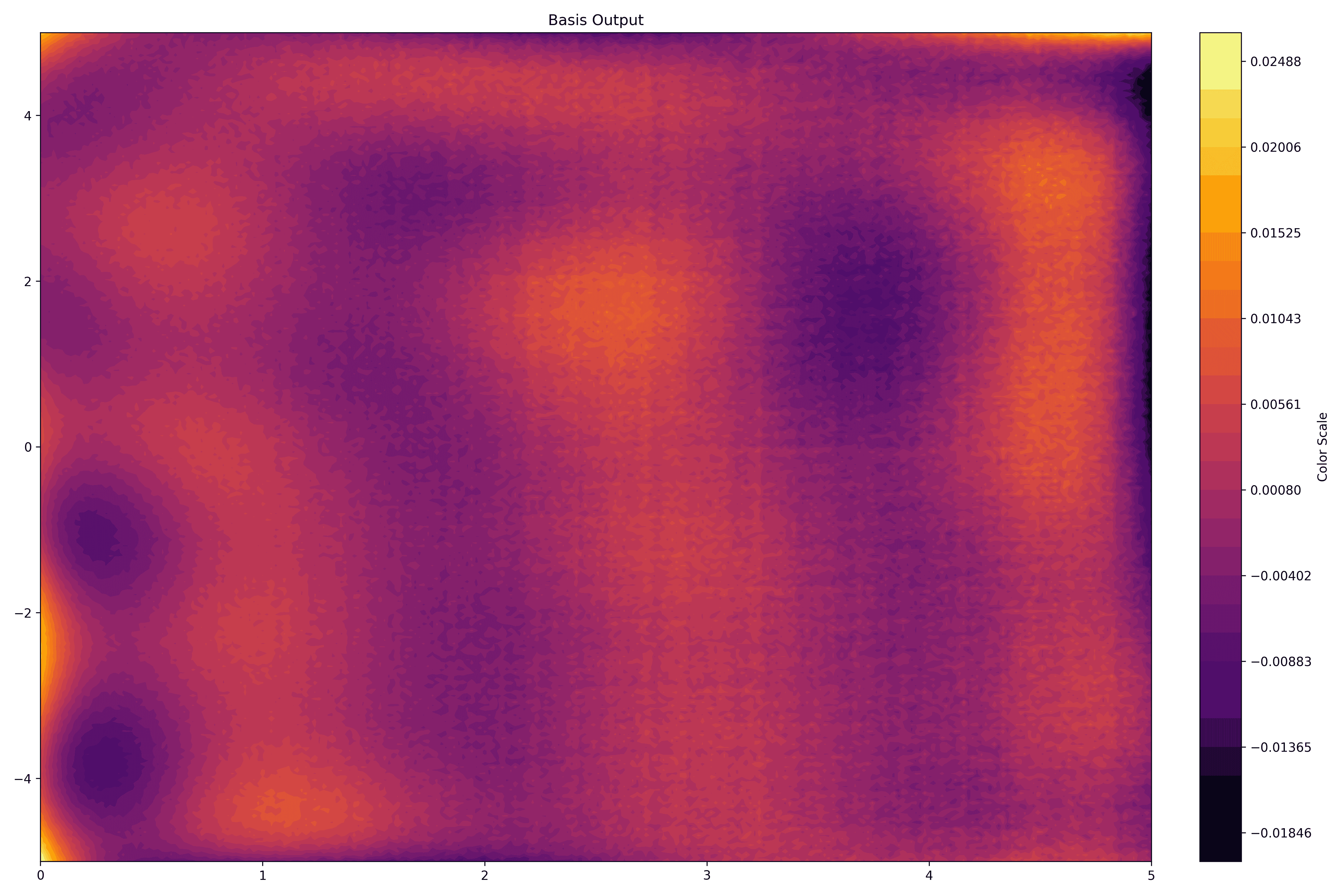}
  \\
  (a) Mode 50 basis function Fusion-DeepONet& (b) Mode 50 basis function Vanilla-DeepONet   
\end{tabular} 
\caption{Mode 50 basis function of trunk network's first hidden layer for (a) Fusion-DeepONet and (b)Vanilla-DeepONet trained on uniform grid.}
    \label{fig:basis_function_compare}
  \end{center}
\end{figure}

\paragraph{Irregular grid:}
Interpreting the prediction of Fusion-DeepONet on an irregular grid is much more challenging than on a uniform grid. The trunk network must construct 64 basis functions corresponding to each training sample for each grid configuration. There exist 28 samples in the training dataset providing 28 different grid configurations. During the training of Fusion and Vanilla-DeepONets, the trunk network must encode information on both the solution structure and mesh point configurations and can generalize both the solution values and mesh point locations. As shown in Fig.~\ref{fig:Neural_comp_un}, Vanilla-DeepONet struggles with predicting the flow field over the changing geometry, while the Fusion-DeepONet can accurately infer the solution. We now investigate the SVD of the output of the three hidden layers in the trunk network of Fusion and Vanilla-DeepONets. Figure~\ref{fig:spectrum_ig}(a)-(f) demonstrates the spectrum of the outputs of the hidden layer for 28 grid configurations. Each plot shows the spectrum of modes for each grid configuration in a particular color. Considering the spectrum on all the 28 grid configurations, the amount of energy captured over the entire modes by the first hidden layer of Fusion-DeepONet is much higher than the Vanilla-DeepONet, as seen in Fig.~\ref{fig:spectrum_ig}(a) and (d). Similarly, the second and the third hidden layers of the Fusion-DeepONet extract more information over the entire range of modes than the Vanilla-DeepONet. Comparing Fig.~\ref{fig:spectrum_ig}(b) and (e), we can observe that the curves corresponding to grid configurations are more scattered in Fusion-DeepONet, while for Vanilla-DeepONet, the curves are clustered together. This clustering of the spectrum for all the grid configurations can hint at the fact that the Vanilla-DeepONet cannot differentiate between various grid configurations and, in other words, cannot learn and generalize on the locations of the coordinate points.

\begin{figure}[t]
  \begin{center}
    \begin{tabular}{ccc}
\includegraphics[width=0.3\textwidth]{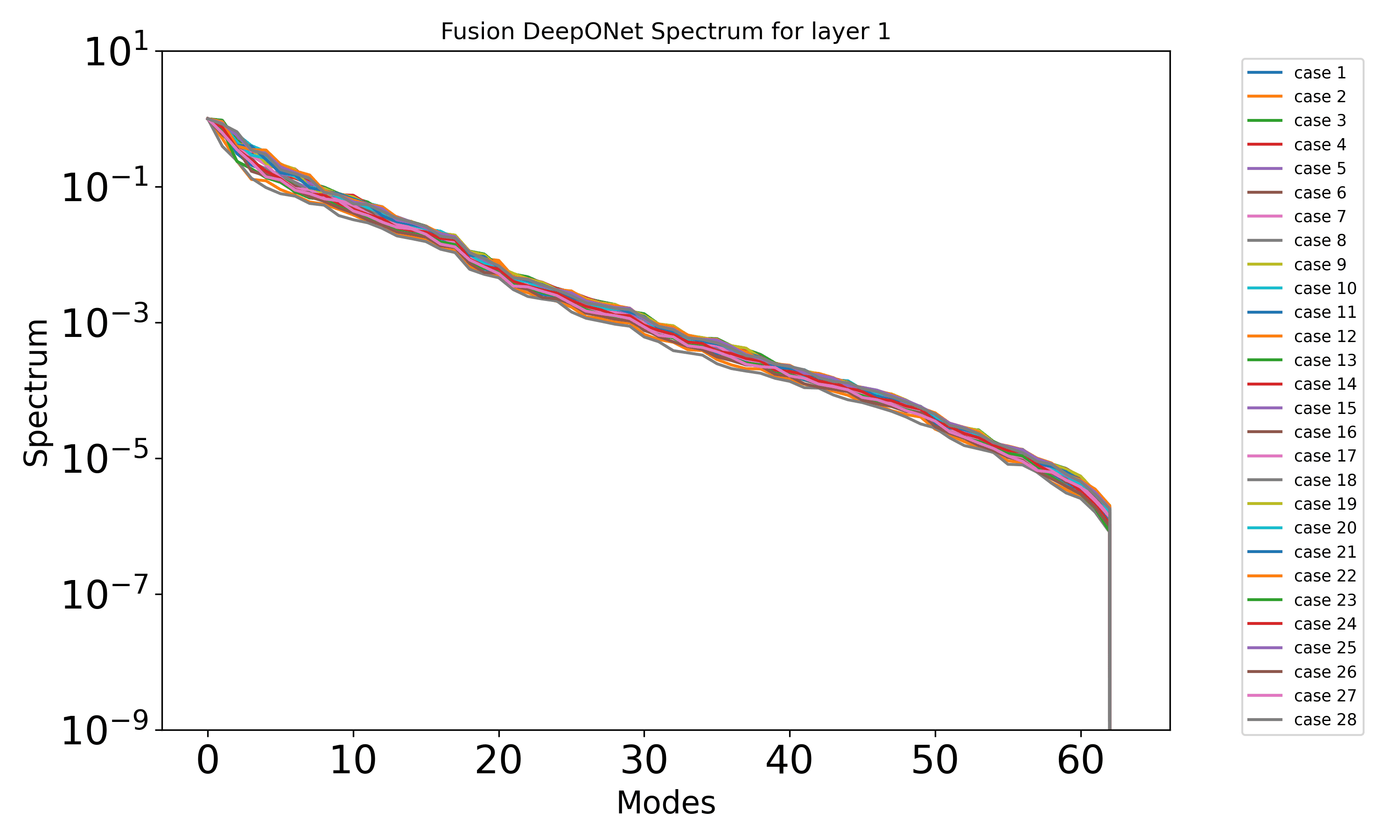}
 &
\includegraphics[width=0.3\textwidth]{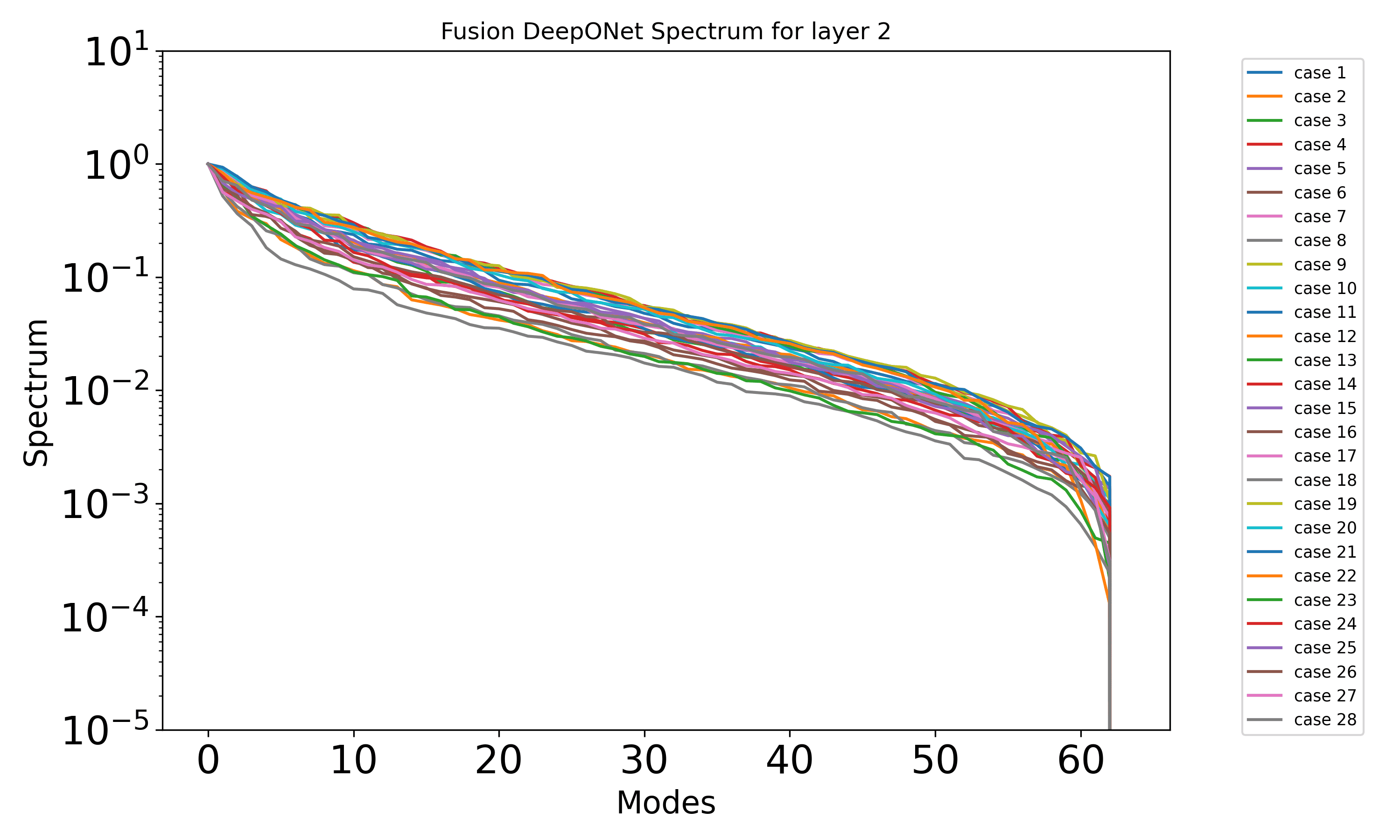}
&
\includegraphics[width=0.3\textwidth]{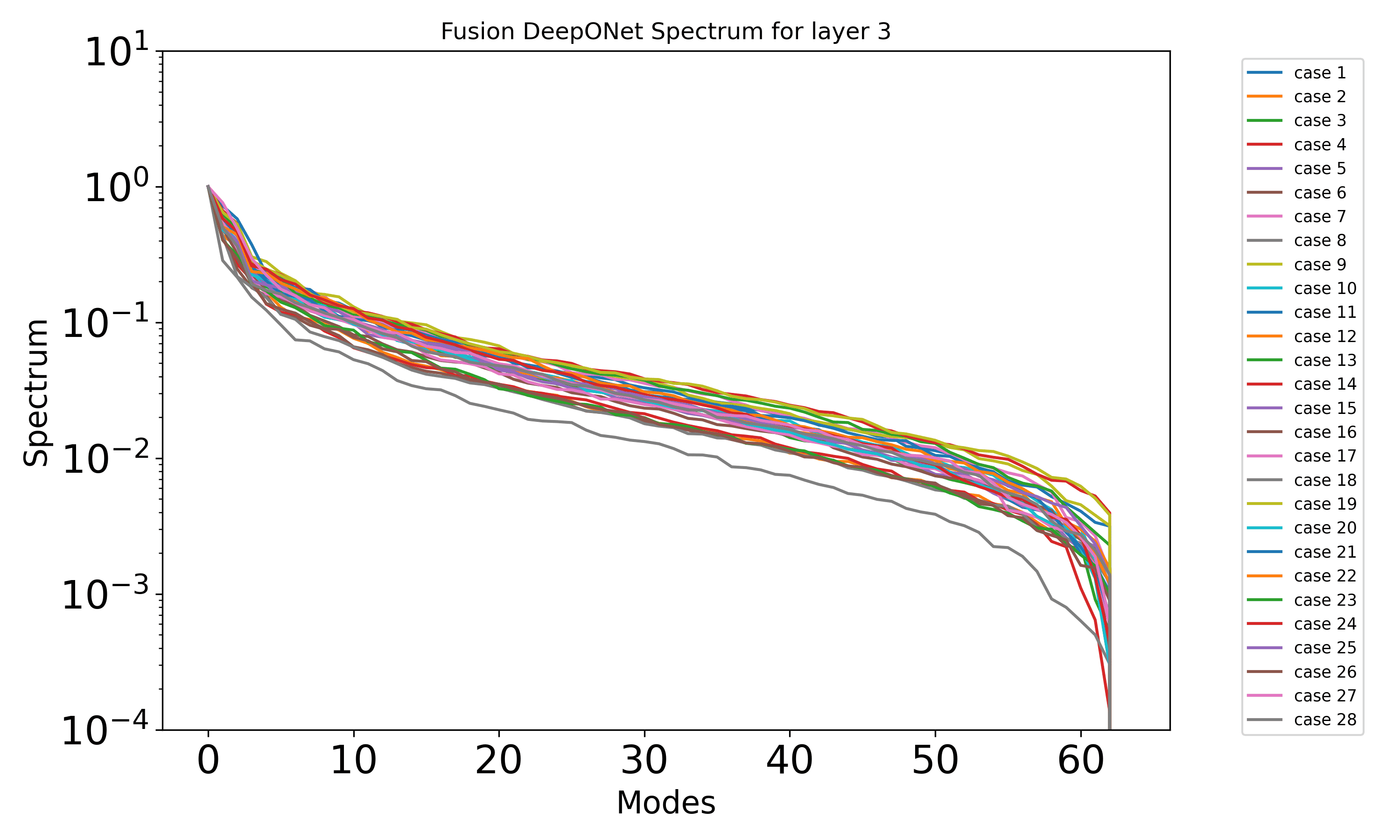}
\\
(a) FD First Layer& (b) FD Second Layer  & (c) FD Third Layer
\\
\includegraphics[width=0.3\textwidth]{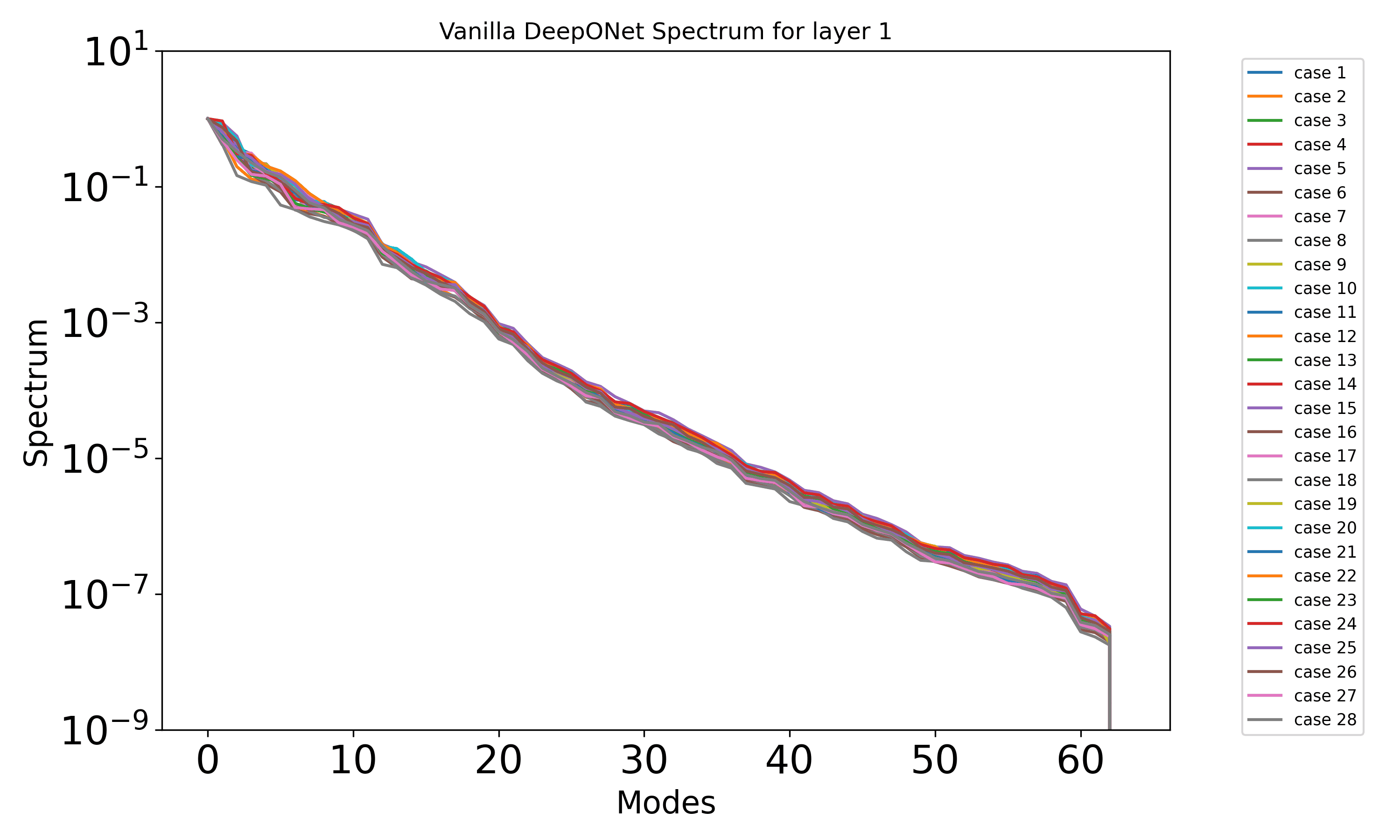}
 &
\includegraphics[width=0.3\textwidth]{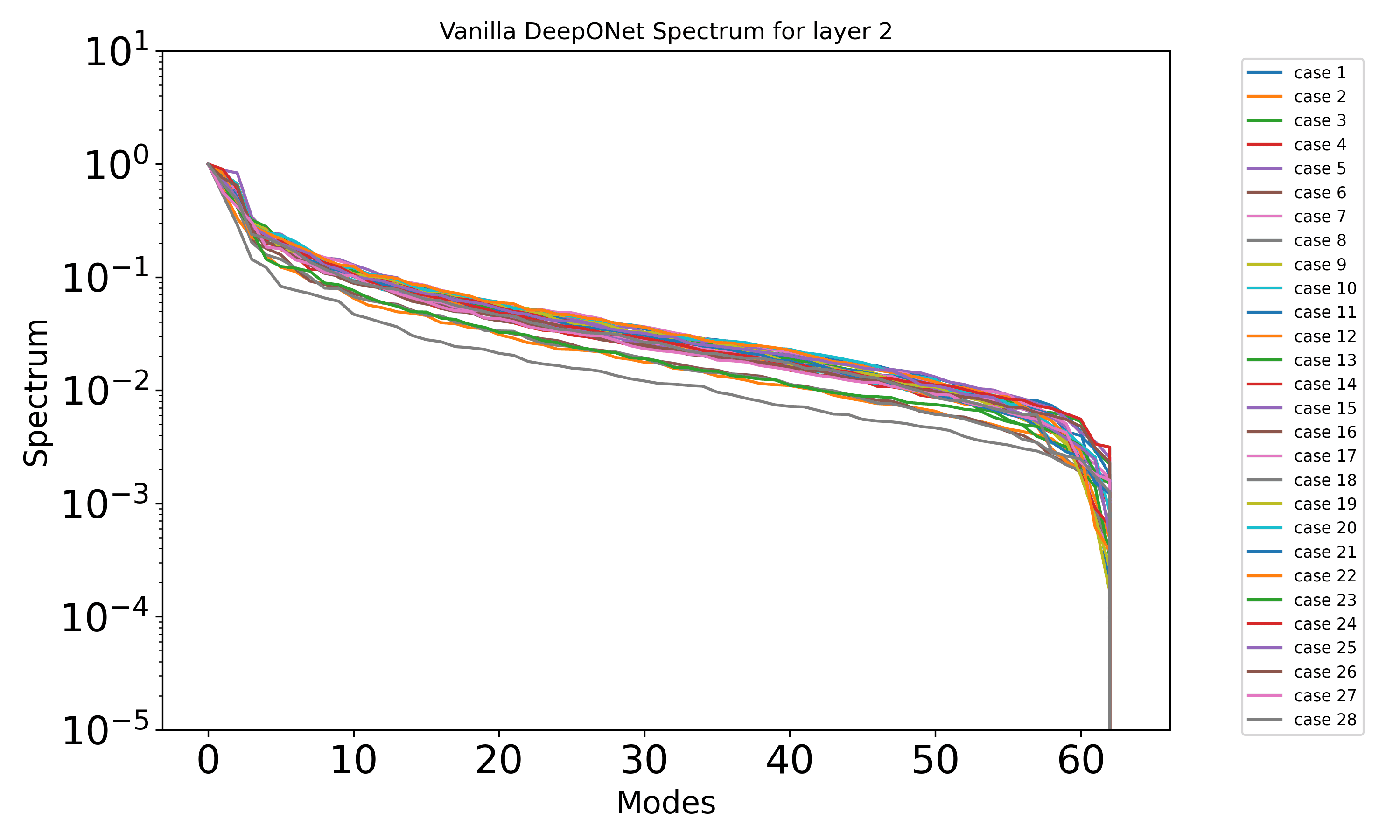}
&
\includegraphics[width=0.3\textwidth]{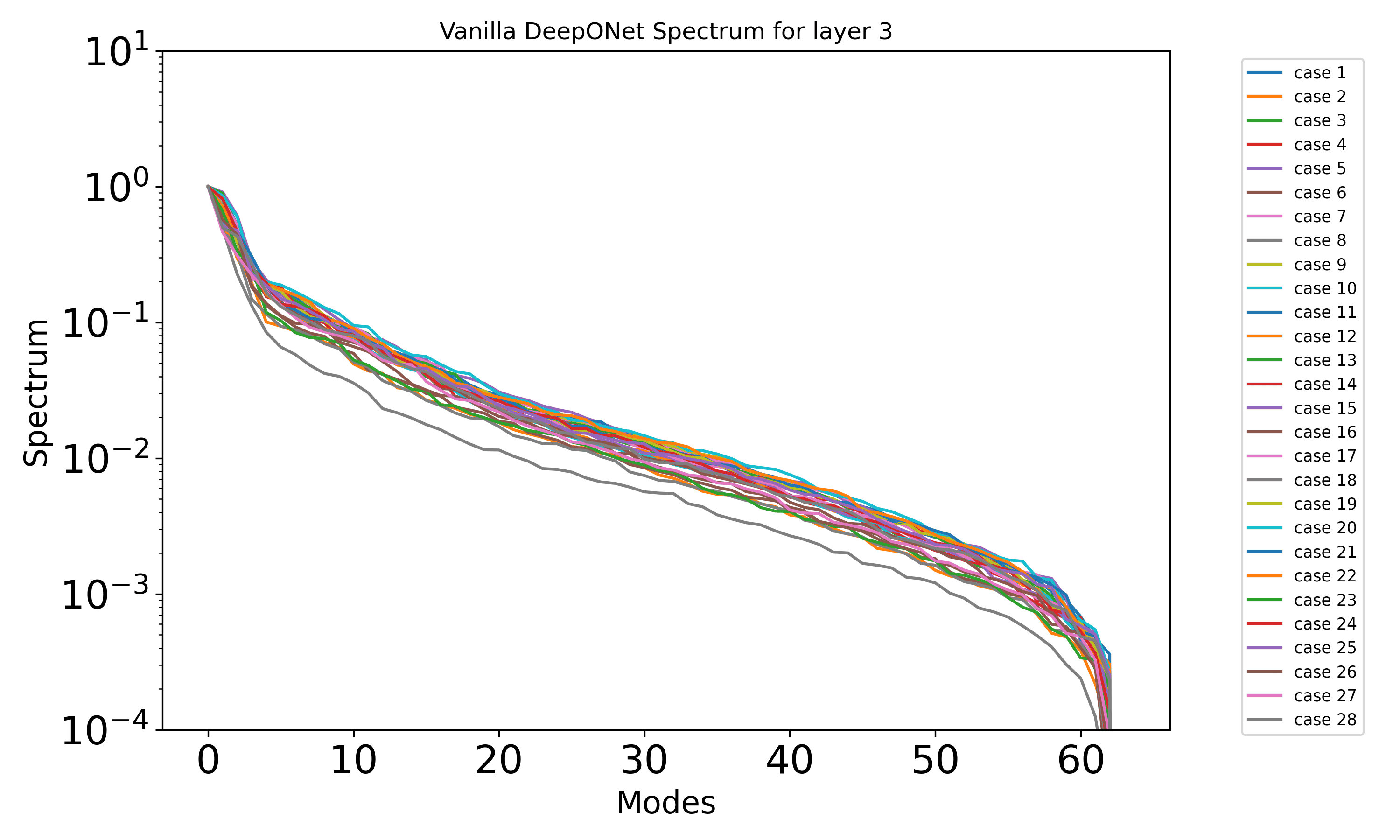}
\\
(d) VD First Layer & (e) VD Second layer  & (f) VD Third layer
\end{tabular} 
\caption{Energy Spectrum of trunk network's hidden layers outputs of Fusion-DeepONet (FD) and Vanilla-DeepONet (VD) for irregular unstructured grid. Each curve represents the Eigen spectrum of extracted data on an individual grid in the training dataset.}
    \label{fig:spectrum_ig}
  \end{center}
\end{figure}

Figure~\ref{fig:basis_ig} depicts the basis functions extracted by the last hidden layer of the Fusion-DeepONet for three grid configurations. Comparing grid configurations, we can observe that the basis functions of each mode exhibit different solution structures. At mode zero, the general shape of the actual solution can be seen for all three grid configurations. According to Fig.~\ref{fig:basis_ig}(a) and (b), mode zero for G=1 resembles the density and pressure solution while mode zero for G=13 and G=23 resembles the $v$ velocity. For each mode, the general range of the color map remains the same across the various grids. At higher modes, the basis function deforms into the outline of the bow shock. At the constant regions of the domain, we can observe that the basis functions at high modes show high-frequency oscillations, which will be canceled out in combination with other basis functions to retrieve the constant state.

\begin{figure}[t!]
  \begin{center}
    \begin{tabular}{cccc}
\includegraphics[width=0.23\textwidth]{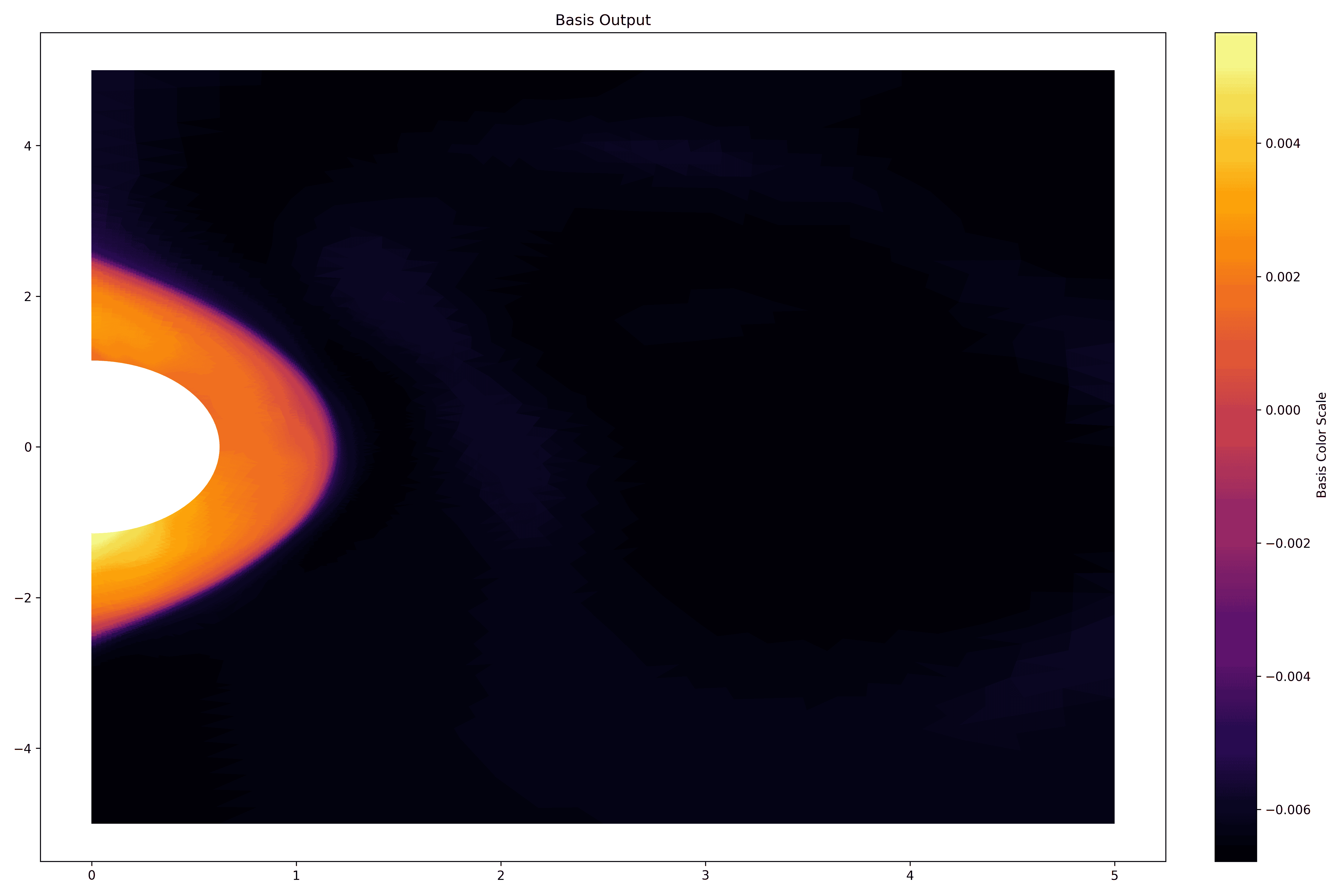}
 &
\includegraphics[width=0.23\textwidth]{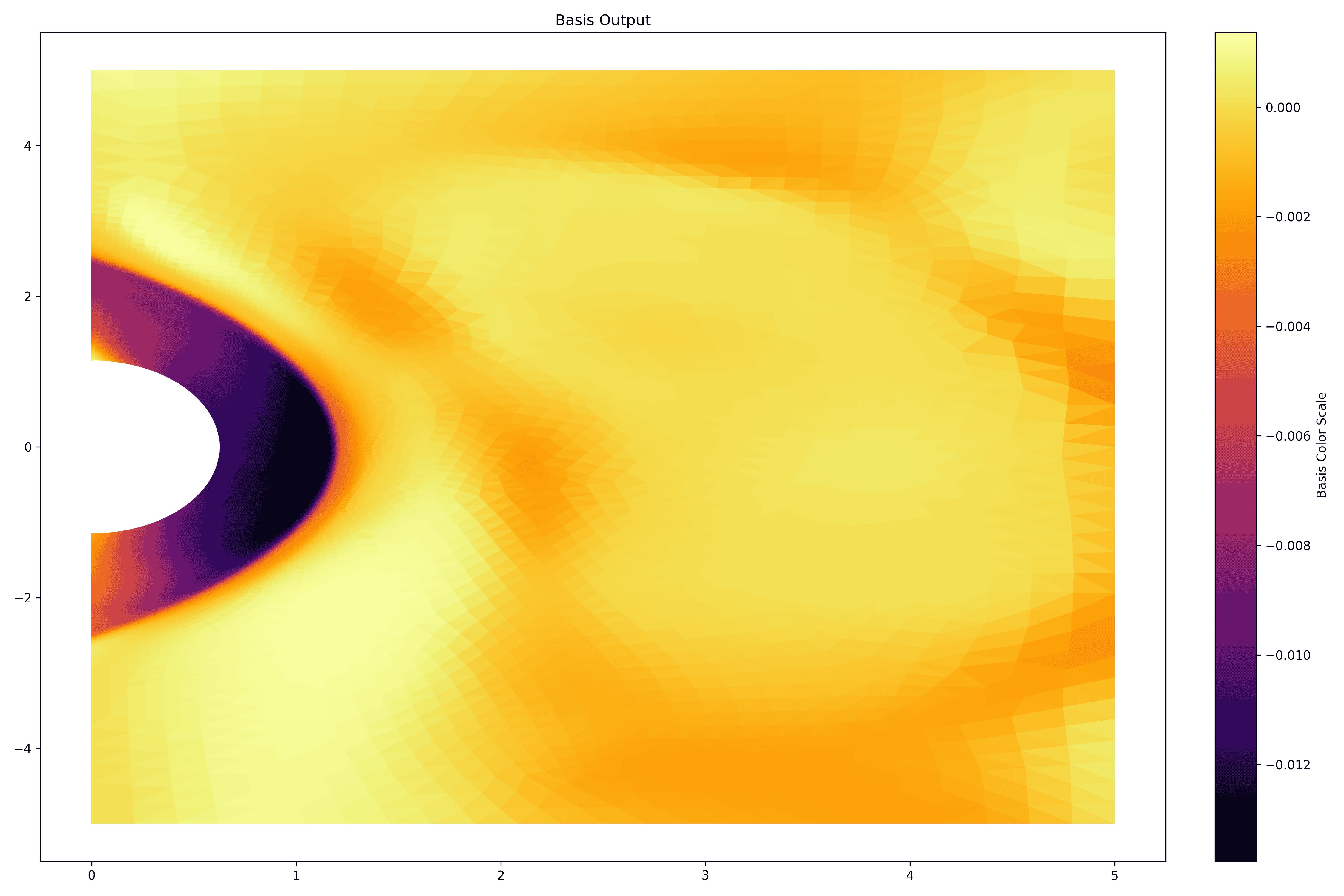}
&
\includegraphics[width=0.23\textwidth]{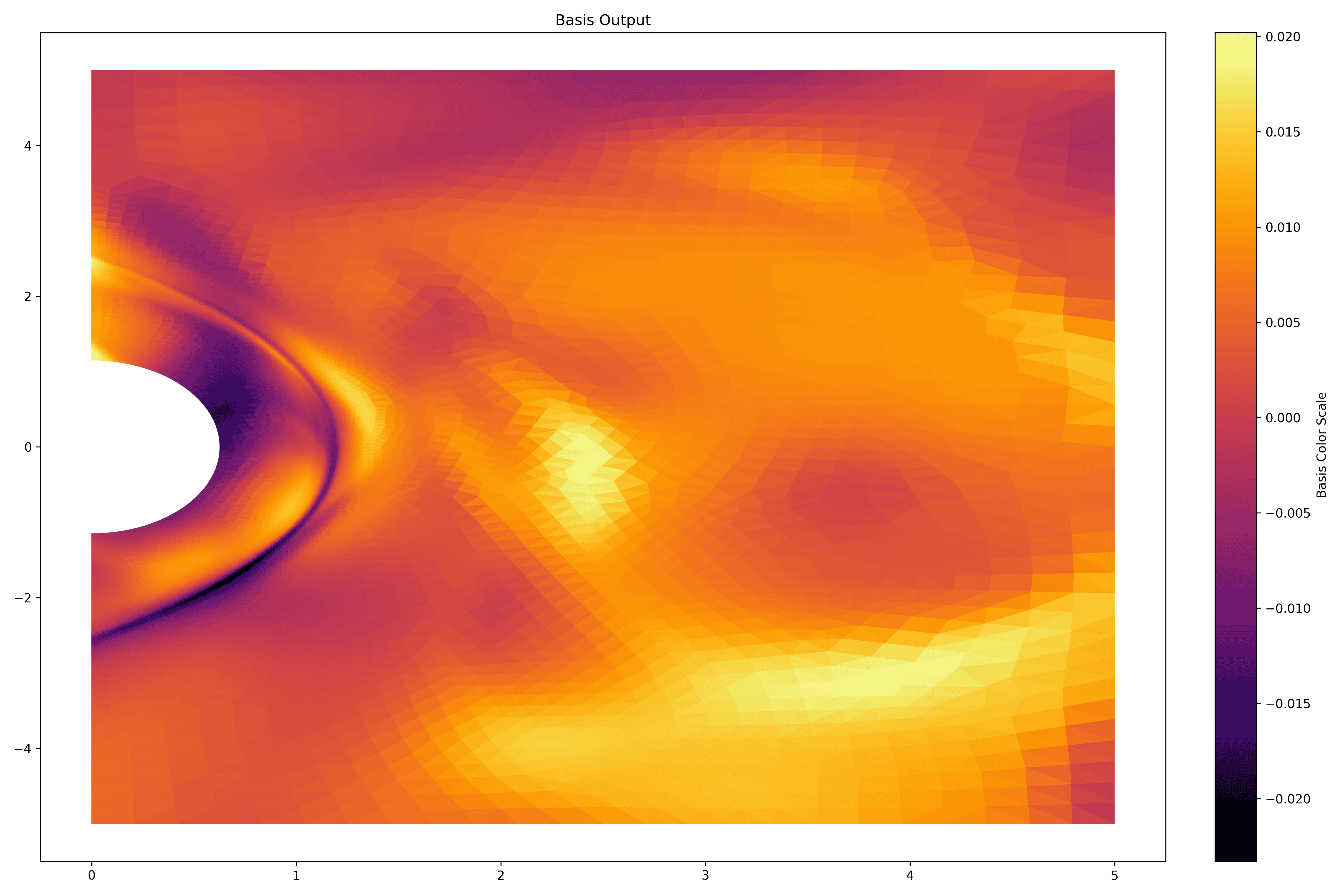}
&
\includegraphics[width=0.23\textwidth]{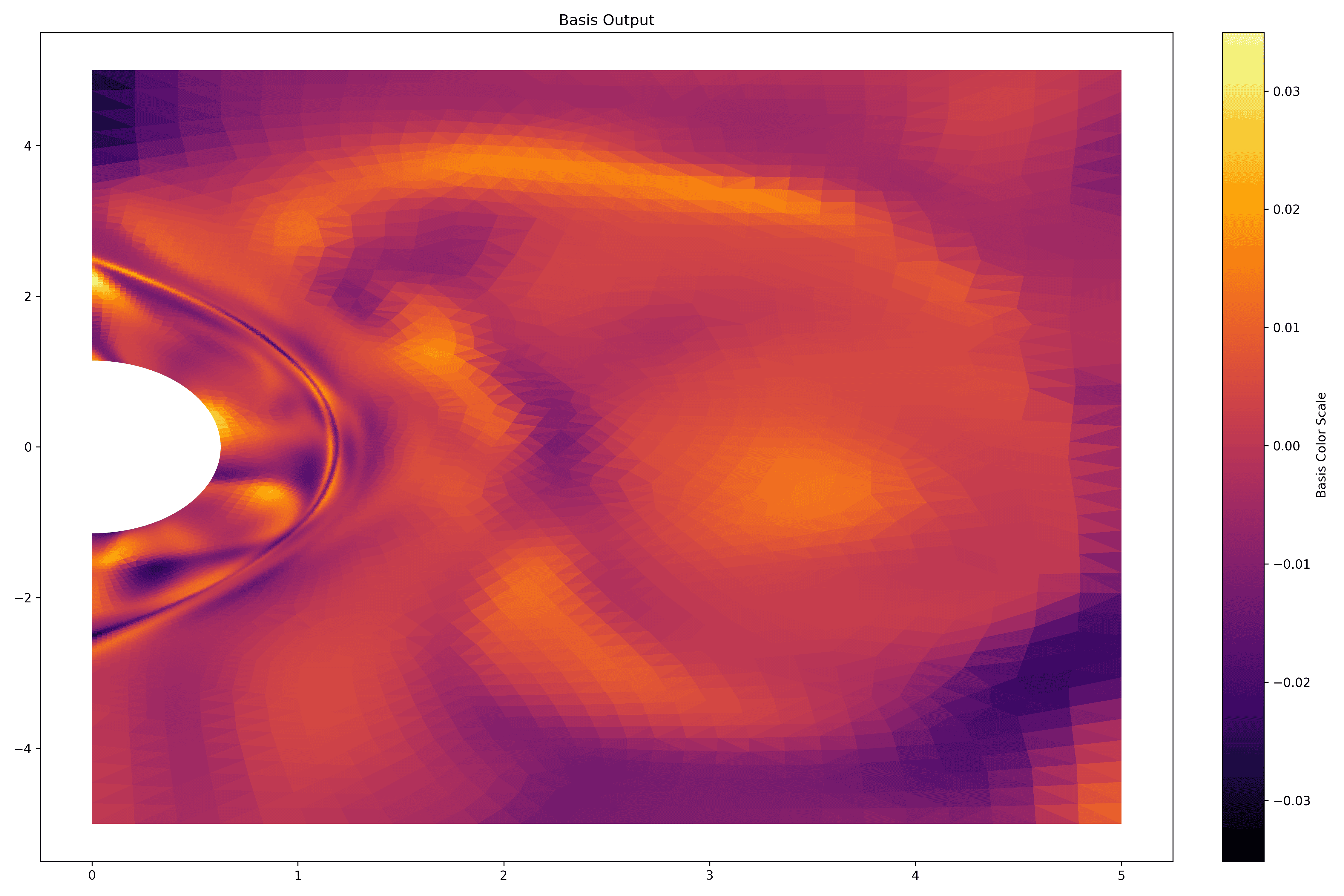}
\\
(a) G=7, M=0& (b) G=7, M=1  & (c) G=7, M=10 & (d) G=7, M=40
\\
\includegraphics[width=0.23\textwidth]{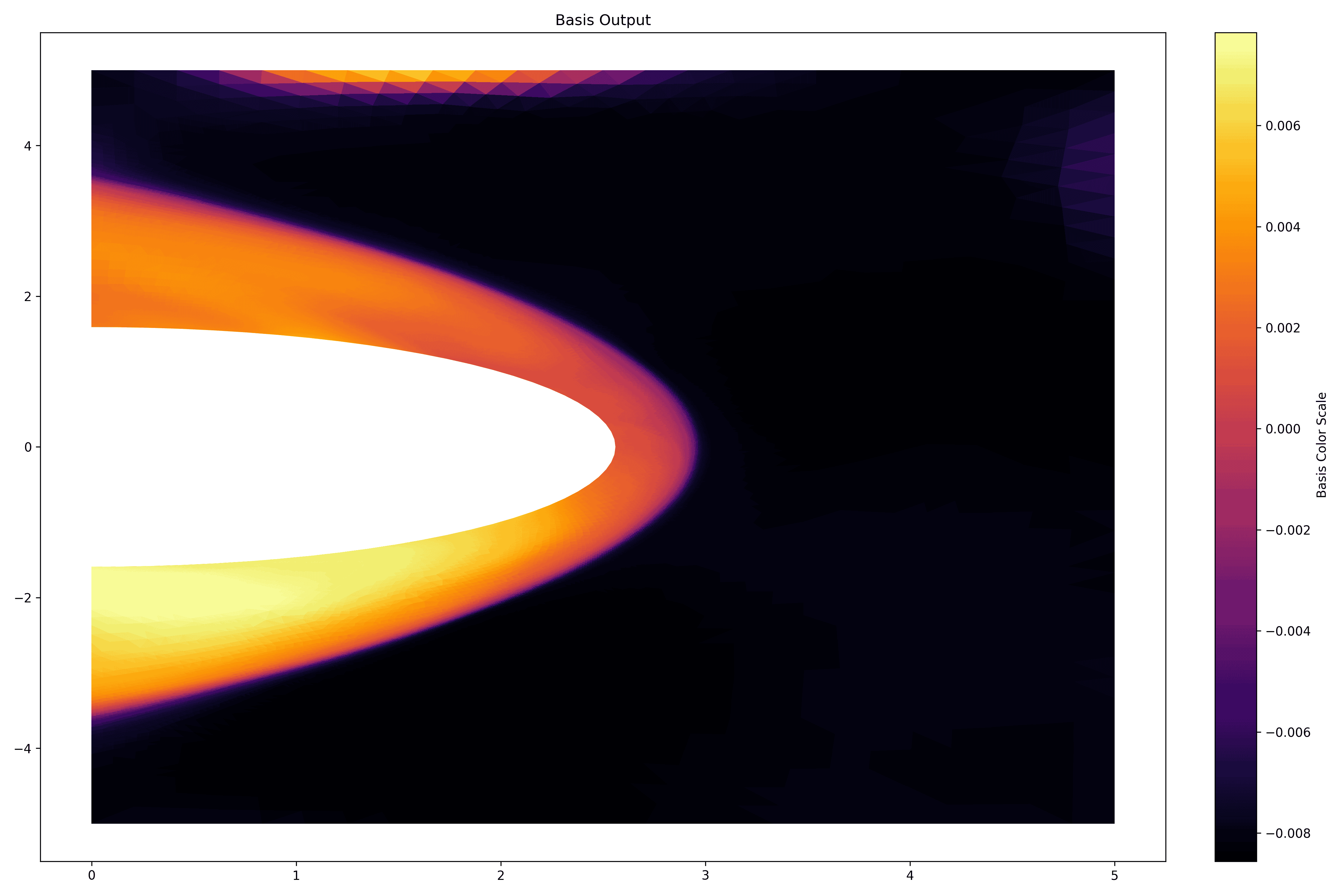}
 &
\includegraphics[width=0.23\textwidth]{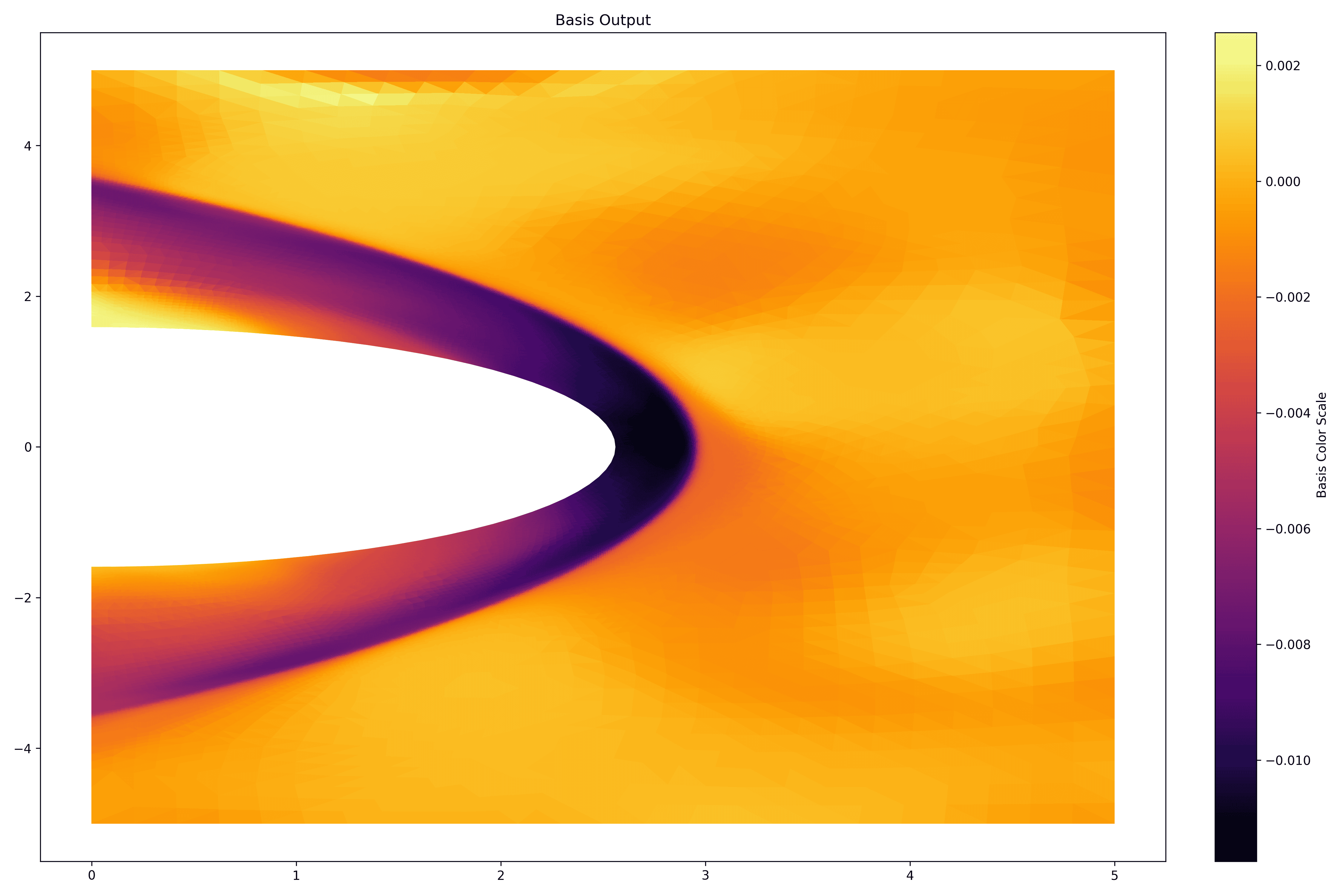}
&
\includegraphics[width=0.23\textwidth]{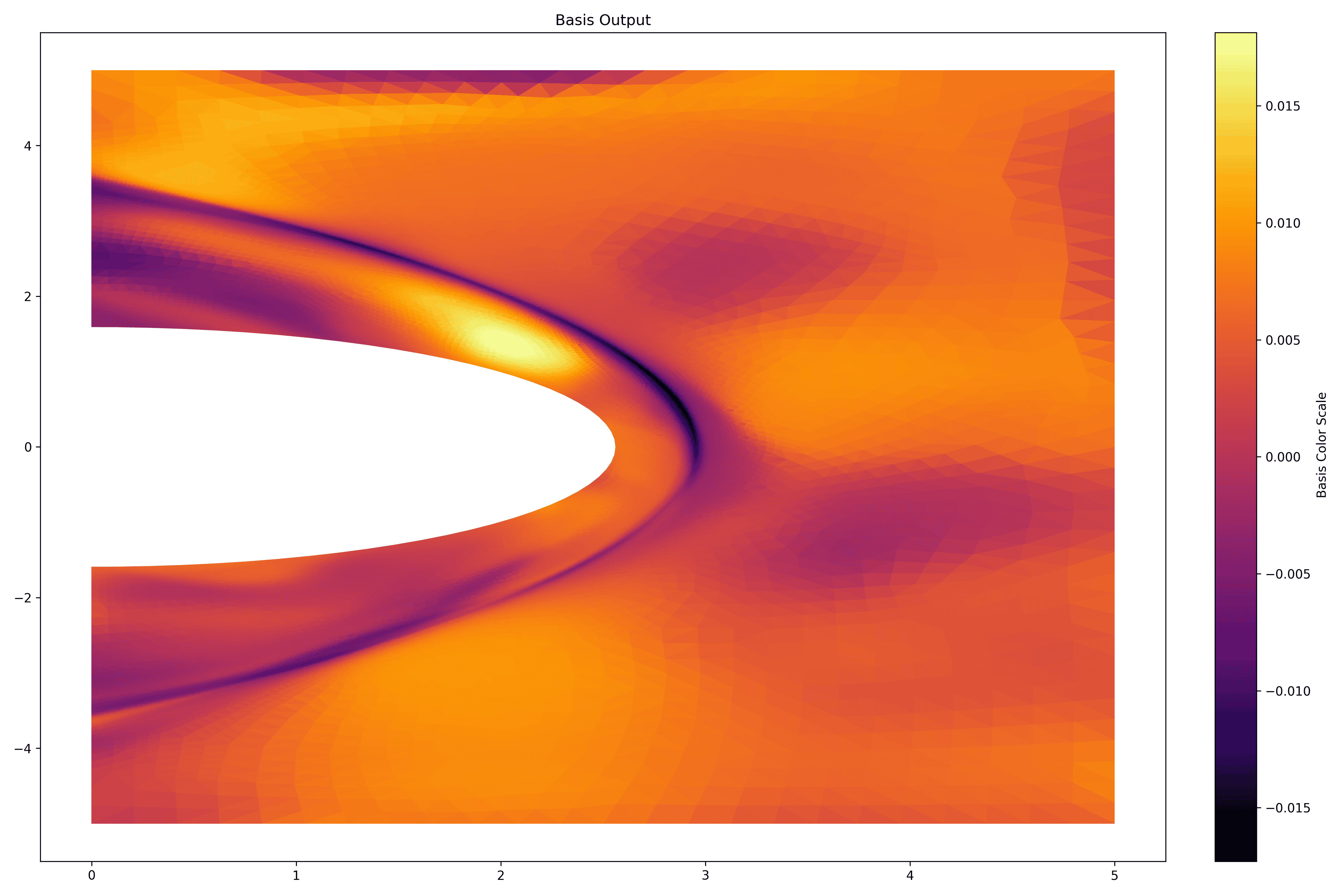}
&
\includegraphics[width=0.23\textwidth]{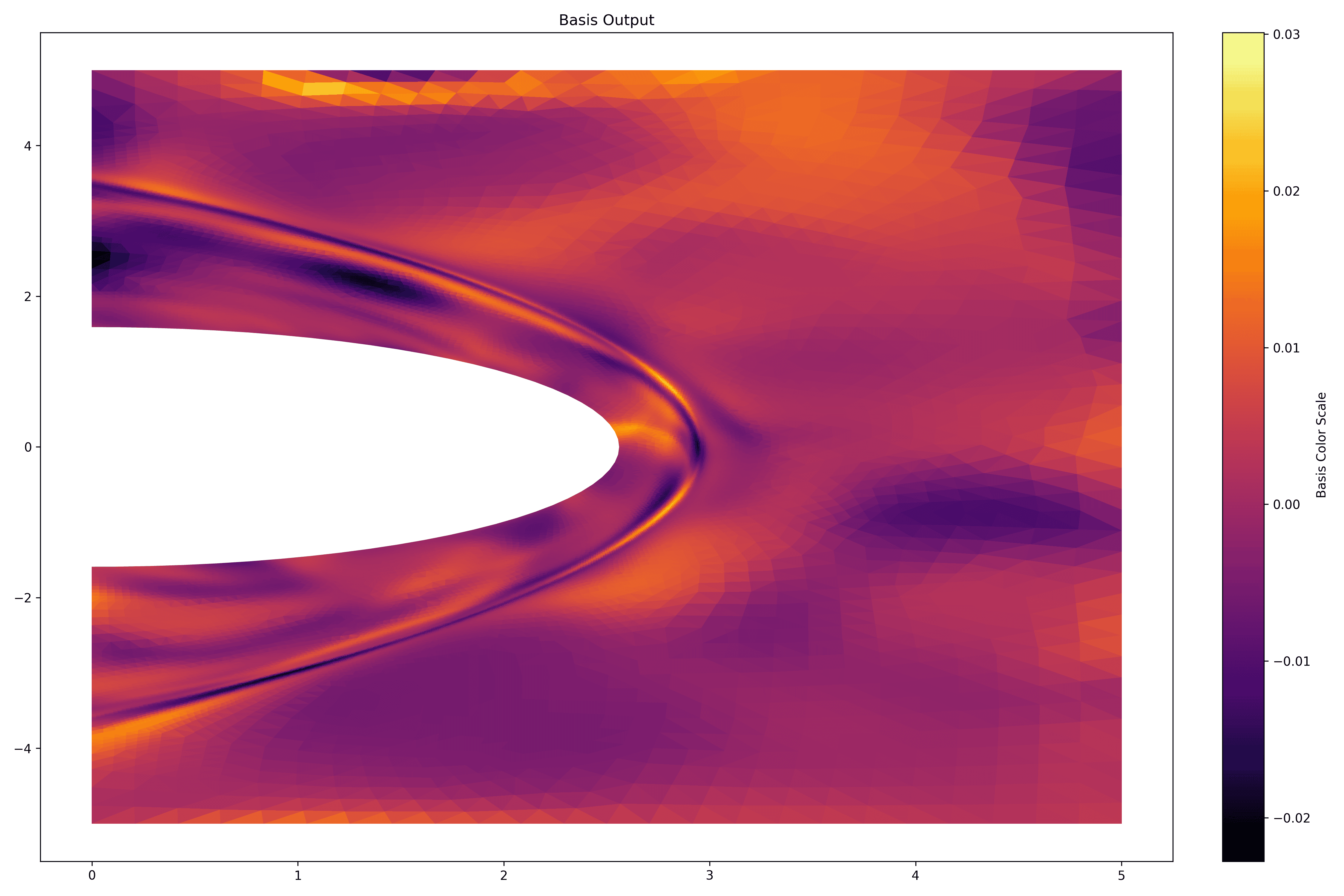}
\\
(e) G=13, M=0& (f) G=13, M=1  & (g) G=13, M=10 & (h) G=13, M=40
\\
\includegraphics[width=0.23\textwidth]{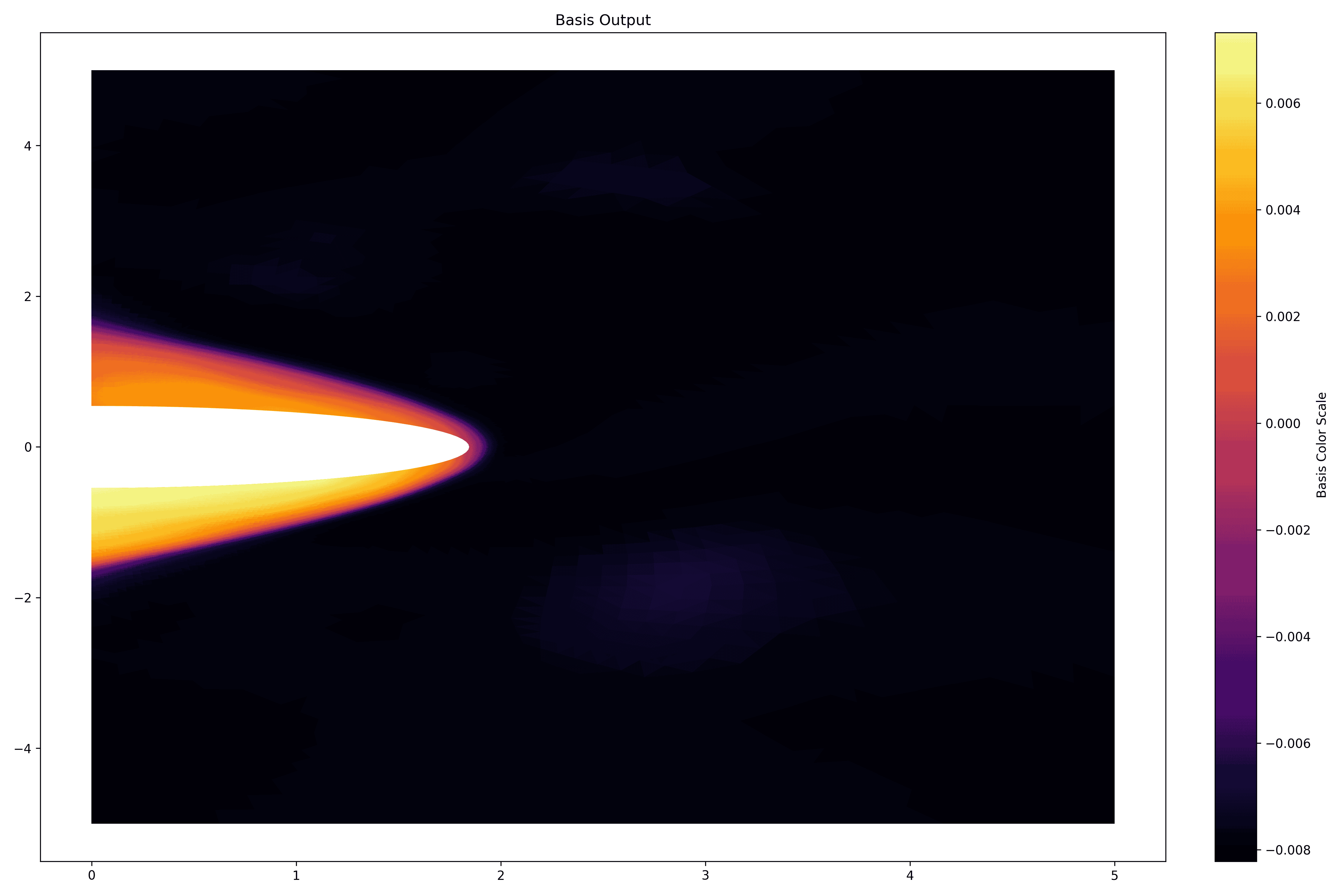}
 &
\includegraphics[width=0.23\textwidth]{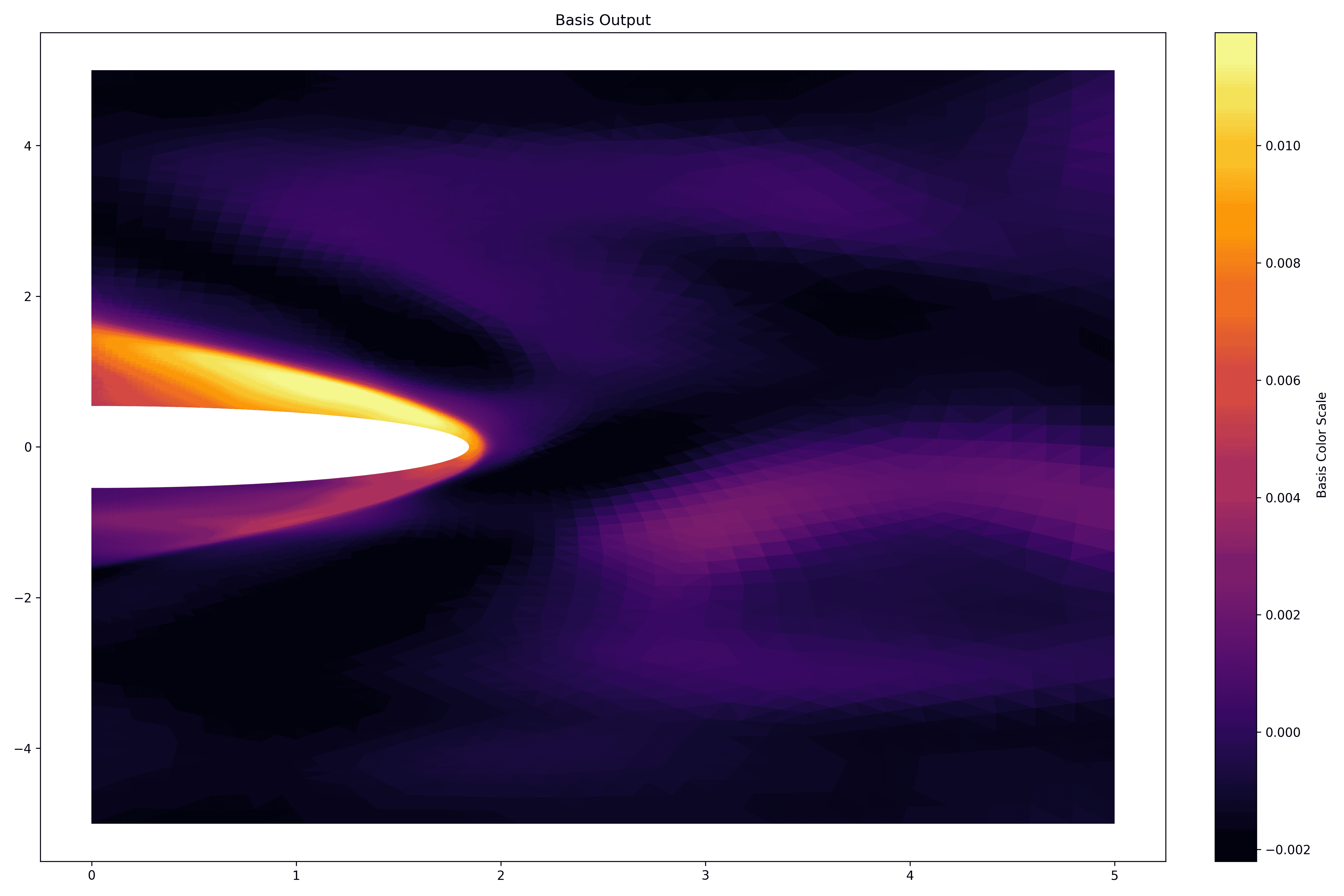}
&
\includegraphics[width=0.23\textwidth]{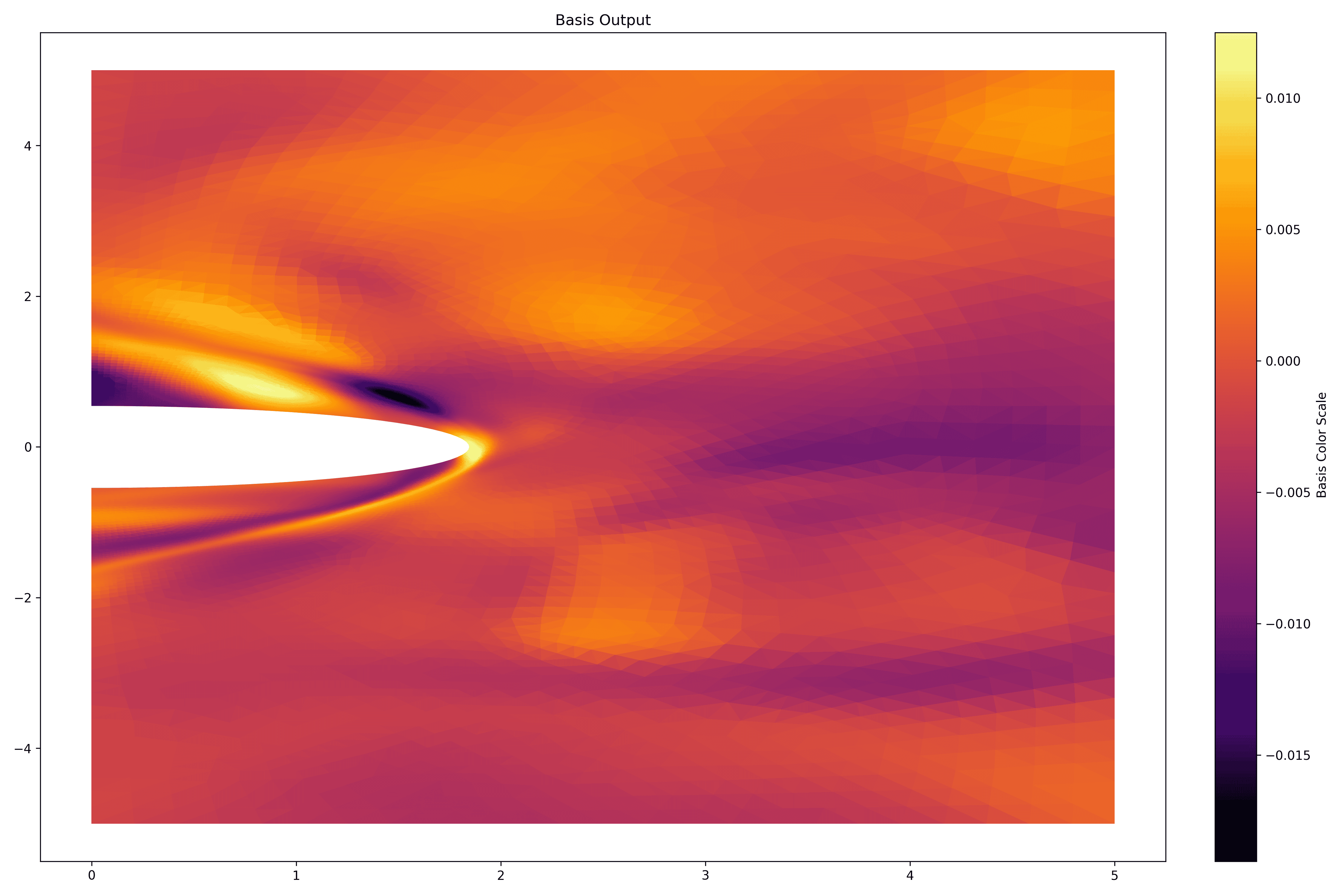}
&
\includegraphics[width=0.23\textwidth]{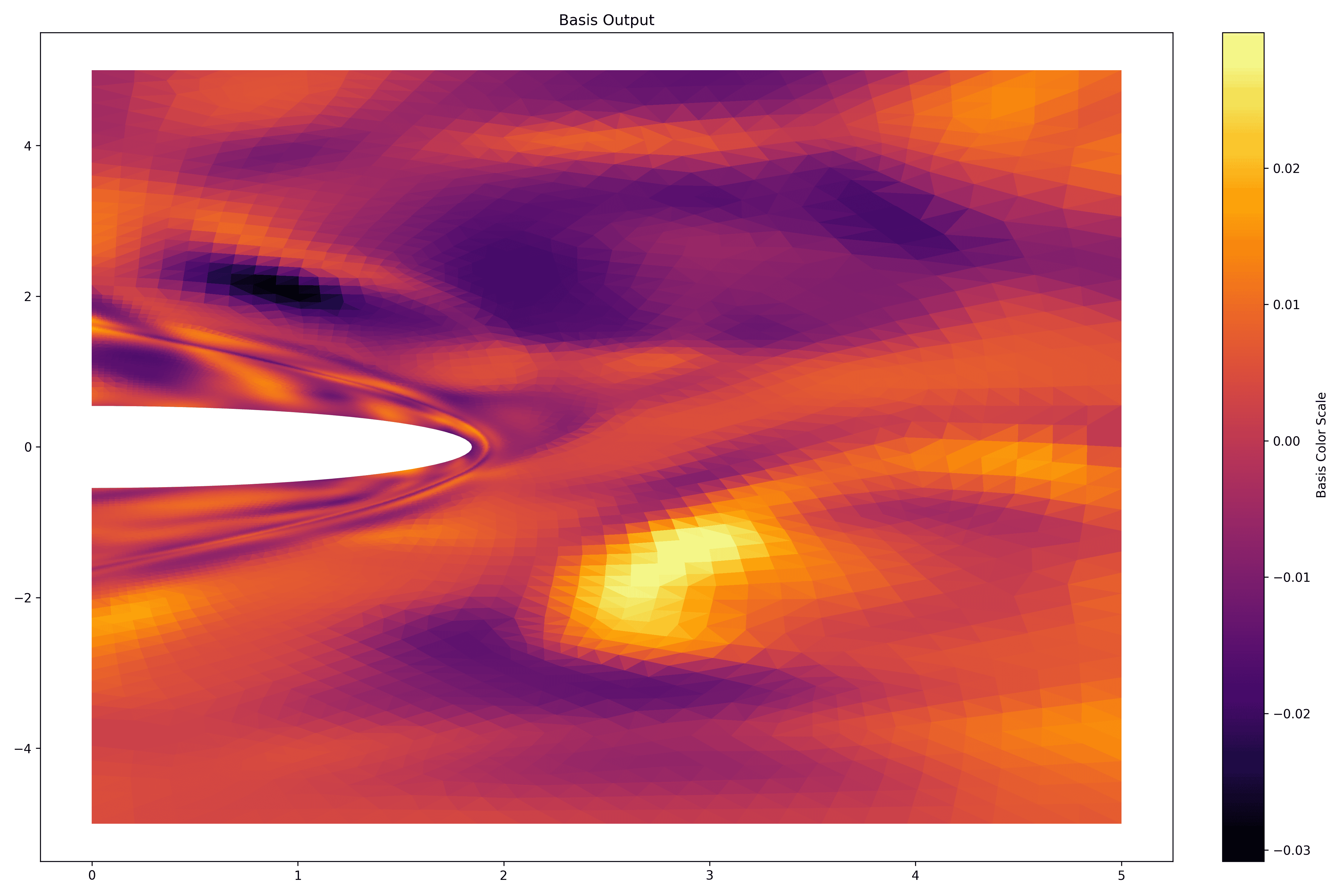}
\\
(i) G=23, M=0& (j) G=23, M=1  & (k) G=23, M=10 & (l) G=23, M=40
\end{tabular} 
\caption{Basis functions extracted by the third hidden layer of the Fusion-DeepONet trunk network. Each row belong to a specific grid configuration denoted by G. Each column is assigned to a specific mode denoted by M.}
    \label{fig:basis_ig}
  \end{center}
\end{figure}
\subsubsection{Error Distribution}
We also compare the sample-wise errors for all test conditions to understand the spread and identify outliers that may skew the error results. Figure \ref{fig:boxplot_err} shows a bar plot comparing the spread of individual sample errors for the four regular and irregular grid fields. We note that the Vanilla-DeepONet and MeshGraphNet show the highest mean errors compared to Fusion-DeepONet, POD-DeepONet, U-Net, and FNO. U-Net and Fusion-DeepONet for regular grids show the lowest error values due to their ability to extract features across multiple scales. U-Net achieves the lowest mean error values on the regular grid for $\rho$, $v$, and $p$, while Fusion-DeepONet performs better than its counterparts for the irregular grid.

\begin{figure}[!t]
\begin{center}
\includegraphics[width=\textwidth]{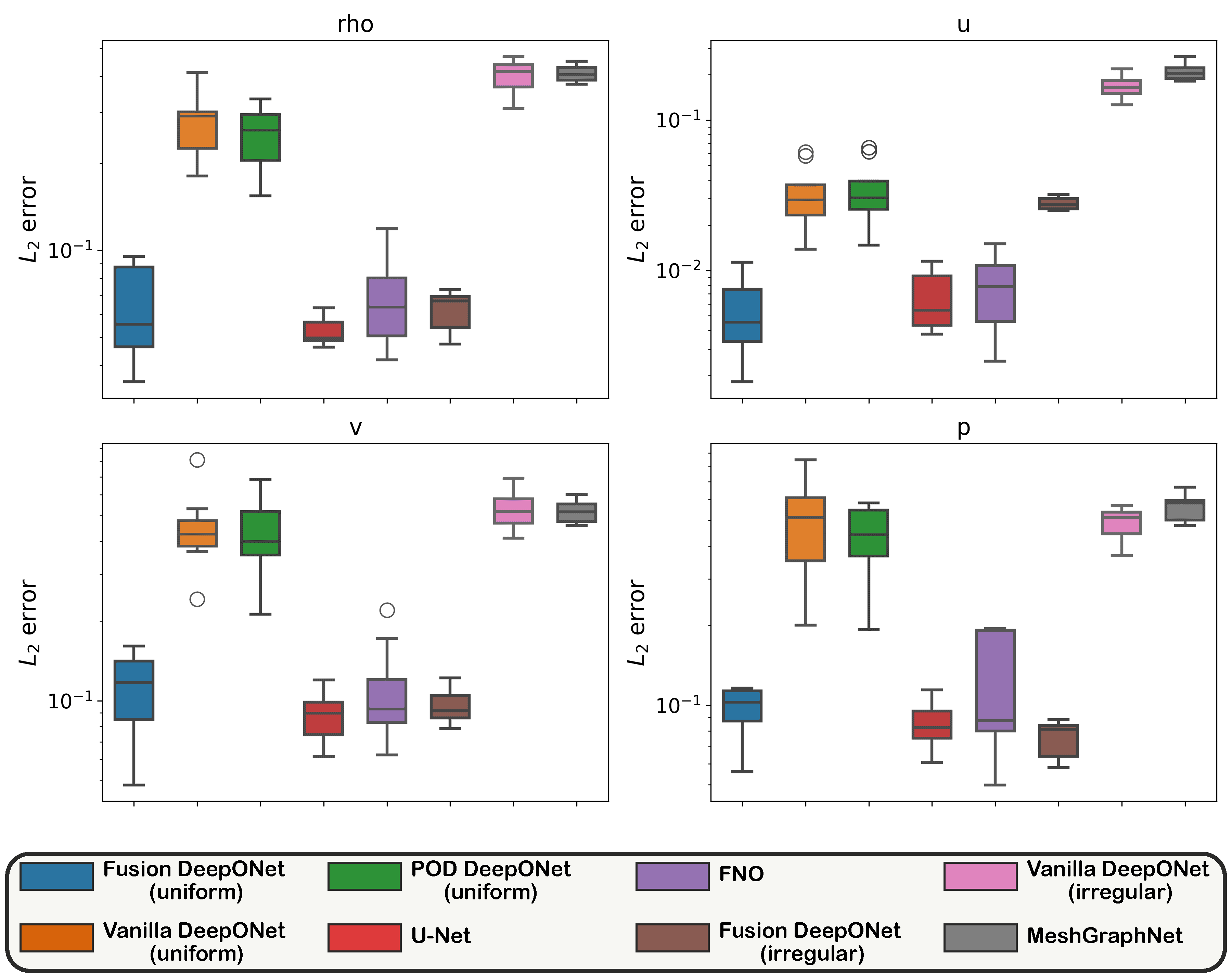}
\caption{Box plot showing sample-wise error comparison for all architectures and grid types. Vanilla-DeepONet and MeshGraphNet frameworks show highest sample-wise error distribution across both mesh types. Error values for u-velocity prediction show the lowest error across all the four outputs predicted in this study. U-Net and Fusion-DeepONet show better error values across all four output predictions. Only the Fusion-DeepONet provides a good error estimate across both regular and irregular grid setup.}
\label{fig:boxplot_err}
\end{center}
\captionsetup{justification=centering}
\end{figure}

\subsection{Hypersonic viscous flow around reentry capsule}

We develop this test case to assess the accuracy of high-speed flow surrogate models in predicting surface heat flux on reentry vehicles—a critical factor in designing thermal protection systems. Our goal is to demonstrate that Fusion-DeepONet can generate surrogates for geometry and varying flow conditions, such as Mach number. As a result, the capsule’s shape can be optimized across the whole reentry trajectory, rather than at a single fixed Mach number. From 60 total simulation runs, we randomly allocated 48 cases for training and 12 for testing. Since we focus on heat-flux prediction, Fusion-DeepONet is configured to predict only the temperature field. The trunk and branch sub-networks comprise five hidden layers of 64 neurons each, followed by a 64-neuron linear output layer. We employ the Rowdy activation function in all hidden layers and train the model using the Adam optimizer for 100,000 epochs on an NVIDIA L40S GPU. Each sample contains 77,171 unique solution points for learning the temperature distribution.

To predict the heat flux, the temperature surrogate model must accurately predict the temperature derivative with respect to the x and y coordinates. The surrogate model generated using Fusion-DeepONet is a continuous function that takes in the geometric parameters, the Mach number, plus the x and y coordinates, and generates the temperature predictions. Since we generate a continuous function, we can efficiently compute its derivative with respect to the inputs, including geometric parameters, Mach number, and coordinates, using automatic differentiation. Ultimately, we compute the heat flux by applying automatic differentiation on the trained Fusion-DeepONet to compute the heat flux on the surfaces of the reentry vehicle geometries shown in Fig~\ref{fig:caspule}. Fusion-DeepONet is trained employing two loss functions. First, we use the mean squared error (MSE) loss function constructed using the predicted temperature and the ground truth. Second, we use the loss functions defined in Eq.~\eqref{eq:loss}, which is a combination of MSE and a loss term for the derivative of the temperature field.

Our objective is to develop a surrogate model for the temperature field around a hypersonic reentry capsule that can accurately predict temperature derivatives while trained on temperature values. A key challenge is enhancing the derivative prediction capability of Fusion-DeepONet. Since we assume no access to the connectivity matrix of the solution points in the computational domain, we avoid using the derivative matrix from the DG solver that generated the training data. Instead, we aim to create an effective method even when working with unstructured point clouds lacking mesh connectivity. To address this, we explore several strategies to enrich Fusion-DeepONet with derivative information from the temperature field.



\begin{figure}[t!]
  \begin{center}
    \begin{tabular}{c}
\includegraphics[width=0.8\textwidth]{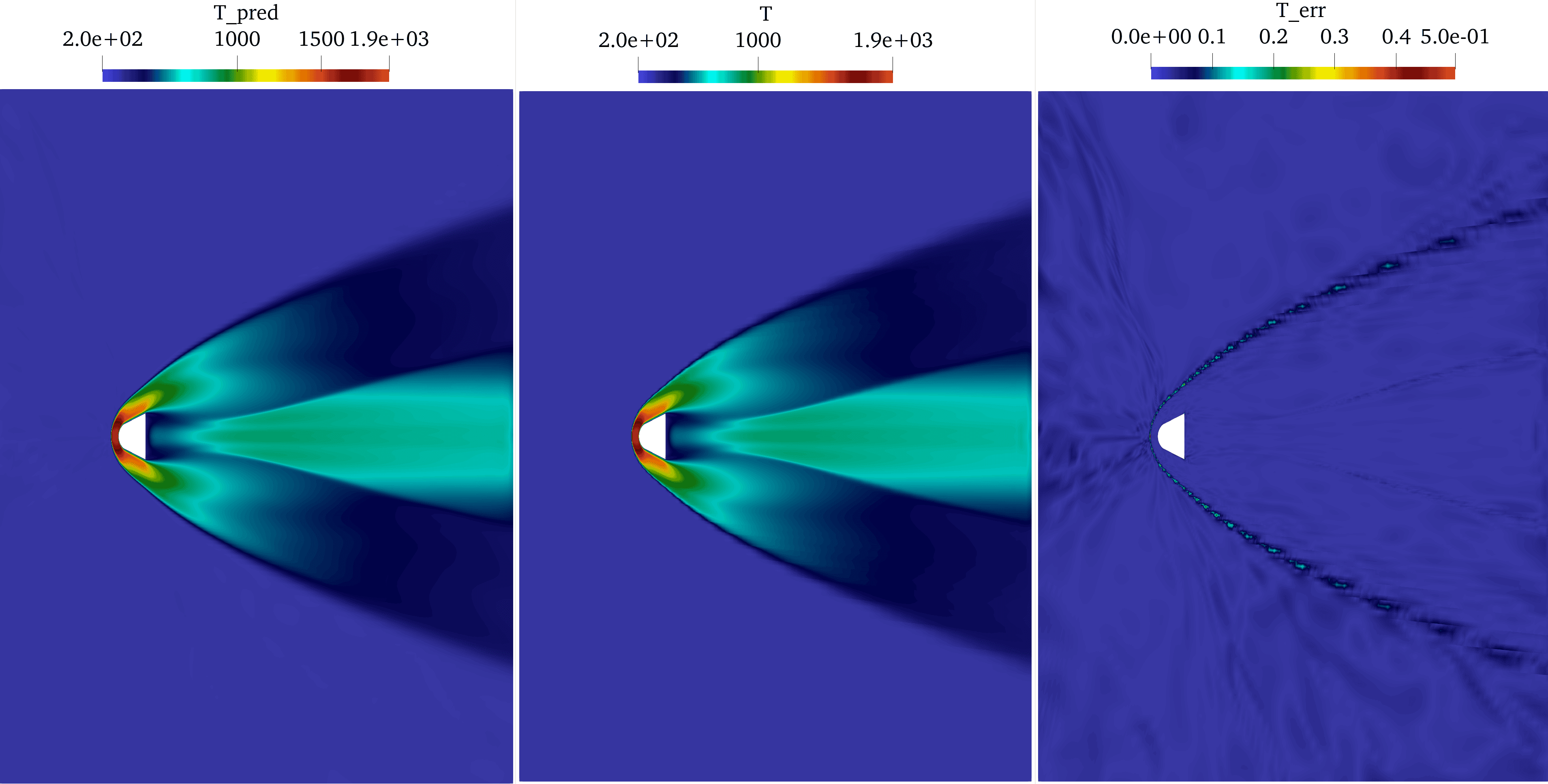}
\\
(a) Temperature field case 4  $(R,\alpha,Ma)=(14.59,44.40,6.24)$\\
\includegraphics[width=0.8\textwidth]{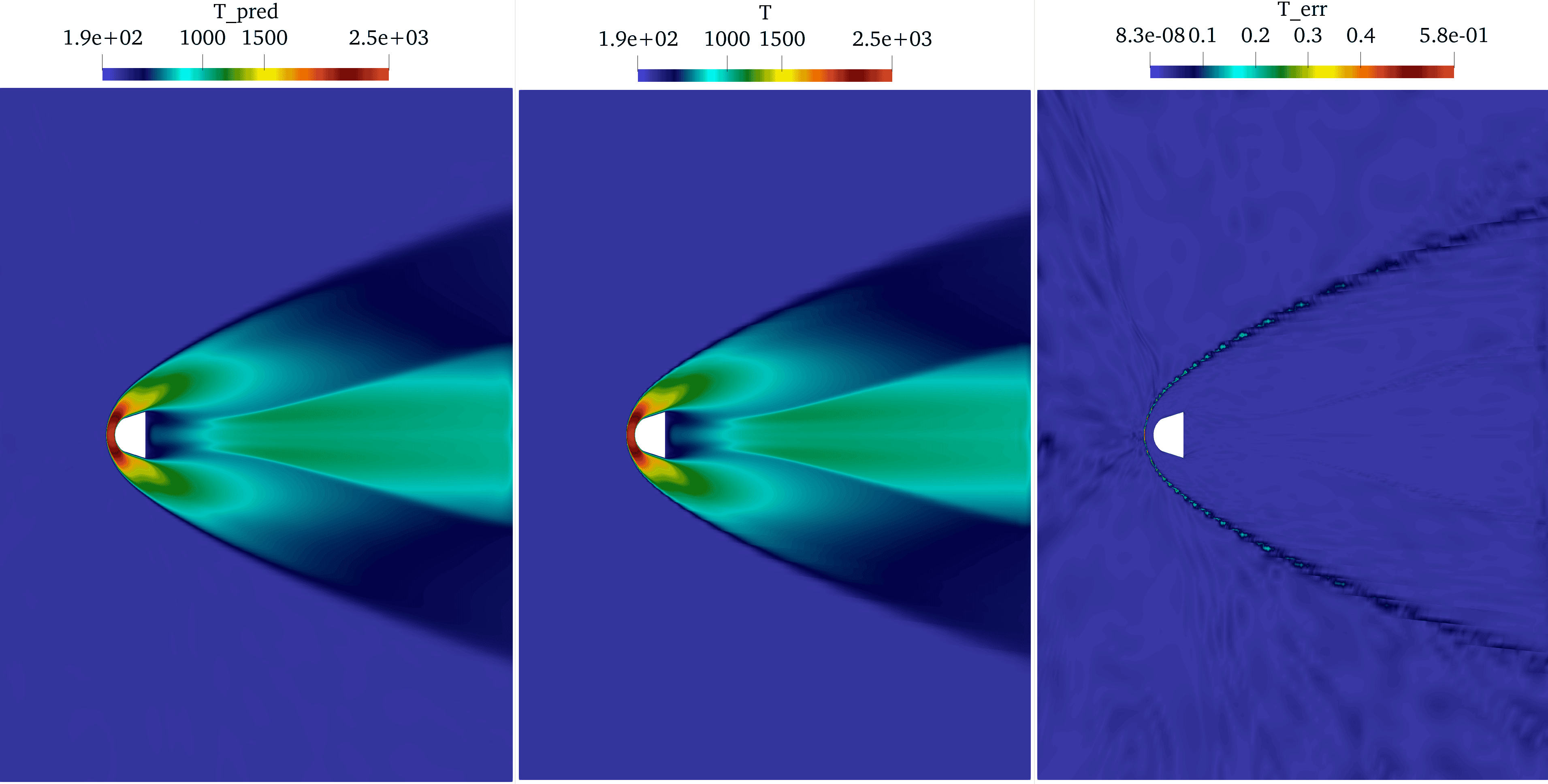}
\\ (b) Temperature field case 11 $(R,\alpha,Ma)=(17.25,54.19,7.36)$

\end{tabular} 
\caption{Two unseen test cases of flow field around hypersonic reentry vehicle predicted by Fusion-DeepONet. (a) Temperature case 4, (b) Temperature case 11. The Fusion-DeepONet was trained using 48 samples and tested for 12 unseen samples. The loss function is MSE of temperature field. from the left first contour plot is prediction, second plot is ground truth, and  third plot is the point-wise error 
 defined as $T_{err}=|T_{pred}-T|$, See appendix.~\ref{append:datasets} for the values of input parameters for test cases 4 and 11.}
    \label{fig:capsule_no_der}
  \end{center}
\end{figure}

Figure~\ref{fig:capsule_no_der} presents the temperature field predictions made by Fusion-DeepONet. In each set of figures, the left plot illustrates the predicted temperature field, the middle plot represents the ground truth, and the right plot shows the point-wise prediction error calculated as $|T_{\text{pred}} - T|$. The results demonstrate that Fusion-DeepONet achieves highly accurate temperature predictions for previously unseen cases, highlighting its excellent generalization capability across varying geometries and Mach numbers. The point-wise errors remain small throughout most of the domain, although the maximum errors are observed near the bow shock region. The average relative $L_2$ norm of the temperature error across the 12 test cases is approximately $0.98\%$. However, as discussed later, despite the temperature field being predicted with less than $1\%$ error, significant errors emerge when computing its spatial derivatives with respect to the $x$ and $y$ coordinates.


Now we focus on the correct prediction of the heat flux on the surface of the reentry vehicle. The heat flux can be determined using the following 

\begin{equation}
Q = \int_{\partial \Omega} -\kappa \nabla T. \Vec{n}d\mathbf{s}
    \label{eq:heat_flux}
\end{equation}
where $\nabla T=(\partial T/\partial x, \partial T/ \partial y)$, $\Vec{n}$ is the normal vector to the line segment $d\mathbf{s}$. $\partial \Omega$ denotes the surfaces of the reentry capsule. The surfaces of the reentry capsule are segmented using the sides of quadrilateral elements placed on the boundary of the capsule. According to the discretization of the DGSEM solver, on each line segment, we have $P+1$ nodes which we use to approximate the integral of Eq.~\eqref{eq:heat_flux}. In the simulations, we employed $P=3$, which results in four points per line segment. For each line segment, we used the trapezoid rule using the four nodes, the normal vector values to the line segment, and the value of $\nabla T$ computed at four nodes to compute the heat flux. After the computation of heat flux of line segments, we sum the heat flux of the line segments on the surface of the reentry capsule to compute the total heat flux.

\subsubsection{Heat flux prediction}
In Section~\ref{method_der}, we introduced two methods to enhance the accuracy of derivative predictions by Fusion-DeepONet:
\begin{enumerate}
\item Discrete Derivative (DD) Method: Use the discrete derivative operator defined in Eq.\eqref{eq:approx_der} to compute derivatives of both the Fusion-DeepONet predictions and the ground truth, as incorporated in the derivative loss term in Eq.\eqref{eq:loss}.
\item Least Squares Differentiation (LSD) Method: Estimate directional derivatives of both the Fusion-DeepONet predictions and the ground truth using the LSD method, and include them in the derivative loss term in Eq.~\eqref{eq:loss}.
\end{enumerate}
We further investigate their accuracy in predicting the temperature field and the temperature derivative without explicitly training DeepONet for the derivative field. Here, we clarify that the ground truth data is generated using the discontinuous Galerkin code, which produces a discontinuous solution at the interfaces of the quadrilateral elements. Therefore, when we compute the derivative of this temperature field even in the ground truth, we obtain discrete derivative contours across elements. To diminish the discontinuous jumps in the temperature derivative for the ground truth, we use spectral element interpolation to interpolate the solution on $P=3$ elements ($4\times4$ nodes in each quad) to $P=7$ elements ($8\times8$ nodes in each quad). Therefore, we generate a higher resolution grid without simulating the cases using Trixi.jl for higher $P$. During training, we employ $P=3$ resolution coordinates, while for inference,  we employ both $P=7$ and $P=3$ for heat flux predictions.

\begin{figure}[t!]
  \begin{center}
    \begin{tabular}{cc}
\includegraphics[width=0.48\textwidth]{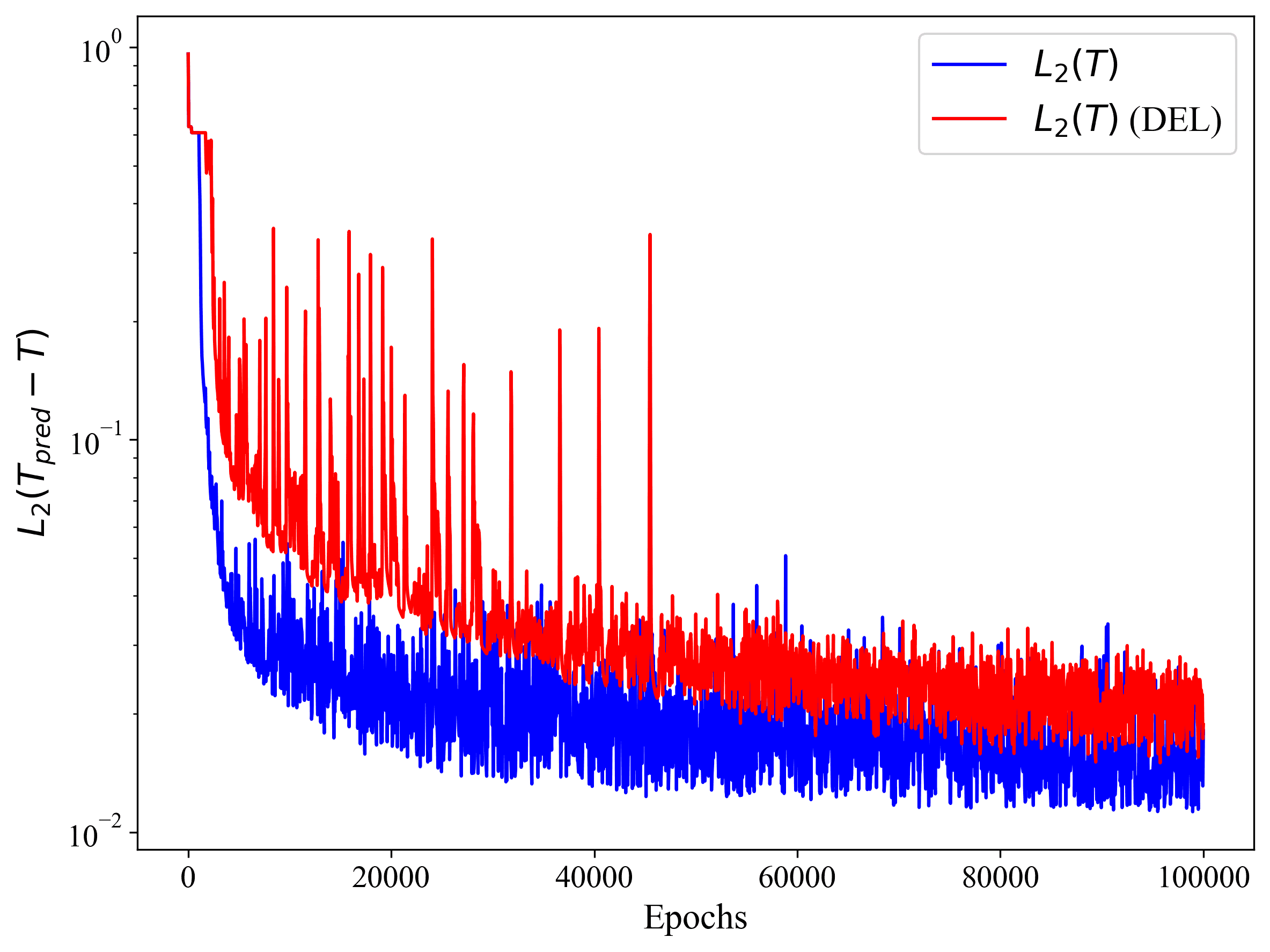} & \includegraphics[width=0.48\textwidth]{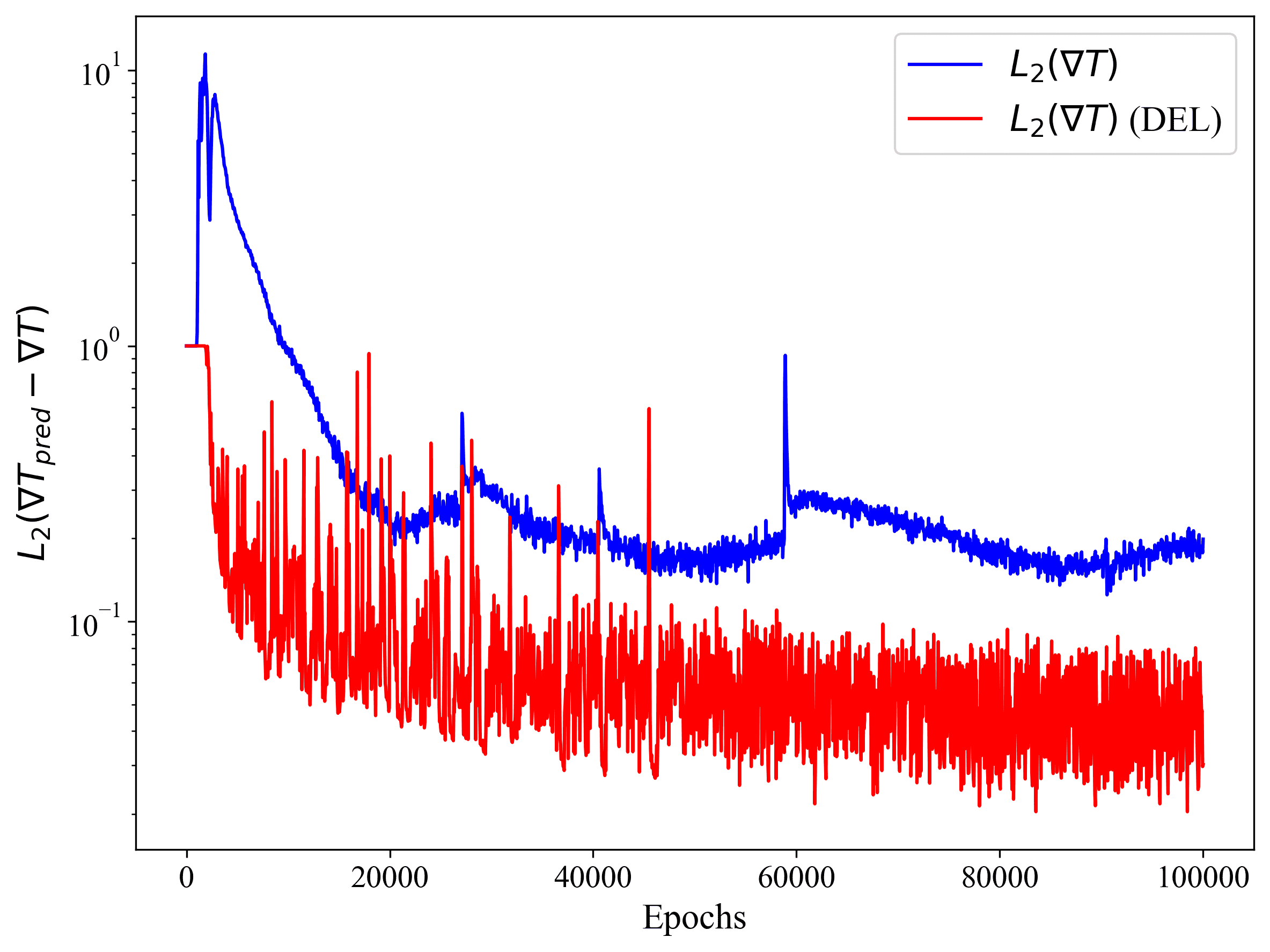}
\\
(a) $||T-T_{pred}||_{L_2}$ & (b) $|||\nabla T|-|\nabla T_{pred}|||_{L_2}$ 
\end{tabular} 
\caption{(a) $L^2$ norm of temperature predictions for testing dataset during the training of Fusion-DeepONet (b)$L^2$ norm of temperature gradient magnitude computed using least square differentiation operator for testing dataset during the training of Fusion-DeepONet. (DEL) refers to derivative enhanced loss function.}
    \label{fig:capsule_metrics}
  \end{center}
\end{figure}

In Fig.~\ref{fig:capsule_no_der}, we show that Fusion-DeepONet can precisely learn the temperature field. However, predicting the temperature field accurately does not guarantee the accuracy of $\partial T/\partial x$ and $\partial T/\partial y$. Figure~\ref{fig:capsule_metrics} (a) compares the relative $L_2$ norm of temperature prediction error for the testing dataset during training for the MSE loss function and the derivative-enhanced loss function (DEL). Moreover, Fig.~\ref{fig:capsule_metrics} (b) compares the trend of the $L_2$ norm of the magnitude of the temperature gradient computed using the least squares differentiation operator. According to Fig.~\ref{fig:capsule_metrics}, the DEL loss function slightly deteriorates the temperature prediction while improving the temperature gradient prediction. We conclude that integrating a derivative loss term can enhance the prediction of the field gradient and, consequently, the heat flux.

\begin{table}[!t]
\caption{Heat flux prediction accuracy over the unseen dataset using various derivative operators. $\%E$ is average relative error of heat flux predictions over the entire testing dataset. $\%L^2(T)$ is the relative $L^2$ norm of temperature field for all the test cases. The temperature derivative computation during the inference is performed using automatic differentiation for all the cases.}
\label{tab:result_error_hf}
\begin{center}
\resizebox{\columnwidth}{!}{%
\begin{tabular}{@{}lccccc@{}}
\toprule
\textbf{Approach} &  $\%E(Q)$, $P=3$&$\%E(Q)$, $P=7$&$\%L^2(T)$, $P=3$&$\%L^2(T)$, $P=7$  & time(h)\\ \midrule
NO DEL&34.15&27.69& 0.98&0.95&8.67\\
LSD&21.95&14.10&1.30&1.33&9.39\\
DD&$\mathbf{6.47}$&$\mathbf{11.34}$&1.34&1.31&8.00\\
\bottomrule
\end{tabular}
}
\end{center}
\end{table}
The derivative loss term can be implemented using two distinct approaches. Table~\ref{tab:result_error_hf} compares the accuracy of heat flux and temperature predictions obtained using various loss functions. We computed the heat flux predictions on a coarse grid ($P=3$) and a fine grid ($P=7$). Our results indicate that using the discrete derivative (DD) operator to construct the derivative-enhanced loss (DEL) term achieves the highest accuracy at the lowest computational cost. However, because the DD approach uses a discrete derivative operator, the neural operator does not exhibit super-resolution capabilities. Conversely, the least squares derivative (LSD) method enforces resolution-independent learning of temperature derivatives, demonstrating that employing a higher-resolution mesh significantly improves heat flux prediction accuracy.

The final column of Table~\ref{tab:result_error_hf} highlights the computational cost associated with training Fusion-DeepONet using different loss functions over 100,000 epochs. The DD operator presents the lowest computational expense. Based on the data presented in Table~\ref{tab:result_error_hf}, we conclude that the optimal choice of derivative loss term depends on the grid resolution used for data generation: DD is preferable for coarse-resolution meshes, while LSD provides superior accuracy for high-resolution meshes. Figure~\ref{fig:heat_pred} illustrates heat flux predictions for various test cases using different loss functions. Fusion-DeepONet was trained on the $P=3$ grid and evaluated on both $P=3$ and $P=7$ grids. According to Fig.~\ref{fig:heat_pred}, the discrete derivative loss consistently delivers the most accurate heat flux predictions across all tested cases.

\begin{figure}[!t]
\begin{center}
\includegraphics[width=0.95\textwidth]{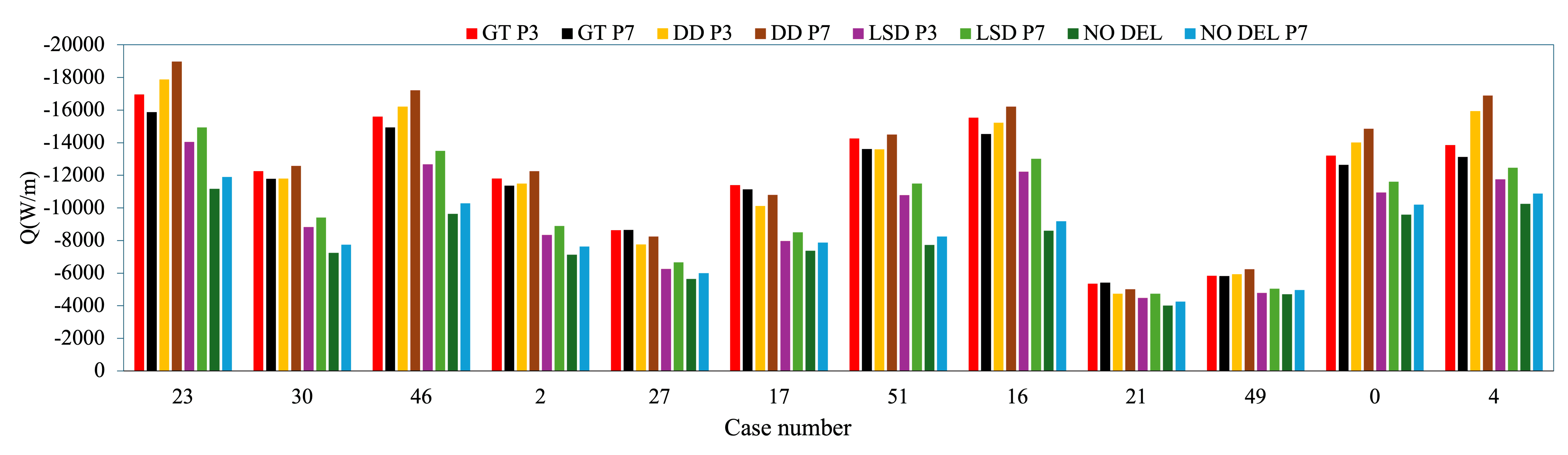}
\caption{Heat flux prediction using various loss functions for test samples. GT: Ground Truth, DD: Discrete Derivative operator, LSD: Least Square Differentiation, NO DEL: No derivative term in the loss function, P3: refers to coarse mesh with polynomial order 3, P7: refers to fine mesh with polynomial order 7.}
\label{fig:heat_pred}
\end{center}
\captionsetup{justification=centering}
\end{figure}

\subsection{Viscous flow inside convergent-divergent nozzle}
In this example, we learn a complex internal flow field inside a converging-diverging nozzle. Fusion-DeepONet learns a mapping from the shape of the nozzle to a steady state flow that includes density, x-velocity, y-velocity, pressure, and temperature. We split 60 cases into 42 training and 18 testing samples in this example. The Fusion-DeepONet architecture is designed to have 128 neurons in five hidden layers for the branch and trunk sub-networks. We use a single trunk network for all the trained variables (density, x-velocity, y-velocity, pressure, and temperature). To predict five variables, we employ $5\times 128$ neurons in the last linear layer of the branch network. We train the model for 100k epochs and select the best model among the saving checkpoints based on the best relative $L_2$ norm of the testing samples' error. We employ 16,751 unique coordinate points extracted from Trixi.jl results for training. For this problem, we did not employ the solution gradient loss terms to enhance the prediction of the solution derivative.


\begin{figure}[t!]
  \begin{center}
    \begin{tabular}{cccc}
\includegraphics[width=0.24\textwidth]{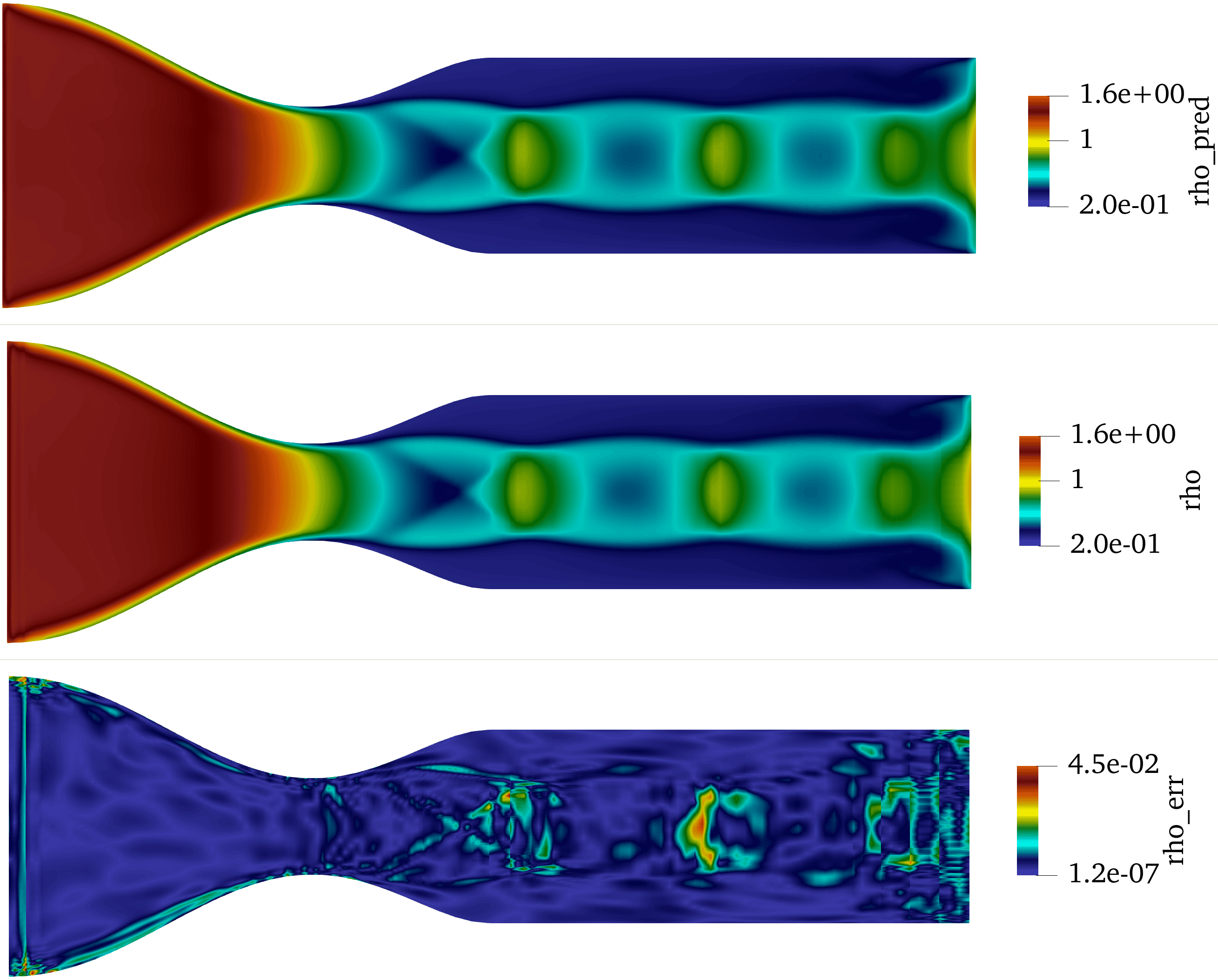}
 &
\includegraphics[width=0.22\textwidth]{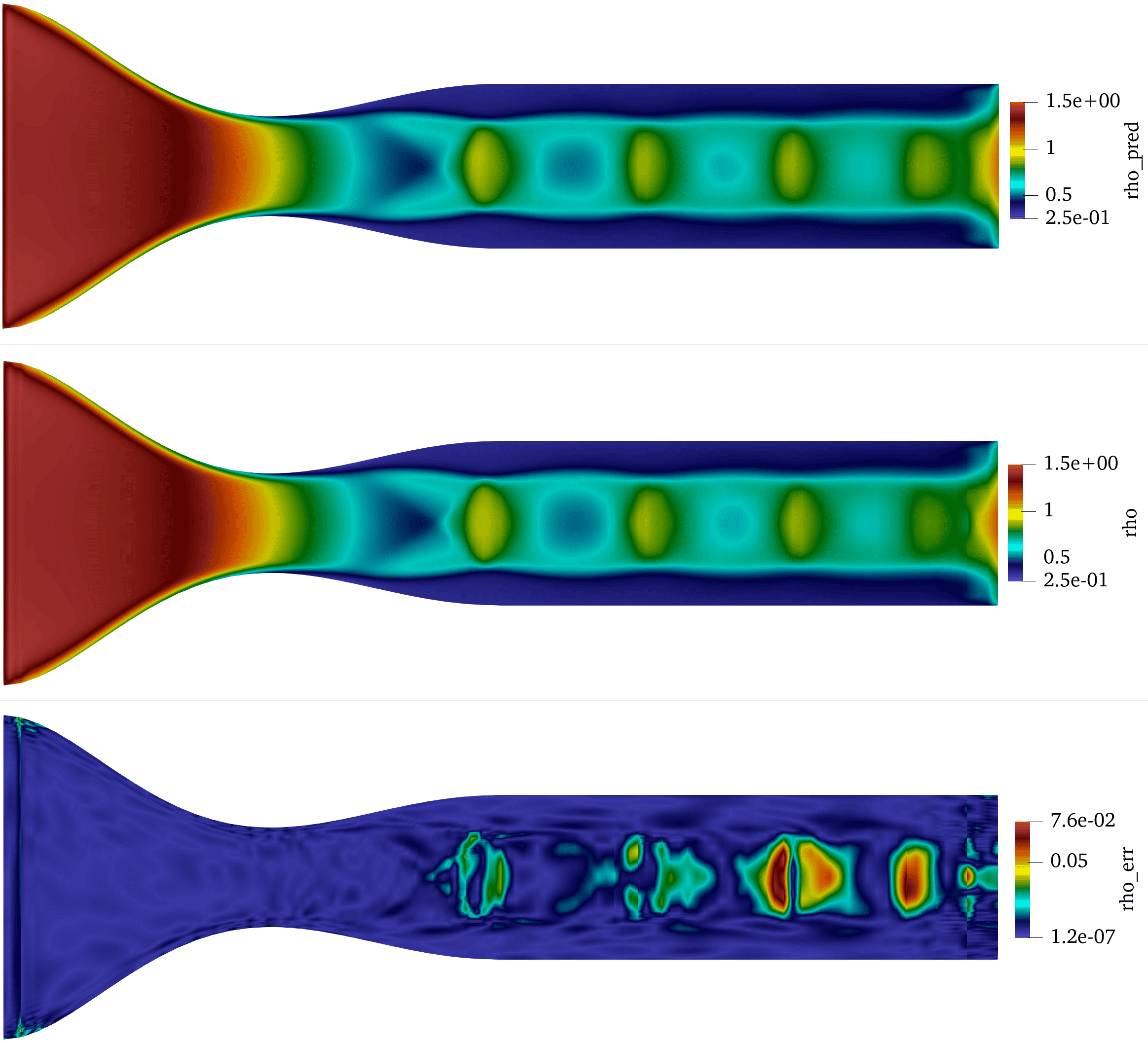}
&
\includegraphics[width=0.24\textwidth]{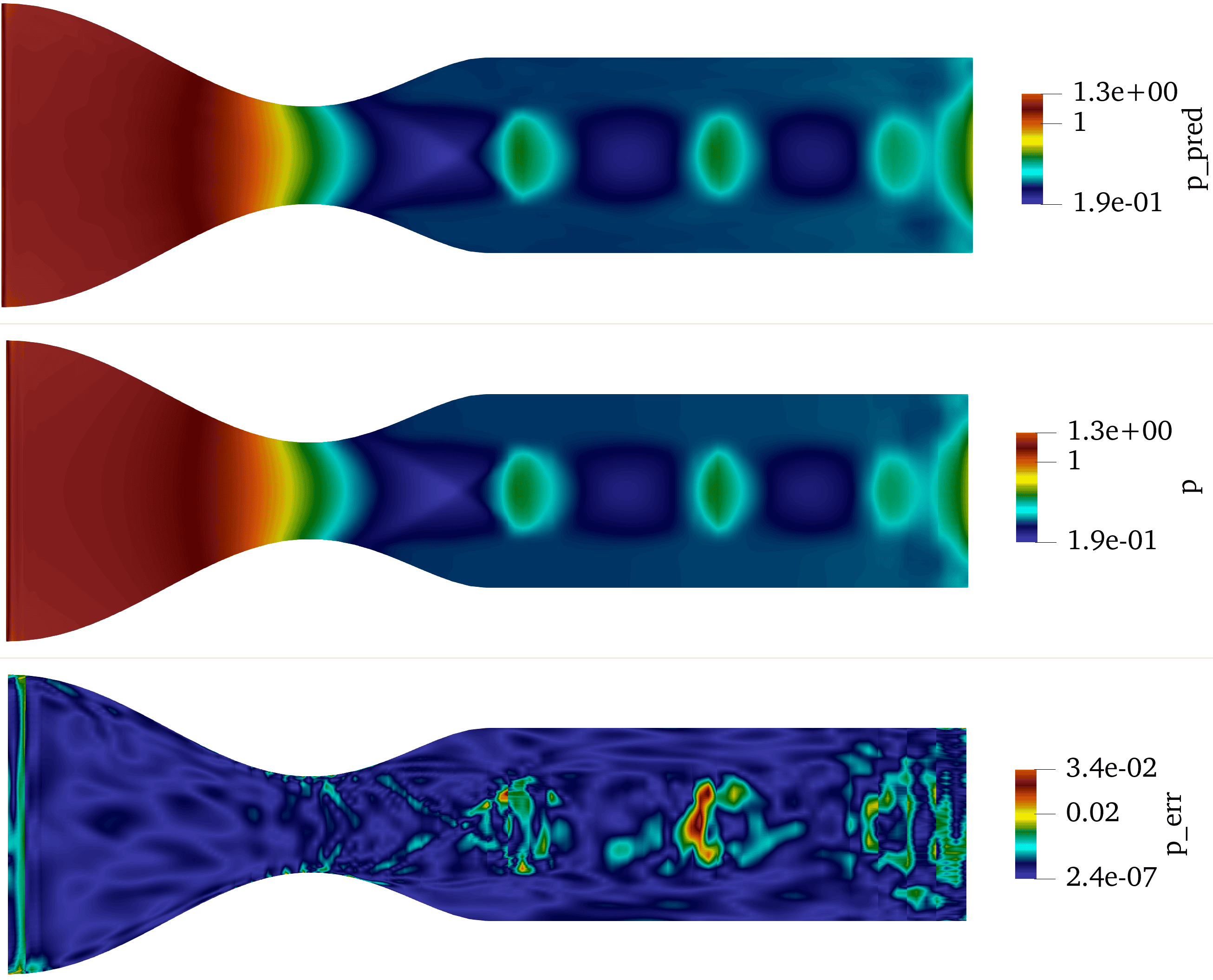}&
\includegraphics[width=0.22\textwidth]{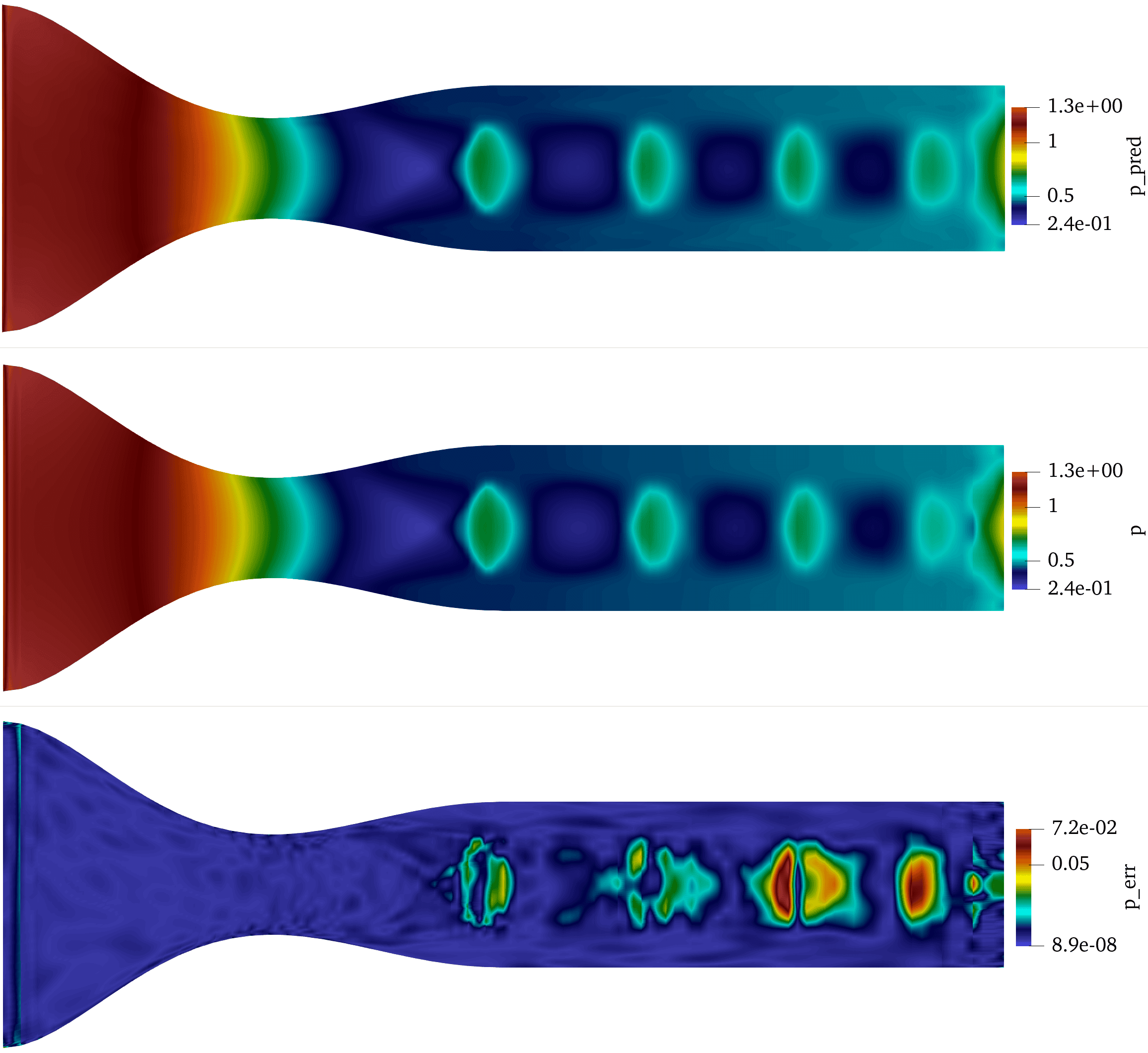}
\\
(a) Density case 1& (b) Density case 12 & (c) Pressure case 1 & (d) Pressure case 12
\\
\includegraphics[width=0.24\textwidth]{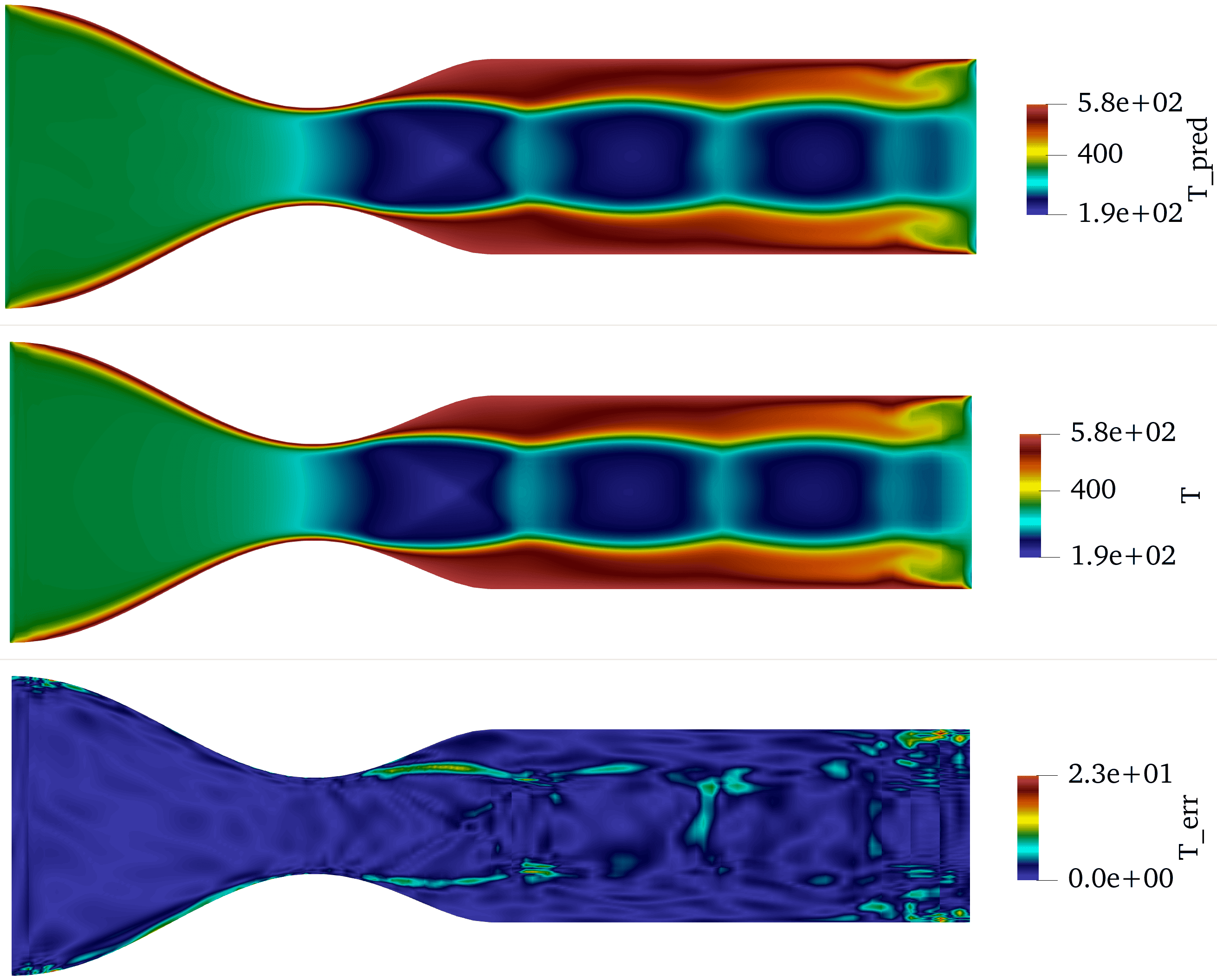}&\includegraphics[width=0.22\textwidth]{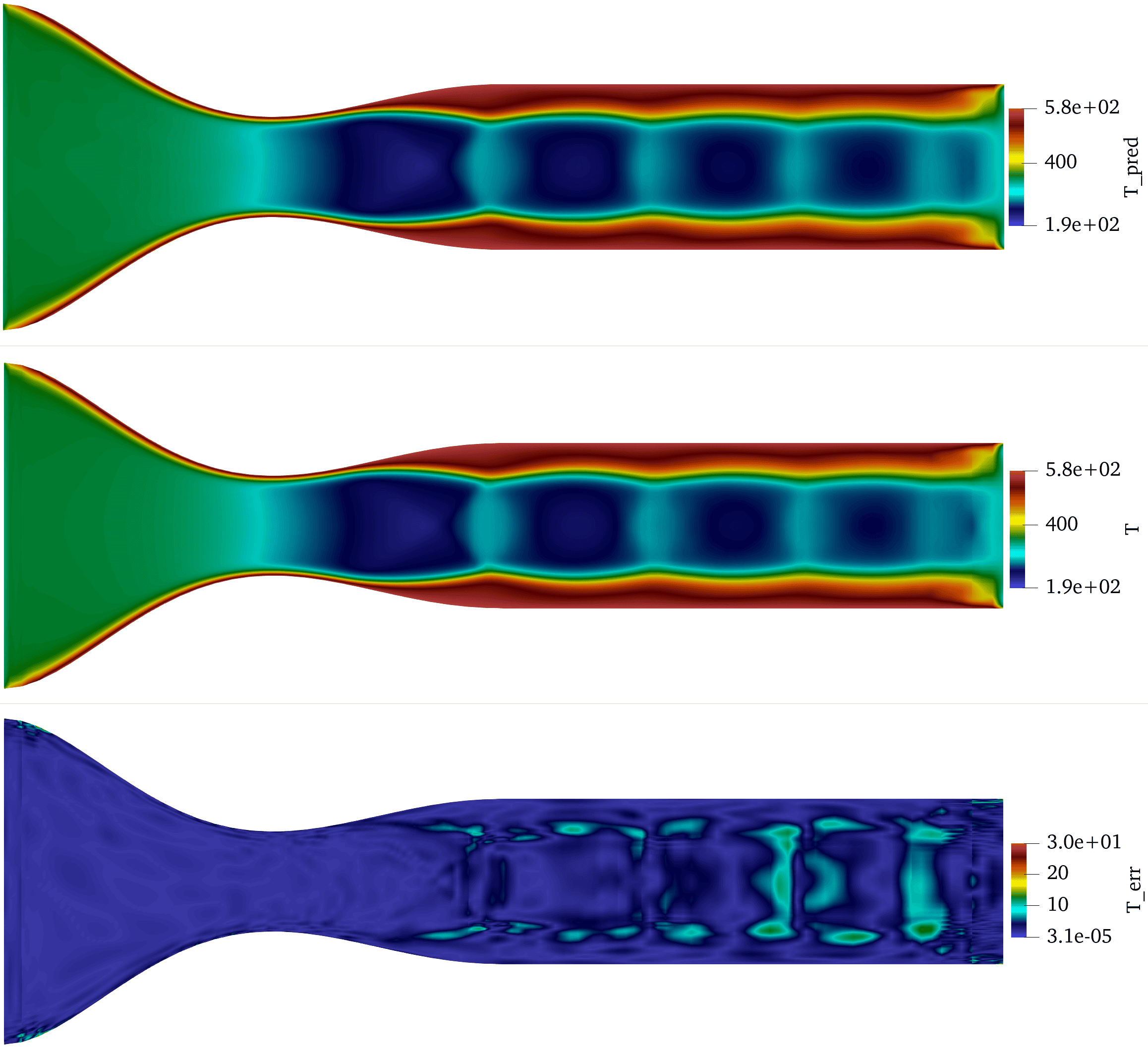}&\includegraphics[width=0.24\textwidth]{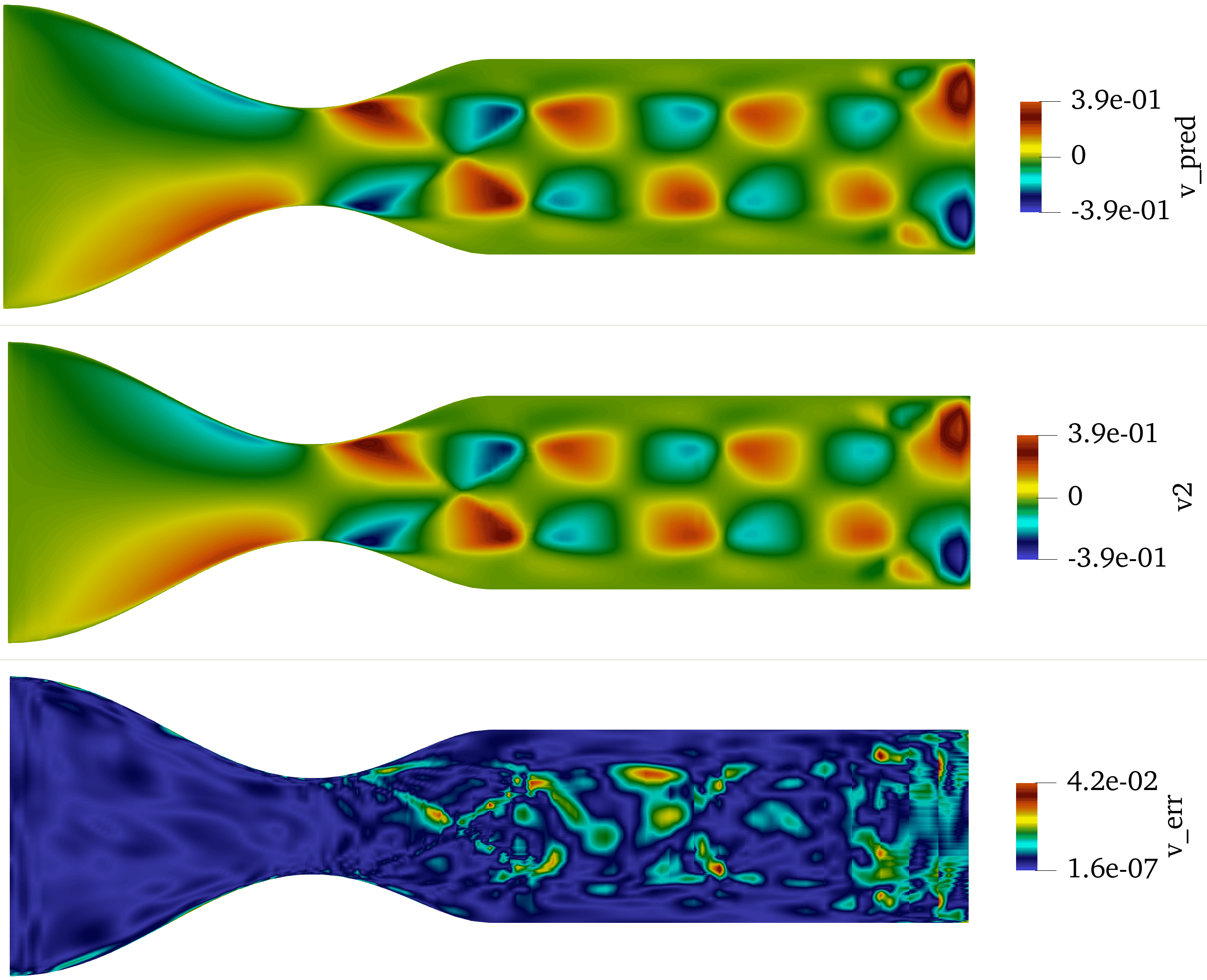}&\includegraphics[width=0.22\textwidth]{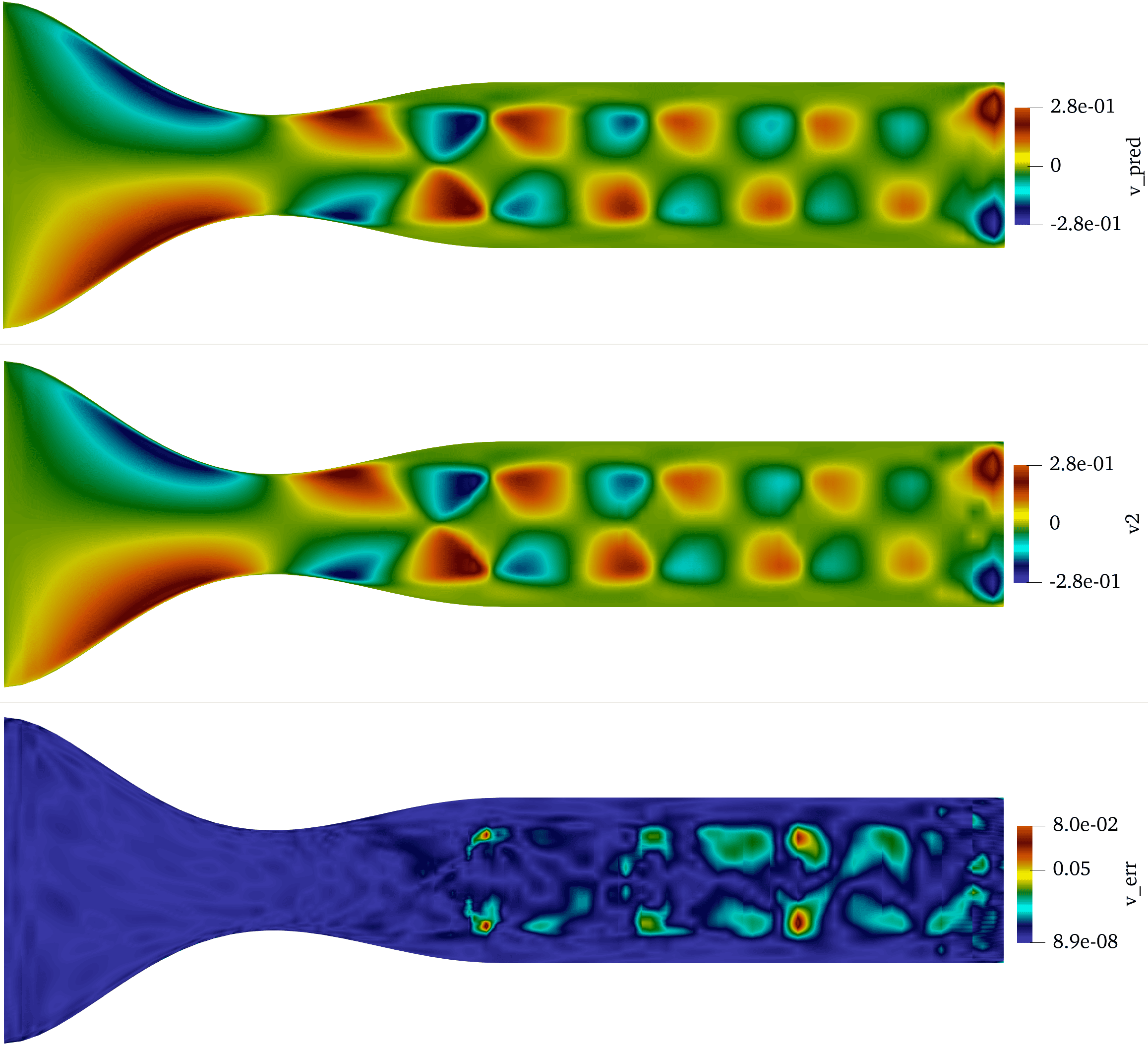}\\
(e) Temperature case 1& (f) Temperature case 12 & (g) y-Velocity case 1 & (h) y-Velocity case 12\\
\multicolumn{2}{r}{\includegraphics[width=0.24\textwidth]{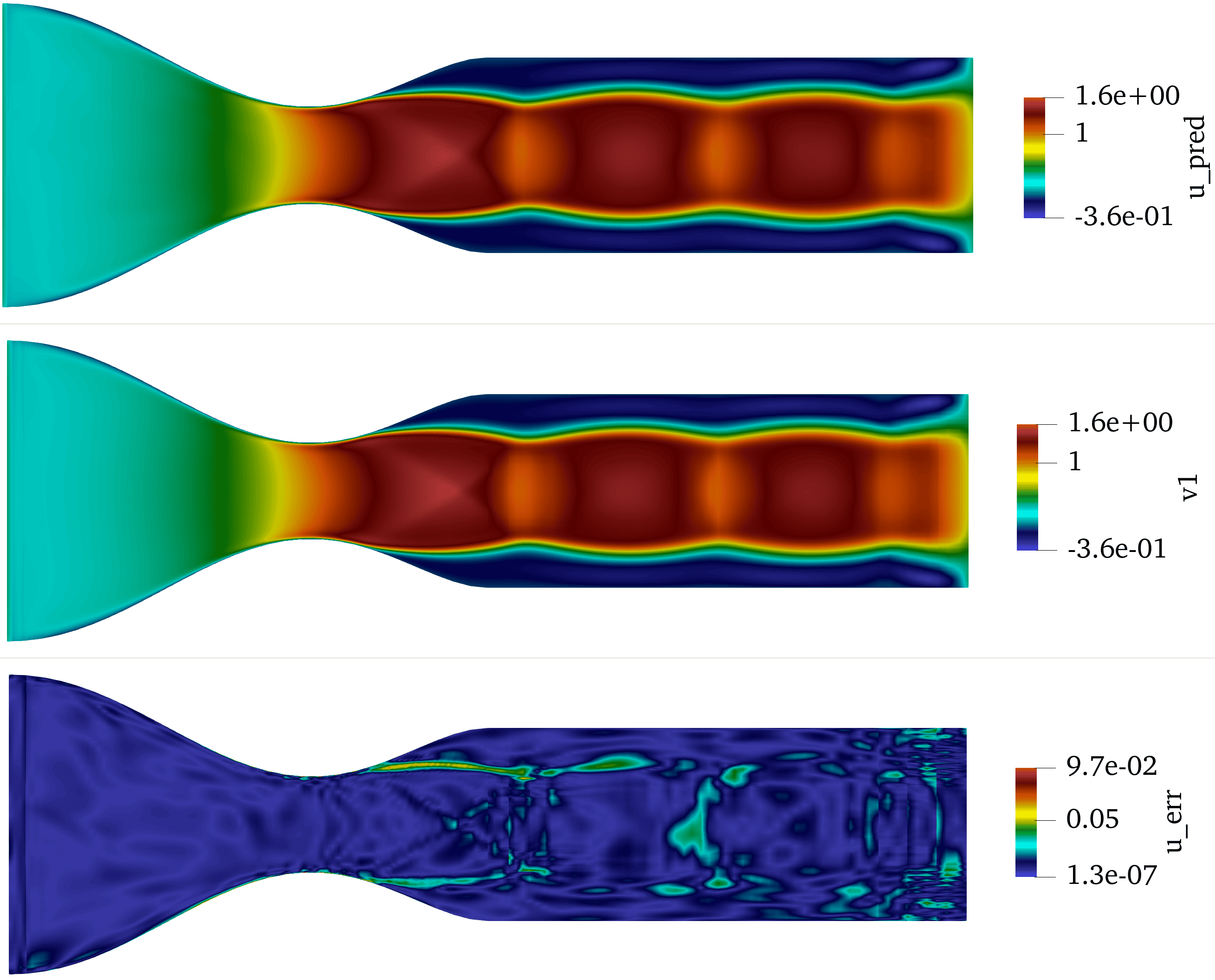}}&\multicolumn{2}{l}{\includegraphics[width=0.22\textwidth]{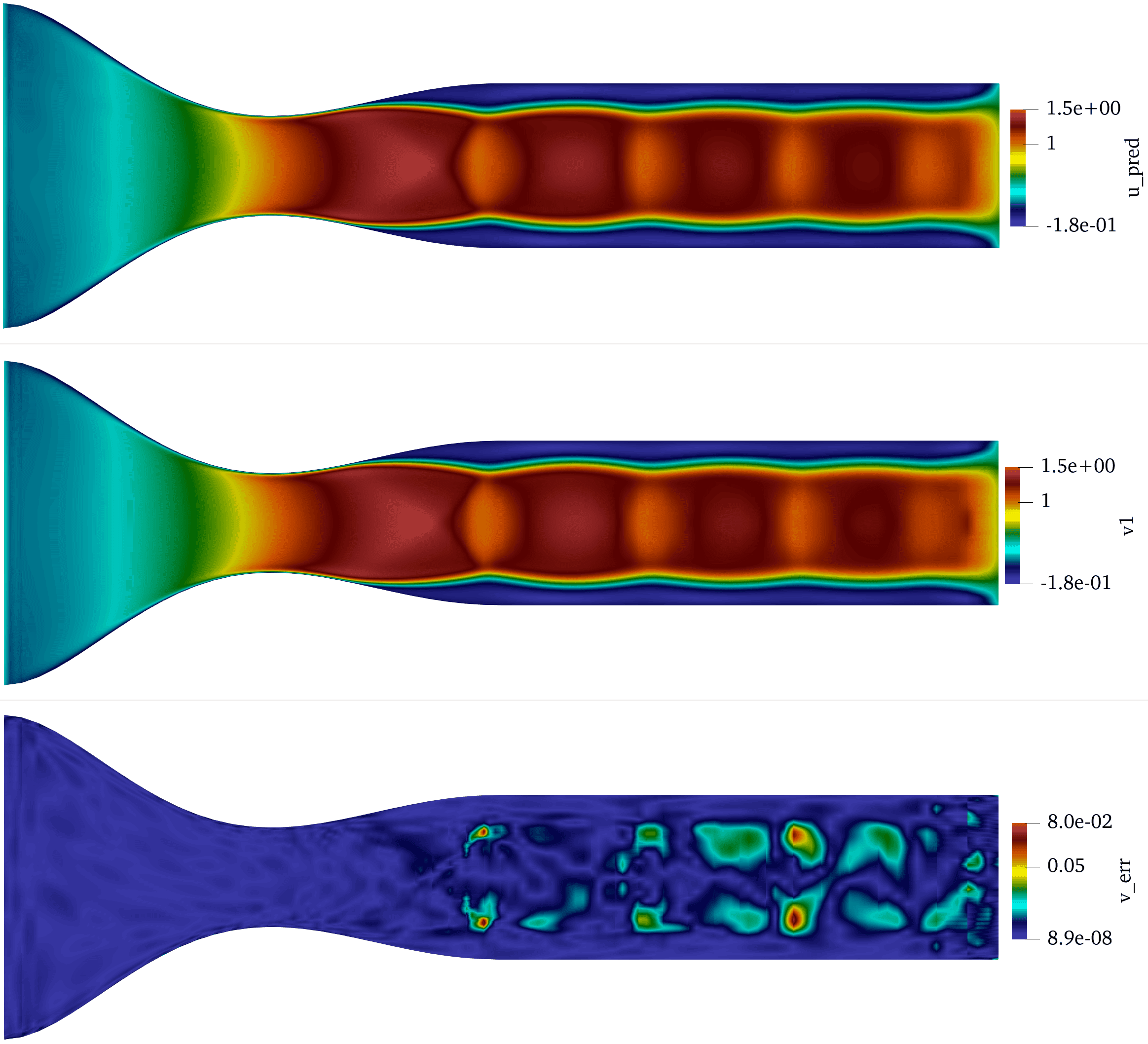}}\\
\multicolumn{2}{r}{(i) x-Velocity case 1}&\multicolumn{2}{l}{(j) x-Velocity case 12}
\end{tabular} 
\caption{Two unseen test cases of converging-diverging nozzle flow field predicted by Fusion-DeepONet. All the variables are non-dimensional except temperature which is reported in Kelvin unit ($K$).(a) Density case 1, (b) Density case 12, (c) Pressure case 1, (d) Pressure case 12, (e) Temperature case 1, (f) Temperature case 12, (g) y-Velocity case 1, (h) y-Velocity case 12, (i) x-Velocity case 1, (j) x-Velocity case 12. The Fusion-DeepONet was trained using 42 samples and tested for 18 unseen samples. In each sub-figure, from the top, the first plot shows prediction, second shows ground truth, and third shows point-wise error defined as $E=|\phi_{pred}-\phi|$ where $\phi$ denotes the primitive variables. See appendix.~\ref{append:datasets} for the values of input parameters for test cases 1 and 12.}
    \label{fig:CD_field}
  \end{center}
\end{figure}

Table~\ref{tab:result_error_cd} compares the accuracy for an unseen testing dataset for five primitive variables predicted by Fusion-DeepONet and MeshGraphNet. The training of Fusion-DeepOnet is performed using the JAX backend on the NVIDIA H100 GPU, while we used NVIDIA 3090 GeForce RTX using PyTorch for MeshGraphNet. Fusion-DeepONet can accurately learn the primitive variables for nozzle shape surrogate modeling better than MeshGraphNet for all the primitive variables. The precision of the Fusion-DeepONet deteriorates when inferring the y-velocity component ($v$). According to Figs.~\ref{fig:CD_field}(a)-(f) and (i) and (j), we can observe that the solution structure for density, pressure, x-velocity, and temperature shows some degree of similarity. However, the solution structure of y-velocity entirely differs from the rest. We mentioned previously that a single trunk sub-network generates the basis functions for predicting five variables, hence a higher error for $v$. Figure~\ref{fig:CD_field} shows that Fusion-DeepONet can predict the flow field near the boundaries, specifically near the wall boundary conditions. Moreover, comparing the flow field of Case 1 with Case 2, we observe that the flow field significantly varies since the geometry of the nozzle changes.

\begin{table}[!t]
\caption{Comparison of prediction error for the converging-diverging nozzle problem, \#P refers to number of trainable parameters. The $L^2$ norm reported is the average of relative $L^2$ norm across 18 unseen samples.}
\label{tab:result_error_cd}
\resizebox{\columnwidth}{!}{%
\begin{tabular}{@{}lccccccc@{}}
\toprule
\textbf{Framework} & \textbf{\#P} & $\%L^2 (\rho)$&$\%L^2 (u)$&$\%L^2 (v)$&$\%L^2 (p)$&$\%L^2(T)$ & time(sec)\\ \midrule
Fusion-DeepONet   & 298,182    & $\mathbf{1.29}$  & $\mathbf{2.41}$   & $\mathbf{11.26}$   & $\mathbf{1.39}$   & $\mathbf{0.97}$   & 5,569.5\\
MeshGraphNet   & 277,061   & 4.00  & 7.25  & 23.33   & 3.29   & 3.66   & 13,000\\
\bottomrule
\end{tabular}%
}
\end{table}

\section{Summary}

In this study, we propose three challenging problems for developing a neural operator-based framework for geometry-dependent hypersonic and supersonic flow problems. We use these three problems to evaluate the newly developed Fusion-DeepONet in learning high-speed flow fields for varying geometries. First, we compare several state-of-the-art neural operators, including parameter-conditioned U-Net, DeepONet, FNO, and MeshGraphNet, with our Fusion-DeepONet in learning a challenging geometry-dependent hypersonic inviscid flow around blunt bodies for uniform Cartesian and irregular grids. For the training dataset, we deliberately employed a limited number of samples. The scarce data set includes the density, x-direction velocity, y-direction velocity, and pressure of the hypersonic flow around a semi-ellipse that is parametrized using values of the minor and major axes. The solution involves a highly non-linear field with strong shocks and high-gradient regions.

Using 28 training samples, we first created a Cartesian uniform grid and interpolated all the samples into the Cartesian grid. This allows us to evaluate and compare different operator models that work on uniform gird. U-Net and Fusion-DeepONet outperformed other neural operators in generalization and prediction accuracy. However, Fusion-DeepONet achieved the same accuracy as U-Net using approximately 200 times fewer parameters than the U-Net framework. Moreover, the training time for Fusion-DeepONet is less than that of U-Net. The primary focus of this study was to develop a framework that can input arbitrary grids and generalize using a limited number of samples. Thereafter, we employ this Fusion-DeepONet to use unstructured and irregular grids for various training samples and predict the highly irregular hypersonic flow field. We employed Vanilla-DeepONet and MeshGraphNet to learn the solution field on an irregular, unstructured grid for comparison. They both performed far worse than Fusion-DeepONet and do not generalize to the irregular grid solution field. We then analyzed the reason for improved performance of Fusion-DeepONet using SVD decomposition of its trained trunk network. Using the decomposition of hidden-layer outputs, we observed that the Fusion-DeepONet operator can extract the maximum amount of information from the training dataset for the low and high modes. The decomposition of the trained trunk network using irregular grids shows that the Fusion-DeepONet can learn the grid-dependent information far better than the Vanilla and POD-DeepONet at their respective lowest generalization error.

Next, we learn viscous hypersonic flow around a reentry capsule, parameterized by two geometric variables and one flow condition (Mach number). The primary quantity of interest in this context is wall heat flux. To enable accurate heat flux prediction, we design a surrogate model that predicts the temperature field, from which surface heat flux can be computed. We augment the mean squared error loss with a derivative-informed term to enhance the model's ability to capture temperature gradients. This additional term improves Fusion-DeepONet's accuracy in predicting temperature derivatives at the boundaries, which is critical for accurate heat flux estimation. We demonstrate that even highly accurate temperature predictions do not necessarily yield accurate heat fluxes unless the model is enhanced to capture temperature gradients. Incorporating the derivative-based loss allows Fusion-DeepONet to serve as a robust surrogate model for hypersonic heat flux prediction, suitable for early-stage aerodynamic design. Moreover, we demonstrate that Fusion-DeepONet can construct a surrogate model dependent on both geometry and Mach number, enabling optimization of the reentry capsule geometry along the entire reentry trajectory across varying Mach numbers.

Finally, we apply Fusion-DeepONet to learn supersonic internal flow under varying geometries. A converging-diverging nozzle connected to an isolator duct is parameterized using three geometric variables. A single Fusion-DeepONet model is trained to simultaneously predict density, x-velocity, y-velocity, pressure, and temperature fields. For this complex scenario, Fusion-DeepONet significantly outperforms MeshGraphNet regarding prediction accuracy.

For the next step, we will apply the Fusion-DeepONet to learn hypersonic flow fields involving real chemistry. Moreover, the Fusion-DeepONet will be employed to learn a mapping from non-parametric geometries to time-dependent hypersonic flows in 2D and 3D spatial configurations. 
%

\section*{Acknowledgments}
This work was supported by the U.S. Army Research Laboratory
W911NF-22-2-0047 and by the MURI-AFOSR FA9550-20-1-0358. 
\appendix

\section*{Appendix}
\section{Appendix}

\subsection{Datasets}
\label{append:datasets}
In this section, we report on the training and testing samples for the three problems reported in this study. Table \ref{tab:semi_ellips} presents the values of the input parameters for the test and training data sets of the semi-ellipse problem. Table~\ref {tab:capsule_dataset} demonstrates the values of geometric and flow conditions input parameters for the reentry capsule problem used for training and testing. In Table~\ref {tab:CD}, we report the input geometric parameters of the converging-diverging nozzle selected for training and testing neural operators.

\begin{table}[!h]
\caption{Training and testing samples for input parameters of surrogate models of inviscid hypersonic flow over semi-elliptic blunt bodies.}
\label{tab:semi_ellips}
\begin{center}
\resizebox{\columnwidth}{!}{%
\begin{tabular}{@{}cccccccccc@{}}
\toprule
Case &  $(a,b)$ Train &Case &  $(a,b)$ Train& Case &  $(a,b)$ Train&Case &  $(a,b)$ Train& case &  $(a,b)$ Test\\ \midrule
1 & (2.78,0.70) & 9  & (2.89,1.26) & 17 & (1.27,1.30) & 25 & (2.36,1.09) & 1 & (0.90,0.80) \\
2 & (1.05,0.88) & 10 & (1.51,1.69) & 18 & (0.53,0.53) & 26 & (1.62,0.84) & 2 & (2.64,1.36) \\
3 & (1.57,1.03) & 11 & (2.08,0.74) & 19 & (2.72,1.53) & 27 & (1.21,1.44) & 3 & (1.44,1.16) \\
4 & (0.77,1.66) & 12 & (1.90,1.39) & 20 & (2.46,1.00) & 28 & (1.79,1.40) & 4 & (0.80,0.94) \\
5 & (2.27,0.64) & 13 & (2.18,1.75) & 21 & (1.69,1.78) &    &             & 5 & (2.41,1.25) \\
6 & (0.66,1.20) & 14 & (2.56,1.59) & 22 & (2.84,1.08) &    &             & 6 & (0.98,1.63) \\
7 & (2.99,0.90) & 15 & (1.10,1.50) & 23 & (1.35,0.59) &    &             & 7 & (2.10,0.76) \\
8 & (0.63,1.15) & 16 & (1.15,1.55) & 24 & (1.85,0.54) &    &             & 8 & (2.00,0.68)\\
\bottomrule
\end{tabular}
}
\end{center}
\end{table}

\begin{table}[!h]
\caption{Training and testing samples for input parameters of surrogate models for reentry capsule.}
\label{tab:capsule_dataset}
\resizebox{\columnwidth}{!}{%
\begin{tabular}{@{}ccccccccccccccc@{}}
\toprule
Case & $(R,\alpha,Ma)$  Train & Case & $(R,\alpha,Ma)$  Train & Case & $(R,\alpha,Ma)$  Train & Case & $(R,\alpha,Ma)$  Train & Case & $(R,\alpha,Ma)$  Train & Case & \multicolumn{2}{c}{$(R,\alpha,Ma)$  Test} & Case & $(R,\alpha,Ma)$  Test \\ \midrule
1    & (19.05,50.56,6.97)     & 11   & (16.59,57.58,5.00)     & 21   & (13.47,41.35,6.42)     & 31   & (18.80,45.39,6.36)     & 41   & (14.19,51.65,5.67)     & 1    & \multicolumn{2}{c}{(15.79,53.26,7.78)}    & 11   & (17.25,54.19,7.36)    \\
2    & (18.61,42.64,6.06)     & 12   & (15.61,56.70,5.27)     & 22   & (13.37,40.35,5.14)     & 32   & (19.24,48.12,8.00)     & 42   & (14.56,43.68,6.13)     & 2    & \multicolumn{2}{c}{(14.89,43.10,6.33)}    & 12   & (16.92,56.39,7.68)    \\
3    & (16.46,51.94,7.40)     & 13   & (13.08,52.46,5.31)     & 23   & (17.60,58.94,7.23)     & 33   & (19.76,53.95,7.52)     & 43   & (18.12,47.01,7.56)     & 3    & \multicolumn{2}{c}{(18.29,44.97,7.28)}    &      &                       \\
4    & (13.80,52.22,5.72)     & 14   & (19.50,41.78,5.56)     & 24   & (14.70,59.20,6.74)     & 34   & (14.94,59.68,7.13)     & 44   & (14.37,46.97,7.00)     & 4    & \multicolumn{2}{c}{(14.59,44.40,6.24)}    &      &                       \\
5    & (19.93,45.29,5.79)     & 15   & (19.81,47.85,5.21)     & 25   & (19.60,42.95,7.47)     & 35   & (16.84,52.97,7.34)     & 45   & (18.70,56.24,6.62)     & 5    & \multicolumn{2}{c}{(17.87,49.40,5.54)}    &      &                       \\
6    & (17.14,57.07,5.83)     & 16   & (17.37,46.63,5.45)     & 26   & (17.76,55.01,6.26)     & 36   & (16.66,50.92,7.85)     & 46   & (17.53,55.51,6.04)     & 6    & \multicolumn{2}{c}{(16.18,40.07,5.97)}    &      &                       \\
7    & (13.21,40.90,6.66)     & 17   & (13.59,49.99,5.88)     & 27   & (18.00,48.95,6.49)     & 37   & (19.18,58.51,5.64)     & 47   & (15.17,54.85,5.09)     & 7    & \multicolumn{2}{c}{(15.54,44.30,6.77)}    &      &                       \\
8    & (16.35,47.57,6.60)     & 18   & (15.08,49.07,7.87)     & 28   & (18.88,45.86,6.92)     & 38   & (16.06,46.03,7.63)     & 48   & (13.97,50.32,7.73)     & 8    & \multicolumn{2}{c}{(14.02,48.62,7.19)}    &      &                       \\
9    & (18.44,42.24,7.93)     & 19   & (14.29,57.90,5.42)     & 29   & (17.09,54.38,7.06)     & 39   & (15.84,55.70,6.16)     &      &                        & 9    & \multicolumn{2}{c}{(19.33,53.46,5.37)}    &      &                       \\
10   & (15.35,58.15,6.85)     & 20   & (13.70,51.21,6.50)     & 30   & (16.22,41.18,5.17)     & 40   & (15.39,43.37,6.89)     &      &                        & 10   & \multicolumn{2}{c}{(18.20,59.52,5.95)}    &      &    \\
\bottomrule
\end{tabular}%
}
\end{table}

\begin{table}[!h]
\caption{Training and testing samples for input parameters of surrogate models for supersonic flow in the converging diverging nozzle.}
\label{tab:CD}
\resizebox{\columnwidth}{!}{%
\begin{tabular}{@{}cccccccccccccc@{}}
\toprule
Case & $(h_i,h_o,x_t)$  Train & Case & $(h_i,h_o,x_t)$  Train & Case & $(h_i,h_o,x_t)$  Train & Case & $(h_i,h_o,x_t)$  Train & Case & $(h_i,h_o,x_t)$  Train & Case & $(h_i,h_o,x_t)$ Test & Case & $(h_i,h_o,x_t)$  Test \\ \midrule
1    & (3.86,2.03,0.58)       & 10   & (3.33,2.41,0.57)       & 19   & (3.18,2.39,0.48)       & 28   & (3.84,1.79,0.58)       & 37   & (3.88,2.43,0.49)       & 1    & (3.73,1.85,0.62)     & 10   & (3.22,1.72,0.53)      \\
2    & (3.80,1.63,0.52)       & 11   & (3.51,2.38,0.45)       & 20   & (3.09,2.06,0.55)       & 29   & (3.58,2.22,0.59)       & 38   & (3.43,1.80,0.63)       & 2    & (3.19,1.85,0.58)     & 11   & (3.69,1.97,0.49)      \\
3    & (3.49,2.10,0.61)       & 12   & (3.37,2.33,0.47)       & 21   & (3.06,1.57,0.54)       & 30   & (3.46,1.56,0.46)       & 39   & (3.40,2.29,0.53)       & 3    & (3.81,2.31,0.56)     & 12   & (3.45,1.50,0.51)      \\
4    & (3.11,2.11,0.50)       & 13   & (3.00,2.12,0.47)       & 22   & (3.04,1.52,0.46)       & 31   & (3.83,1.77,0.54)       & 40   & (3.34,1.67,0.58)       & 4    & (3.64,2.28,0.52)     & 13   & (3.36,1.71,0.57)      \\
5    & (3.99,1.76,0.50)       & 14   & (3.93,1.59,0.49)       & 23   & (3.65,2.45,0.60)       & 32   & (3.89,1.91,0.65)       & 41   & (3.16,2.08,0.49)       & 5    & (3.30,2.24,0.46)     & 14   & (3.14,1.93,0.60)      \\
6    & (3.59,2.35,0.51)       & 15   & (3.97,1.89,0.46)       & 24   & (3.24,2.46,0.57)       & 33   & (3.97,2.20,0.62)       & 42   & (3.22,1.68,0.53)       & 6    & (3.13,2.02,0.63)     & 15   & (3.90,2.17,0.47)      \\
7    & (3.02,1.54,0.56)       & 16   & (3.62,1.83,0.48)       & 25   & (3.94,1.65,0.61)       & 34   & (3.27,2.48,0.59)       &      &                        & 7    & (3.39,2.16,0.64)     & 16   & (3.74,2.48,0.51)      \\
8    & (3.47,1.88,0.56)       & 17   & (3.08,2.00,0.51)       & 26   & (3.68,2.25,0.53)       & 35   & (3.54,2.15,0.61)       &      &                        & 8    & (3.26,1.66,0.54)     & 17   & (3.60,2.21,0.61)      \\
9    & (3.78,1.61,0.65)       & 18   & (3.29,1.95,0.64)       & 27   & (3.71,1.95,0.55)       & 36   & (3.52,2.05,0.64)       &      &                        & 9    & (3.75,1.75,0.60)     & 18   & (3.56,2.32,0.63) \\
\bottomrule
\end{tabular}%
}
\end{table}


\subsection{High pressure ratio Sod problem (LeBlanc problem)}
\label{sod}

For completeness, we evaluated the Fusion-DeepONet on the challenging high-pressure-ratio Sod problem, also known as the LeBlanc problem. This problem is complicated due to the steep pressure discontinuity it involves. Recent research has explored various neural operators designed to handle sharp discontinuities in this context. For example, Peyvan \emph{et al.} \cite{riemannonet} and Shih \emph{et al.} \cite{SHIH2025117560} utilized approaches such as parameter-conditioned UNet, two-step Rowdy DeepONet, and Transformers to tackle the high-pressure-ratio Sod problem effectively.

We solve the 1D compressible Euler equations over the spatial domain $x \in [-20,20]$. The initial conditions for this problem are defined as: 

\begin{equation}
\left(\rho, u , p\right)=\begin{cases}
\left(2.0,0.0,p_l\right) & x \le -10 \\ 
\left(0.001,0.0,1.0\right) & x > -10, 
\end{cases}
\label{eq:low_p}
\end{equation}
where $p_l \in [10^9,10^{10}]$. 

We use the exact solution method to compute results at $t_f=0.0001$. 500 equispaced $p_l$ values are sampled, with 400 cases randomly assigned for training and 100 for testing. We construct the loss function to train the neural networks using the logarithm of density and pressure values. We want to create a mapping from $p_l$ to the final solution for $\rho$, $u$ and $p$. During inference, the exponential function is applied to the predicted logarithmic values of density and pressure, converting them back to their physical values. For Fusion-DeepONet, we chose three hidden layers, each with 100 neurons in the branch and trunk networks. We employed a single trunk network to predict all three variables, including $\rho$, $u$, and $p$. We then trained the network for 100,000 epochs with an exponential decay profile for learning rate. The exponential decay setup is identical to the one we used for the semi-ellipse geometry-dependent problem. Table.~\ref{tab:lebalnc} demonstrates a comparison of Fusion-DeepONet accuracy versus 2-step DeepONet \cite{riemannonet}, UNet \cite{riemannonet}, and Transformer \cite{SHIH2025117560} in predicting the discontinuous solution of the LeBlanc problem. According to Table.~\ref{tab:lebalnc}, Fusion-DeepONet performs better than all its counterparts in predicting all the flow variables. This test showcases the generalization of the Fusion-DeepONet prediction in learning discontinuous solutions.

\begin{table}[!t]
\caption{Comparison of L$_2$ norms of the error for all the unseen test cases of the LeBlanc problem predicted by the trained 2-step DeepONet \cite{riemannonet}, UNet \cite{riemannonet}, Transformer \cite{SHIH2025117560}, and Fusion-DeepONet.}
\label{tab:lebalnc}
\begin{center}
\begin{tabular}{@{}lcccc@{}}
\toprule
\textbf{Framework}& $\%\textrm{L}_2 (\rho)$&$\%\textrm{L}_2 (u)$&$\%\textrm{L}_2 (p)$&$\%\textrm{L}_2$\\ \midrule
2 step DeepONet \cite{riemannonet} & 0.66  & 3.39   & 2.86   & 2.31 \\
UNet\cite{riemannonet} & 1.00  & 2.65   & 2.27   & 1.97  \\
Transformer \cite{SHIH2025117560} & 1.42  & 2.27   & 2.23   & 1.97\\
Fusion-DeepONet  & \textbf{0.61}    & \textbf{0.59}      & \textbf{1.23}   &\textbf{0.81} \\
\bottomrule
\end{tabular}%
\end{center}
\end{table}

\section{Neural Operators}
\label{append:neural_ops}
This section describes the architectures of various neural operators used to learn the hypersonic flow field around blunt bodies and supersonic flow in converging-diverging nozzle. Our goal is to learn the flow by using the object's geometry as input.
\subsection{Vanilla-DeepONet}
Lu \emph{et al.} \cite{lu2019deeponet} developed deep operator networks (DeepONet) inspired by the universal approximation theorem of operators to map from an infinite dimension functional input to another infinite dimension functional output, both defined in the Banach space. The DeepONet framework consists of two distinct networks, namely branch and trunk nets. The branch net encodes the input parameters, and the trunk net encodes the independent variables, namely spatial coordinates and time. For this study, we focus on learning the solution at a specific time and avoiding learning the time-dependent dynamic. Therefore, we only input spatial coordinates into the trunk net and geometric parameters into the branch network. We multiply the trunk net output by the branch net output to construct the final prediction of DeepONet. The output of the DeepONet consists of four flow variables: density, x-velocity, y-velocity, and pressure. With DeepONet, we try to learn an operator $\mathcal{G}$ that 
\begin{equation}
\mathcal{G}: h_g \mapsto \mathbf{U}(x,y)
    \label{eq:mapping}
\end{equation}
The operator $\mathcal{G}$ maps from geometric parameters denoted as $h_g$ to the solution field $\mathbf{U}=\left(\rho(x,y),u(x,y),v(x,y),p(x,y)\right)^T$ within a physical domain described in Eq.~\eqref{Uconserv}. The DeepONet framework can be described using a compact form as

\begin{equation}
\mathbf{U}_s(x,y)=\sum_{k=0}^{l_d}B_{k,s}(h_g;\theta)T_k(x,y;\Gamma)
    \label{eq:deeponet}
\end{equation}
where $B_{k,s}$ denotes the branch network for the $s^\textrm{th}$ output variable and at $k^\textrm{th}$ latent dimension. Let $HB$ be the total number of hidden layers in the branch network and $HT$ be the total number of hidden layers in the trunk network. We can describe the branch and trunk outputs as 
\begin{equation}
B_{k,s}=W_{ij}^{b,o}\mathbf{z}_{j}^{HB}+b_i^{b,o},
    \label{eq:branch},
\end{equation}
and 
\begin{equation}
T_{k}=W_{ij}^{t,o}\mathbf{a}_{j}^{HT}+b_i^{t,o},
    \label{eq:trunk}
\end{equation}
where $\mathbf{z}^{HB}_j$ and $\mathbf{a}^{TB}_j$ denote the outputs of the last hidden layer of branch and trunk networks, respectively. The outputs of the $l^\textrm{th}$ hidden layer of the branch and trunk networks can be formulated as 

\begin{align}
\mathbf{z}^{l}=W_{ij}^{b,l}\sigma\left(\mathbf{z}^{l-1}_j\right)+b_{i}^{b,l}
\\
\mathbf{a}^{l}=W_{ij}^{t,l}\sigma\left(\mathbf{a}^{l-1}_j\right)+b_{i}^{t,l}
    \label{networks}
\end{align}
and then for the input layer ($l=1$) of the branch and trunk network, we have 
\begin{align*}
    \mathbf{z}^{0}_j=\mathbf{H}_j,&\quad j=1,\cdots,p\\
    \mathbf{a}^{0}_j=\mathbf{X}_j,&\quad j=1,2
\end{align*}
where $\mathbf{H}_j\in \mathbb{R}^{N_s}$ and $\mathbf{X}_j \in \mathbb{R}^{N_p}$ indicate the input vectors of branch and trunk networks for the $j^\textrm{th}$ parameter and $j^\textrm{th}$ coordinate, respectively. In Eq.~\eqref{networks}, $\sigma$ denotes the activation function and superscript $b$, and $t$ refers to the quantity assigned to the branch and trunk networks.

\subsection{POD-DeepONet}
Lu \emph{et al.} \cite{lu2022comprehensive} introduced POD-DeepONet that replaces trunk
net with the training data's proper orthogonal decomposition (POD) modes. POD-DeepONet employs only a fully connected network for the branch net that learns the coefficients of the POD basis to construct the predictions. For POD-DeepONet, the predictions are approximated using the following formula:
\begin{equation}
\mathbf{U}_s(x,y)=\sum_{k=0}^{l_d}B_{k,s}(h_g;\theta)T_k(\xi)+T_0(\xi)
    \label{POD-deeponet}
\end{equation}
where $T_0(\xi)$ is the mean function of training target functions such as $\rho$, $u$, $v$, and $p$ variables. $T_k(\xi)$ are the POD basis functions computed using the target functions after removing the mean values.

\subsection{U-Net with parameter conditioning for uniform grid} \label{sec: UNet_desc}
U-Net, introduced by Ronneberger \cite{UNet2015}, was developed as an image segmentation convolution network. Later, \cite{gupta_unet} introduced a modified version of U-Net with parameter conditioning for generalizing the solution to PDEs across multiple conditions using the same network. For learning hypersonic flow field across multiple elliptical shapes, we utilize the parameter conditioning of U-Net and use a framework similar to \cite{gupta_unet} consisting of two network types: a multi-layer perceptron network (MLP) and a convolution encoder-decoder network. See Figure \ref{fig:UNet_arch} showing network details. Each MLP block consists of a linear layer followed by \textit{Batchnorm} and \textit{Activation} layers. We use `swish' activation function defined as $\sigma = \mathbf{x} \cdot sigmoid(\mathbf{x})$. The U-Net encoder and decoder modules contain a series of Encoder and Decoder blocks. Each \textit{Encoder} block consists of two sequential convolution layers, each followed by a \textit{Groupnorm} and \textit{Activation} layers. The \textit{Decoder} modules consist of two types of blocks: a \textit{Decoder} block and a \textit{Decoder Convolution} block. The \textit{Decoder} block consists of transposed convolution layers that upsample the image embeddings generated by the prior encoder layers. The \textit{Decoder Convolution} block consists of convolution layers.  

The inputs to the Convolution blocks are provided in the form of a binary image consisting of a grid-based sequence of zeros and ones. Domain points contained within the geometry are labeled with 0's while the remaining domain points are labeled with 1's. The U-Net convolution modules encode the multi-spatial embedding of the object's shape and map the geometric shape represented as a binary image and the output hypersonic field around the object. We condition the model further by using the two parameters defining the geometry: the lengths of major ($a$) and minor axes ($b$) of the semi-ellipse. These parameters are used as inputs to the MLP block. The output embeddings generated from the MLP block are combined with the outputs from each \textit{Encoder} block through multiple projection layers. The combination process provides the conditioning for creating a bijective mapping between input and output functions. The final layer converts the Decoder embeddings to the appropriate size for reconstructing the output fields over the domain.

\begin{figure}[h]
\begin{center}
\includegraphics[width=\textwidth]{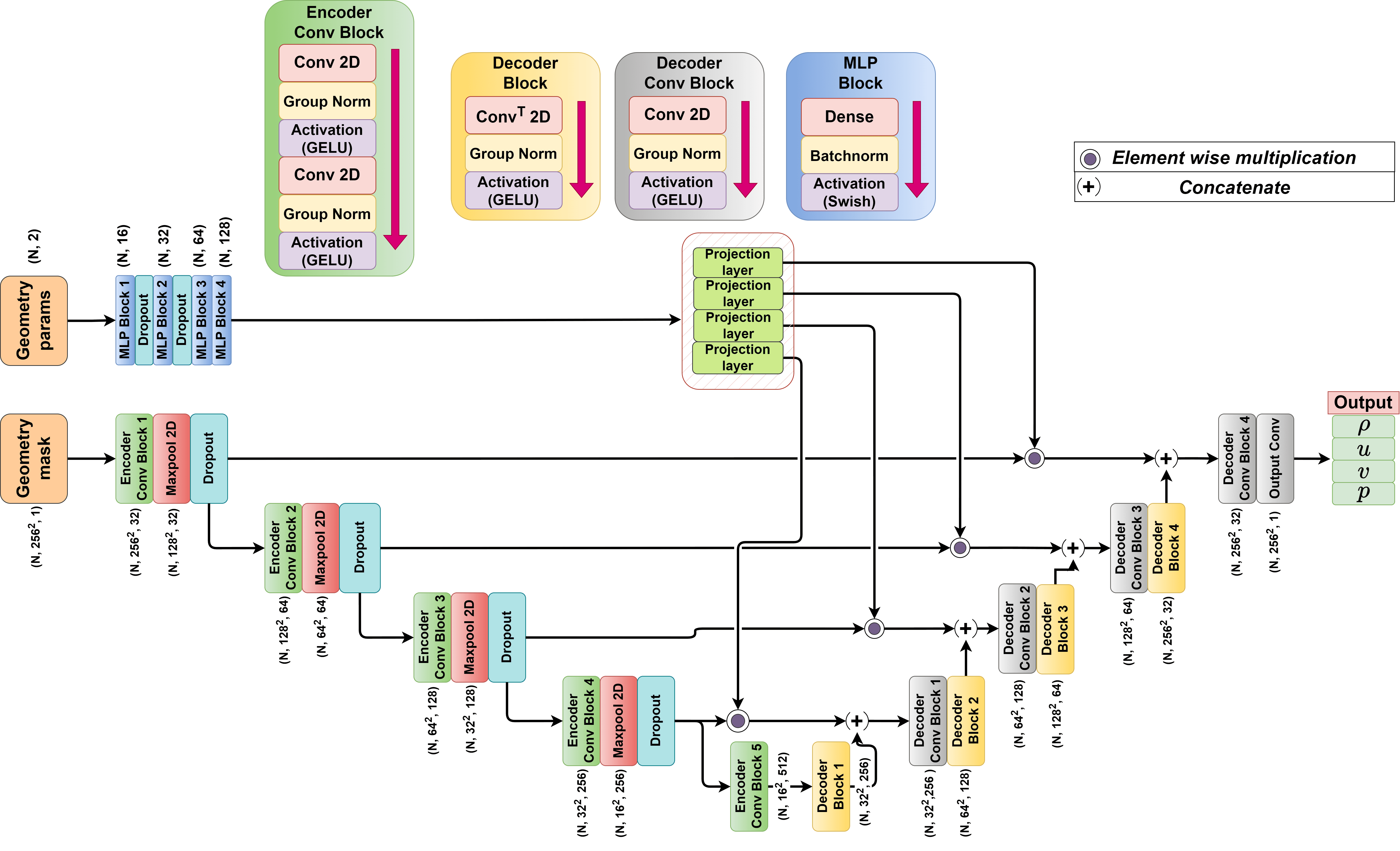}
\caption{Parameter conditioned U-Net architecture used for predicting density field across varying geometric shapes. The inputs consist of the geometry parameters (major and minor axes of ellipse) and the binary representation of the geometry. }
\label{fig:UNet_arch}
\end{center}
\captionsetup{justification=centering}
\end{figure}

\subsection{Fourier Neural Operator for uniform grid}
Fourier Neural Operator (FNO) introduced by Li et al. \cite{FNO} utilizes Fourier transformations of the input functions to learn the operator mapping between input and output for solving PDEs in a given domain  $D \subset \mathbb{R}^d$. For two infinite-dimensional bounded spaces $\mathcal{A} = \mathcal{A}(D; \mathbb{R}^{d_{in}})$ and $\mathcal{U} = (\mathcal{U};\mathbb{R}^{d_{out}})$, the operator mapping $\mathcal{G}_\theta: \mathcal{A} \rightarrow \mathcal{U}$ represents the approximation of the operator $u_i = \mathcal{G}_*(a_i)$, where $\{a_i, u_i\}, i = 1, \cdots, N$ are the pairs of data observations. Here, the input function $v(x) \in \mathcal{A}$ is first lifted to a higher dimension through the transformation $\mathcal{P}(v(x)): \mathbb{R}^{d_{in}} \rightarrow \mathbb{R}^{d_z}$ using a shallow fully-connected linear layer. A sequence of Fourier integral operations follows this applied iteratively on $v_l(x) \in \mathbb{R}^{d_z}$, defined as
\begin{equation}
    \begin{aligned}
        \mathcal{K}(\phi)v_l(x):= \mathcal{F}^{-1}  \left[\mathcal{R}_{\phi} \cdot (\mathcal{F}v_{l}) \right](x), \quad \forall x \in D
    \end{aligned}
\end{equation}
where $\mathcal{R}_{\phi}$ represents the Fourier transformation of a periodic function $\kappa: \Bar{D} \rightarrow \mathbb{R}^{d_v \times d_v}$ with parameters $\phi$, $\mathcal{F}$ and $\mathcal{F}^{-1}$ represent the Fourier and inverse Fourier transformations respectively. For frequency $k, (\mathcal{F}v_{l})(k) \in \mathbb{C}^{d_v}$ and $\mathcal{R}_\phi(k) \in \mathbb{C}^{d_v \times d_v}$ and since $\kappa$ is assumed to be periodic, it admits Fourier series expansion which allows working with discrete modes $k \in \mathbb{Z}^d$. The Fourier transform $\mathcal{F}$ is applied to each channel of $v_l(x)$, and the higher modes are truncated to filter the high-frequency elements and include only modes $k$ within:
\begin{equation}
    \begin{aligned}
        k_{max} = Z_{k_{max}} := \{k \in \mathbb{Z}^d: |k_j|\leq k_{max, i}, i = 1, \cdots d\}
    \end{aligned}
\end{equation}
For each mode of $\mathcal{F}(v_l)$, a different weight matrix $\mathcal{R_\phi} \in \mathbb{C}^{d_v \times d_v}$ is applied to form a complex-valued weight tensor $\mathcal{R}_l \in \mathbb{C}^{k \times d_v \times d_v}$. This output of each Fourier layer is combined linearly with the residual projection of $v_l(x) \in \mathbb{R}^{d_z \times d_z}$ through weight matrices $W_l \in \mathbb{R}^{d_v \times d_v}$. Thus, the output of the $l^{th}$ Fourier layer is given by
\begin{equation}
    \begin{aligned}
        v_{l+1} = \sigma \left( \mathcal{F}^{-1} (\mathcal{R}_l \cdot \mathcal{F}(v_l)) + W_l \cdot v_l + \mathbf{b}_l  \right), \label{eqn:FNO_comb_out}
    \end{aligned}
\end{equation}
where $\sigma$ is a non-linear activation and $\mathbf{b}_l \in \mathbb{R}^{d_z}$ represents the layer bias. 

In this work, the input to FNO is generated by concatenating the grayscale binary image $(256 \times 256 \times 1)$ and the domain coordinates $\{x, y\} \in \mathbb{R}^2$ to create a three-channeled input. For FNO, we utilize the vanilla version of FNO implementation \footnote{\url{https://github.com/neuraloperator/Geo-FNO/blob/main/airfoils/naca_interp_fno.py}} and modify the network architecture and hyper-parameters to achieve lowest errors during validation. The discretized input ($a_i$) and output function ($u_i$) are normalized to a scale of $0 - 1$ using the transformation $x_{norm} = \left[(x - x_{min}) / (x_{max} - x_{min})\right]$. The framework consists of five spectral convolution layers that are initialized as complex weights $W_{conv} \in \mathbb{C}^{d_{in} \times d_{out} \times k_1 \times k_2}$ multiplied with a scaling factor $\dfrac{1}{d_{in} + d_{out}}$. The input $v(x)$ is projected into a higher dimension through initial transformation $\mathcal{P}(v(x): \mathbb{R}^{3} \rightarrow \mathbb{R}^{32}$. In the Fourier layer, this high dimension embedding $v_l(x)$ is transformed using a Fourier transformation $\mathcal{F}v_l(x)$ where the higher spectrum modes $k1$ and $k2$ are filtered before performing a dot product with $W_{conv}$. The Fourier space embedding is converted back into a real domain using an inverse Fourier transformation $\mathcal{F}^{-1}$. The output from the Fourier layers is combined with the residual projection of $v_l(x)$ as per Equation \ref{eqn:FNO_comb_out}. The output is obtained by passing the embeddings obtained from the Fourier layers through two fully connected layers. Notably, the activation function is required only outside the Fourier layers, and here, we use ReLU as our choice of activation. We use Adam as the choice of optimizer with an initial learning rate of $1e^{-3}$ with a weight decay rate of $1e^{-4}$. The network is trained for 2500 epochs using relative $\mathcal{L}_2$ norm as the loss function. Architecture details for the FNO framework are shown in Table \ref{tab:FNO_details}. 

\begin{table}[h]
\caption{Architecture details for FNO}
\label{tab:FNO_details}
\resizebox{\textwidth}{!}{%
\begin{tabular}{@{}lcccc@{}}
\toprule
\multicolumn{1}{c}{\textbf{Layer   type}} & \textbf{Number of layers} & \textbf{Filter size/Number of nodes} & \textbf{Filtering modes} & \textbf{Activation} \\ \midrule
Spectral convolution (Wconv) & 5 & {[}64, 64, 128, 128, 256{]} & {[}64, 64, 128, 128, 256{]} & None \\
2D convolution               & 5 & {[}64, 64, 128, 128, 256{]} & None                        & None \\
Fully connected              & 2 & {[}128, 4{]}                & None                        & ReLU \\ \bottomrule
\end{tabular}%
}
\end{table}

\subsection{MeshGraphNets for unstructured irregular grid}
MeshGraphNets \cite{MeshGraphNet} combines graph neural networks (GNNs) with mesh-based geometric representations of simulation domains for capturing spatial relationships. Unlike conventional convolution-based frameworks, graph-based frameworks can operate directly on the mesh data that is commonly used for scientific tasks and, hence, do not necessarily require uniform grids. MeshGraphNets utilize the conventional messaging passing scheme in an Encoder-Process-Decoder framework \cite{graph_sanchez} for mapping the input graph to an output graph with the same structure but different node features. Here, the graph nodes correspond to the mesh nodes, while the graph edges correspond to the edge connections between the mesh nodes. The encoder $\mathcal{I}: \mathcal{X} \rightarrow \mathcal{G}$ embeds the node features $X$ as latent graph, $G^l = \mathcal{I}(X)$, where $G:=\{V, E, X\}$ with $V := {v_1, v_2, \cdots, v_N}$ representing the spatial location of mesh points and $E := \{e_{ij}:(i,j) \in V \times V\}$ representing the edges connecting the mesh points, and $X \in \mathbb{R}^{N \times d}$ denoting the features of each node (if available). Separate encoders are used to encode the node and edge features from the graph. The edge attributes for each edge are determined by calculating the spatial separation between the node pair $\{v_i, v_j\}$ connected by the edge and appending it with the norm $||v_i - v_j||_{\mathcal{L}_2}$. The encoder $I$ embeds the node and edge feature into a latent vector at each node and edge using a sequence of MLPs. The processor consists of sequential, fully connected layers for message passing and node and edge features aggregation. The mesh edge $e_{ij}$ and node embeddings $v_i$ are updated by
\begin{equation}
    \begin{aligned}
    e_{ij}^\prime \leftarrow \mathcal{N}_{\theta_E}\left(e_{ij}, v_i, v_j \right) \\
    v_i^\prime \leftarrow \mathcal{N}_{\theta_V} ( v_i, \sum_{j} e_{ij}^\prime ) + v_i
    \end{aligned}
\end{equation}
where $\mathcal{N}_{\theta_E}$ and $\mathcal{N}_{\theta_V}$ represent the processor MLP layers for edge and node, respectively. The decoder module uses a series of fully connected networks and decodes the latent node embeddings generated by the processor into one or more output node features $p_i$. The network is trained using a mean-squared error loss function.

In this work, we use the nodal coordinates of the mesh as node features $v_i$ and the edge connectivity matrix as inputs to MeshGraphNet. The output labels consist of the four state variables $\rho_i, u_i, v_i, p_i$ defined at each node. The node and edge encoding encoder consists of five sequential linear layers with ReLU6 activation function \cite{Relu6}. We incorporate Dropouts after the first three encoder layers to prevent overfitting, followed by a layer normalization before applying the activation. The processor layer consists of eight sequential MLP layers with layer normalization and ReLU activation for processing the edge and node vectors. The decoder module contains four MLP layers, with layer normalization and ReLU6 activation applied to every layer except the output. Table \ref{tab:Meshgraphnet_arch} shows the network details used for this study. We choose the Adam optimizer with an initial learning rate of $1e^{-3}$ with weight decay $\lambda = 0.005$ and train the network for 5000 epochs with a batch size of fourteen. 

\begin{table}[ht]
\caption{Details of MeshGraphNet architecture.}
\label{tab:Meshgraphnet_arch}
\resizebox{\textwidth}{!}{%
\begin{tabular}{@{}lclccc@{}}
\toprule
\multicolumn{1}{c}{\textbf{Layer   type}} &
  \textbf{Number of  layers} &
  \multicolumn{1}{c}{\textbf{Number of nodes}} &
  \textbf{Activation} &
  \textbf{Layernorm} &
  \textbf{Dropout} \\ \midrule
Encoder   & 5 & {[}32, 64, 64, 64, 64{]}             & ReLU6 & Yes & Yes  \\
Processor & 8 & {[}64, 64, 64, 64, 64, 64, 64, 64{]} & ReLU  & Yes & None \\
Decoder   & 4 & {[}64, 64, 64, 4{]}                  & ReLU6 & Yes & None \\ \bottomrule
\end{tabular}%
}
\end{table}

\bibliographystyle{elsarticle-num} 
\bibliography{ref, reference,  reference_hypersonic_body}
\end{document}